\documentclass[lettersize,journal]{IEEEtran}
\usepackage{graphicx}
\usepackage{subcaption}
\usepackage{multirow}
\usepackage{booktabs}
\usepackage[table]{xcolor}
\usepackage{color}
\usepackage{colortbl}
\usepackage{tabularx}
\usepackage{bm}
\usepackage{url}
\usepackage{stackengine}
\usepackage{float}
\usepackage{verbatim}
\usepackage{array}
\usepackage{cite}
\usepackage{algorithm}
\usepackage{algorithmic}
\usepackage[colorlinks,linkcolor=blue]{hyperref}
\usepackage{amsmath,amssymb} 
\usepackage{amsfonts}
\usepackage{comment}
\usepackage{svg}
\usepackage{orcidlink}
\usepackage{textcomp}
\usepackage{stfloats}
\begin{document}


\title{S2C: Learning Noise-Resistant Differences for Unsupervised Change Detection in Multimodal Remote Sensing Images}

\author{Lei~Ding\orcidlink{0000-0003-0653-8373}, Xibing~Zuo, Danfeng~Hong,~\IEEEmembership{Senior Member,~IEEE,}, Haitao~Guo, Jun~Lu, Zhihui~Gong and Lorenzo~Bruzzone\orcidlink{0000-0002-6036-459X},~\IEEEmembership{Fellow,~IEEE}~

\thanks{L. Ding is with the Information Engineering University, Zhengzhou, China, and also with the Aerospace Information Research Institute, Chinese Academy of Sciences, Beijing, China (E-mail: dinglei14@outlook.com).}

\thanks{Xibing~Zuo, Haitao~Guo, Jun~Lu, and Zhihui~Gong are with the Information Engineering University, Zhengzhou, China.}

\thanks{D. Hong is with the Aerospace Information Research Institute, Chinese Academy of Sciences, Beijing, 100094, China, and also with the School of Electronic, Electrical and Communication Engineering, University of Chinese Academy of Sciences, 100049 Beijing, China. (e-mail: hongdf@aircas.ac.cn)}

\thanks{L. Bruzzone is with the Department of Information Engineering and Computer Science, University of Trento, 38123 Trento, Italy (E-mail: lorenzo.bruzzone@unitn.it).}

\thanks{This document is funded by the Natural Science Foundation of China under Grant 42201443. It is also funded by the Henan Provincial Key Technologies R \& D Program under Grant 242102211047. (Corresponding author: Lei Ding.)}}

\markboth{Manuscript under review}%
{Shell \MakeLowercase{\textit{et al.}}: Bare Demo of IEEEtran.cls for IEEE Journals}

\maketitle

\begin{abstract}
Unsupervised Change Detection (UCD) in multimodal Remote Sensing (RS) images remains a difficult challenge due to the inherent spatio-temporal complexity within data, and the heterogeneity arising from different imaging sensors. Inspired by recent advancements in Visual Foundation Models (VFMs) and Contrastive Learning (CL) methodologies, this research aims to develop CL methodologies to translate implicit knowledge in VFM into change representations, thus eliminating the need for explicit supervision. To this end, we introduce a Semantic-to-Change (S2C) learning framework for UCD in both homogeneous and multimodal RS images. Differently from existing CL methodologies that typically focus on learning multi-temporal similarities, we introduce a novel triplet learning strategy that explicitly models temporal differences, which are crucial to the CD task. Furthermore, random spatial and spectral perturbations are introduced during the training to enhance robustness to temporal noise. In addition, a grid sparsity regularization is defined to suppress insignificant changes, and an IoU-matching algorithm is developed to refine the CD results. Experiments on four benchmark CD datasets demonstrate that the proposed S2C learning framework achieves significant improvements in accuracy, surpassing current state-of-the-art by over 31\%, 9\%, 23\%, and 15\%, respectively. It also demonstrates robustness and sample efficiency, suitable for training and adaptation of various Visual Foundation Models (VFMs) or backbone neural networks. The relevant code will be available at: \href{github.com/DingLei14/S2C}{github.com/DingLei14/S2C}.
\end{abstract}



\section{Introduction}\label{sc1}

Change detection (CD) represents one of the most earliest and widely utilized technologies in remote sensing (RS)~\cite{bruzzone2009domain, wu2023fully}. Different from CD in surveillance \cite{lanza2011statistical}, streetview\cite{taneja2015geometric}, industrial \cite{roth2022towards} and medical \cite{li2020siamese} applications that mainly address image changes taken by fixed or mounted cameras, CD in RS leverages over-head platforms to identify and segment change regions on the Earth surface at different times \cite{bruzzone2012novel}. It has wide applications in urban development management \cite{benedek2011building}, land cover monitoring \cite{robin2010contrario}, damage assessment \cite{brunner2010earthquake}, etc. Recently, great advances have been achieved in RS CD utilizing deep learning (DL) techniques. State-Of-The-Art (SOTA) methodologies have obtained accuracy levels surpassing 90\% in the $F_1$ metric across various benchmark datasets for CD \cite{chen2021remote, ding2024samcd}. However, typical DL-based CD methods require large amounts of high-quality labeled data for training, which are difficult to collect due to the scarcity of change samples. Consequently, deployment of CD algorithms in real-world applications still faces significant challenges.

\begin{figure}[!t]
\centering
    \includegraphics[width=1\linewidth]{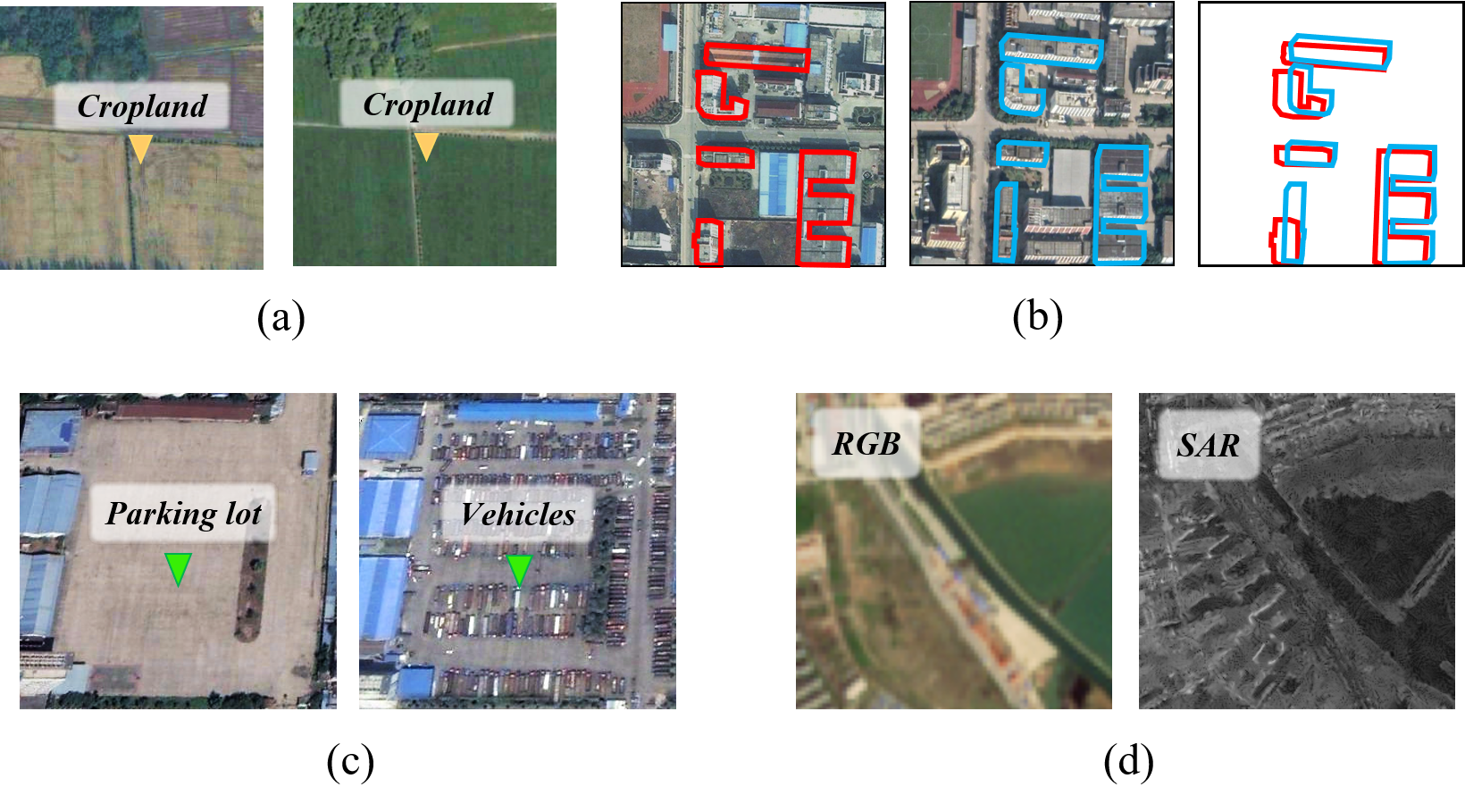}
    \caption{The major types of temporal noise in CD on HR RSIs include: (a) spectral/seasonal variations, (b) spatial misalignment, (c) insignificant changes, and (d) multimodal heterogeneity.}
\label{fig.challenge}
\end{figure}

To reduce dependence on training data, an increasing number of studies have been conducted on unsupervised change detection (UCD) in recent years. However, most of these investigations focus on UCD in medium-resolution multitemporal images \cite{chen2021self}. UCD in HR RS images presents greater challenges due to the increased spatial, temporal, and spectral complexity \cite{bruzzone2012novel}. Fig.\ref{fig.challenge} illustrates the main instances of noise encountered in the CD of RS images, including \textit{i) Spectral variations.} This can be attributed to seasonal variations or differences in imaging sensors and illumination conditions \cite{hong2018augmented}. \textit{ii) Spatial misalignment}. This may arise from varying image acquisition angles, optical distortion, or errors in image registration. \textit{iii) Insignificant changes.} In CD applications, only specific semantic changes are of interest, such as building changes in urban management and cropland changes in agriculture monitoring. Certain temporary randomness changes, such as the parked vehicles depicted in Fig. \ref{fig.challenge}(c), are often deemed as irrelevant noise. iv) Multimodal heterogeneity. Multi-temporal observations captured by different sensors are likely to exhibit substantial differences in radiometric patterns. Consequently, deep neural networks (DNNs) face significant challenges in learning to differentiate between relevant semantic changes and temporal radiometric differences in an unsupervised manner.
 
In recent studies, contrastive learning (CL) \cite{sohn2020fixmatch} and Visual Foundation Models (VFMs) \cite{Kirillov2023Segment} have been identified as two effective methodologies to mitigate data dependence in CD. The former leverages the intrinsic consistency within data, while the latter incorporates external knowledge to learn common semantic representations. Significant advances have been made in a semisupervised CD on HR RSIs leveraging CL \cite{yang2023revisiting, chen2022self} and VFMs \cite{ding2024samcd, zheng2024segment}. However, UCD in HR RSIs using either CL or VFM remains challenging due to the inherent spatio-temporal complexity within this task. In CL-based CD, various studies follow a consistency regularization framework \cite{sohn2020fixmatch}. Although this approach significantly enhances the generalization and robustness of feature representations, it still requires a certain proportion of training data. VFM-based CDs often employ VFMs, such as the Segment Anything Model \cite{Kirillov2023Segment}, to exploit semantic features and decode change masks \cite{zheng2024segment}. Nonetheless, a significant domain gap exists between the training domains of VFM and RS images \cite{ji2024segment}, adversely affecting their recognition capabilities. Furthermore, translating semantic features into CD results still necessitates supervised learning.

This paper aims to investigate the integration of CL and VFM to accomplish UCD in HR RS images. It is observed that these two methodologies effectively complement each other: CL provides self-supervised training objectives essential for adapting VFMs to the RS domain and for mapping the changes. Conversely, VFMs embed pixel-level semantic representations, a capability typically absent within conventional CL frameworks. Furthermore, we extend the existing CL frameworks to incorporate the spatio-temporal correlations unique to CD. Instead of focusing on learning robust and consistent embeddings, our approach also models the consistency and differences across multi-temporal observations.

The major technical contributions in this study can be summarized as follows:

\begin{itemize}
    \item Formulating a UCD framework that explicitly models the unsupervised learning of semantic changes. The proposed S2C framework, to the best of our knowledge, is also the first work that incorporates VFM into CL in the context of CD. It integrates multiple innovative designs and technologies, including CL, VFM, and Low-Rank Adaptation (LoRA), as well as an IoU-based refinement algorithm.
    \item Introducing two novel CL paradigms for UCD: Consistency-regularized Temporal Contrast (CTC) and Consistency-regularized Spatial Contrast (CSC), to capture differences and consistency across multi-temporal observations. Notably, the CTC presents a novel triplet learning strategy to model semantic changes, complementing literature studies that predominantly focus on similarities.
    \item Proposing a grid sparsity regularization that aims at promoting sparse and compact change mapping. The sparse calculations are executed at grid scales to avoid training collapse and ensure efficiency. This training objective is also applicable to other segmentation tasks that require sparse or compact results.
    \item Further extending S2C to unsupervised Multimodal Change Detection (MMCD). The inherent semantic alignment mechanism makes the S2C independent of the specific imaging modalities. With modifications to the image encoders and learning process, cross-modality change modeling is established within the S2C framework.
\end{itemize}

The developed methodology demonstrates significant improvements over the state-of-the-art (SOTA) methods. Specifically, the accuracy improvements are approximately $31\%$, $9\%$, and $23\%$ in $F_1$ across three HR benchmark datasets for CD. Moreover, the application of the proposed S2C learning framework to the Wuhan MMCD benchmark demonstrated an increase of $F_1$ by $15\%$ over the SOTA methods. These experimental findings establish the proposed S2C framework as a robust, accurate, sample-efficient, and modality-independent framework for UCD in RS images.
\section{Related Work}\label{sc2}

This section reviews the literature on unsupervised CD, MMCD, as well as the investigated key techniques including VFM and contrastive learning.

\subsection{Unsupervised Change Detection}

Unsupervised CD poses great challenges to DL-based approaches due to the absence of explicit supervision. To address this challenge, three main strategies have been developed, including feature difference mapping, generative representation, and knowledge transfer.

Difference mapping is an essential step in the CD that transforms the deep features into change representations. Literature techniques on feature difference analysis include Principal Component Analysis (PCA) ~\cite{Bruzzone2000diff, Gao2016Automatic}, change vector analysis \cite{saha2019unsupervised, wu2021unsupervised}, slow feature analysis ~\cite{Wu2014Slow, Du2019Unsupervised}, half-sibling regression \cite{Kondmann2022Spatial}, etc. Due to the absence of explicit supervision, these methods typically employ pretrained deep neural networks (DNNs) for feature extraction, resulting in weak semantic representations. Meanwhile, generative representation-based methods typically employ image generation methods to reduce the style differences between multitemporal observations, a strategy known as generative transcoding. The frequently employed image generation approaches include AutoEncoders (AEs) \cite{Chen2022Unsupervised} and Generative Adversarial Networks (GANs) \cite{noh2022unsupervised}. Differently from generative transcoding, Wu et al. \cite{wu2023fully} introduce a generative framework to iteratively optimize CD results. 

\subsection{Multimodal Change Detection}

In Earth monitoring applications, it is common that the multi-temporal images are captured by sensors with different imaging mechanisms \cite{hong2023cross}. MMCD extracts change regions with RS images of different modalities, such as optical, synthetic aperture radar (SAR) images, hyperspectral \cite{liu2019HyperCDReview}, digital surface models, etc. Key challenges in unsupervised MMCD involve aligning semantic representations from different image modalities. Recent literature methodologies to solve this challenge include: \textit{i) Generative transcoding.} Generative networks such as AEs~\cite{liu2018coupling, wu2022CommonalityAE} and GANs \cite{saha2021building, Luppino2024CAAE} are employed to learn transition patterns between different image modalities, thereby aligning their representations. \textit{ii) Graph modeling.} Graphs are employed to model the modality-invariant structural information and segment the difference regions~\cite{sun2021itertivegraph, Chen2022Unsupervised, sun2024LPEM}. \textit{iii) Metric learning.} Local distribution patterns or statistics are collected from multimodal images to measure similarity or distances \cite{prendes2015Multivariate, touati2019multimodal}. 

However, due to difficulties in the registration and annotation of MMCD datasets \cite{lv2022cdhetreview}, most MMCD methods are developed on a single pair of images with very few instances. For instance, several literature methods exhibit sensitivity to the tuning of hyperparameters. Their generalization to diverse areas with various change instances remains largely unevaluated.

\subsection{Contrastive Learning}

Contrastive learning (CL), a type of self-supervision approach, constructs and compares positive and negative pairs to exploit the semantic consistency in unlabeled data. An established paradigm of CL in visual recognition is to introduce weak-to-strong perturbations, thus regularizing the DNNs to learn robust semantic representations \cite{sohn2020fixmatch}. These perturbations can be introduced into input images or embedded features \cite{yang2023revisiting}.

In CD, bi-temporal images of the same/different regions are often utilized to construct contrastive pairs. In \cite{chen2021self} a simple contrastive learning paradigm for CD is introduced, where change pairs are constructed with cropped RSIs at the same and different locations. Differently, Bandara et al. \cite{bandara2022revisiting} introduce perturbations on the bi-temporal difference features and perform consistency-regularized CL. In \cite{mall2023changeaware} a CL framework for CD with long-term temporal observations is introduced.

Overall, CL is mostly utilized in semi-supervised CD \cite{yang2023revisiting, bandara2022revisiting} and weakly-supervised CD ~\cite{zhao2024pixellevel} to leverage the sparse supervision signals available. Literature methods on CL-based UCD predominately exploit the temporal similarity embeddings \cite{chen2021self}, yet there exists a notable gap in the exploration of temporal difference embeddings. In this study, we investigate the joint exploitation of temporal consistency and difference embeddings employing CL techniques.

\begin{figure*}[!t]
\centering
    \includegraphics[width=1\linewidth]{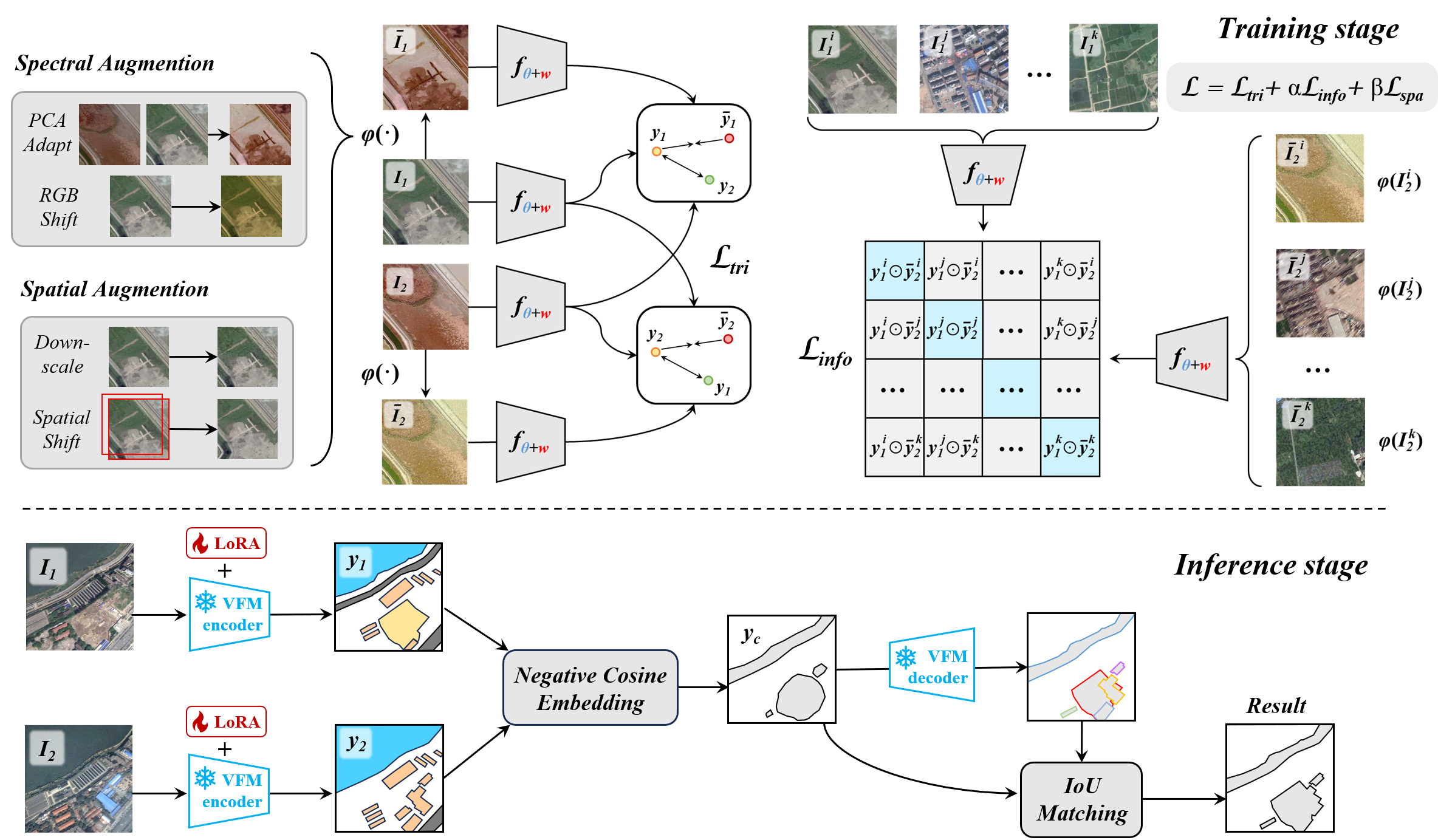}
    \caption{Overview of the proposed S2C framework for UCD. Triplet losses are calculated with bitemporal images and their augmented copies to learn temporal differences; discriminative losses are calculated between bitemporal images of different regions to learn temporal consistency. Random perturbations are introduced to simulate the spectral and spatial variations.}
\label{fig.flowchar}
\end{figure*}

\subsection{VFM-Based Change Detection}

In recent years, there has been a trend towards developing Visual Foundation Models (VFMs), such as CLIP \cite{radford2021CLIP} and Segment Anything Model (SAM) \cite{Kirillov2023Segment}, to acquire comprehensive recognition capabilities. VFMs are trained on large datasets to capture universal features applicable to various tasks. However, since these VFMs are mostly trained on common natural scenes, they demonstrate biases while applied to the recognition of RS scenes \cite{ji2024segment}. Considering the spectral and temporal characteristics of RSIs, several RS foundation models (FMs) have been developed, including SpectralGPT \cite{hong2024spectralgpt} and SkySense \cite{guo2024skysense}. However, employing these models for CD still necessitates incorporating CD-specific modules and performing fully supervised fine-tuning.

Considering that FMs contain implicit knowledge of the image content, several recent methods have explored employing FMs to achieve sample-efficient CD. SAM-CD\cite{ding2024samcd}, the first work on adapting VFMs to the RS domain using a semantic alignment technique, obtains high accuracy and demonstrates label-efficiency over several semisupervised CD methods. 

In \cite{wang2023cs}, SAM is utilized to generate pseudo labels from vague change maps used as prompts. In \cite{chen2024change}, Chen et al. employed SAM to achieve unsupervised CD between optical images and map data. In \cite{zheng2024segment}, zero-shot CD is achieved by measuring the similarity of SAM-encoded features. Dong et al. \cite{dong2024changeclip} utilized CLIP to learn visual-language representations to improve CD accuracy.

Despite these previous works on leveraging VFM for CD, UCD in HR RSIs is still a challenging tasks and the SOTA accuracy is limited. In this research, we further improve SAM-CD by incorporating self-supervision techniques to replace explicit supervision.

\section{Proposed Approach}\label{sc3}

This section first explains the motivation for learning noise-resistant representations, then introduces the S2C learning framework, and subsequently details the proposed technologies including contrastive change learning, grid sparsity loss, and the change mapping algorithms. Finally, the S2C methodologies are expanded to address unsupervised MMCD.

\subsection{Noise-resistant Semantic Embedding}\label{sc3.A}

DL-based CD essentially learns to project multi-temporal RS images ${I_1, I_2}$ into a binary change map $\mathbf{y}_c$. Let $f_\theta$ denote an encoding function parameterized by $\theta$, and $g$ being a projection function. This process can be represented as:
\begin{equation}\label{eq.cd}
    \mathbf{y_1} = f_\theta(I_1), \mathbf{y_2} = f_\theta(I_2), \mathbf{y_c} = g(\mathbf{y_1}, \mathbf{y_2})
\end{equation}
where $\mathbf{y_1}, \mathbf{y_2} \in \mathbb{R}^{c \times h \times w}$ are the learned semantic latent, $\mathbf{y_c} \in \mathbb{R}^{h \times w}$ is a change probabilistic map, $s$ and $h, w$ are the channel and spatial dimensions, respectively.

Let us denote the imaging process as $\Phi$, ground semantics as $s$, and semantic changes as $\delta$. In an ideal case where there is no temporal noise, this process can be re-written as:
\begin{equation}\label{eq.imaging}
    \mathbf{y}_s = f_\theta[\Phi_1(s)], \mathbf{y}_{s+\delta} = f_\theta[\Phi_2(s+\delta)], \mathbf{y_c} = g(\mathbf{y}_s, \mathbf{y}_{s+\delta})
\end{equation}
Since $\mathbf{y}_s$ and $\mathbf{y}_{s+\delta}$ share a common semantic space, $g$ can be implemented using a simple linear transformation with normalization. However, when considering practical cases that involve temporal noise (refer Fig.\ref{fig.challenge}), there exist insignificant changes (denoted $\epsilon$), spatial variance (denoted $\Omega$), and sensor differences ($\Phi_1 \neq \Phi_2$). Consequently, $I_2$ is projected into a different space:
\begin{equation}\label{eq.noise}
    f_\theta(I_2)=f_\theta\{\Omega[\Phi_2(s+\delta+\epsilon)]\}=\mathbf{y^{\prime}}_{s+\delta+\epsilon}
\end{equation}

Thus, the distance between $\mathbf{y}_s$ and $\mathbf{y^{\prime}}_{s+\delta+\epsilon}$ is nonlinear, and optimizing subsequent change embedding with $g$ typically requires supervised learning with task-specific labels.

To achieve unsupervised change learning, we utilize spatial and spectral augmentations to simulate various types of temporal noise, thus training a noise-resistant $f_\theta$ and formulating a training-free $g$. First, we apply augmentations functions $\phi$ on $I_1$ simulating the noise in Eq.(\ref{eq.noise}):
\begin{equation}\label{eq.phi_I1}
    \phi(I_1)= \hat{\Omega}\{\hat{\Phi}_{1 \rightarrow 2}[\Phi_1(s)]\} = \hat{\Omega}[\hat{\Phi}_{1,2}(s)],
\end{equation}
where $\hat{\Omega}$ is a set of spatial augmentations, $\hat{\Phi}_{1 \rightarrow 2}$ is a set of spectral augmentations that adapt the spectral distribution of $I_1$ approaching $I_2$. The augmentations are performed with stochastic parameters to simulate the imagining process with random noise, generating diverse training pairs. Further utilizing CL techniques elaborated in Sec.\ref{sc3.CL}, we can train $f_\theta$ towards projecting $I_1$ and $\phi(I_1)$ into similar representations, i.e., learning noise-invariant semantic latent:
\begin{equation}
    f_\theta(I_1) = \mathbf{\hat{y}}_s, 
    f_\theta[\phi(I_1)] = f_\theta\{ \hat{\Omega}[\hat{\Phi}_{1,2}(s)] \} \approx \mathbf{\hat{y}}_{s}
\end{equation}
These operations are also symmetrically performed on $I_2$ as:
\begin{equation}\label{eq.phi_I2}
\begin{aligned}
    \phi(I_2) = \hat{\Omega}\{\hat{\Phi}_{2 \rightarrow 1}[\Phi_2(s+\delta+\epsilon)]\} \approx \hat{\Omega}[\hat{\Phi}_{1,2}(s+\delta)],\\
    f_\theta(I_2) = \mathbf{\hat{y}}_{s+\delta}, f_\theta[\phi(I_2)] \approx f_\theta\{ \hat{\Omega}[\hat{\Phi}_{1,2}(s)] \} \approx \mathbf{\hat{y}}_{s+\delta}
\end{aligned}
\end{equation}
where $\epsilon$ denoting the subtle changes can be diminished since $\hat{\Omega}$ includes spatial filtering operations.
This enables modeling the semantic differences between $\mathbf{\hat{y}}_s, \mathbf{\hat{y}}_{s+\delta}$ in a common latent space, enabling a training-free formulation of $g$:
\begin{equation}
    \mathbf{y_c} = g(\mathbf{\hat{y}}_s, \mathbf{\hat{y}}_{s+\delta})
\end{equation}

\subsection{Overview of the S2C framework}
The theoretical analysis in Sec.\ref{sc3.A} offers a simplified understanding of the optimizing goals. In the following, we elaborate on the S2C architecture that trains $f_\theta$ to exploit noise-invariant semantic representations.

As depicted in Fig.\ref{fig.flowchar}, S2C is a UCD framework consisting of distinct stages of training and inference. During the training phase, a set of augmentation operations are first performed on the input images, corresponding to $\phi(\cdot)$ in Eq.(\ref{eq.phi_I1})(\ref{eq.phi_I2}). The augmentations comprise a sequential combination of spectral and spatial operations executed randomly at each training iteration, including \textit{RGBshift}, \textit{PCA adaptation}, \textit{down-sampling} and \textit{random shifting}. \textit{PCA adaptation} blends the spectral distribution of $I_1$ and $I_2$. The spatial operations are performed to simulate spatial misalignment and imaging degradation/distortion, while the spectral operations replicate imaging and seasonal variations. Collectively, their combination simulate stochastic injection of diverse temporal noise.

Subsequently, utilizing the constructed pre- and after-augmentation image sequences, two self-supervised CL strategies are jointly employed to learn task-specific semantic features. Although this training process can be performed from scratch using common DNNs, employing VFM as feature encoders empirically leads to better accuracy. In addition, considering the limitations of VFM in processing RS images \cite{ji2024segment}, we employ a VFM augmented with additional parameters $w$, denoted as $f_{\theta + w}$. $w$ is a trainable parameter implementing LoRA \cite{hu2021lora} (with $rank=4$), a parameter-efficient technique extensively used to adapt VFMs to a particular domain of interest. $\theta$ is frozen to retain pre-trained visual knowledge, whereas $w$ are the LoRA weights trained with CL paradigms to exploit temporal semantic features. The VFM can be any off-the-shelf models, as its inner structure is not modified in our S2C architecture.

The training process is conducted within two CL paradigms to learn the CD-relevant semantic representations. The two CL paradigms are introduced to embed difference and consistency representations, respectively, which are elaborated in Sec.\ref{sc3.CL}. The associated loss functions are $\mathcal{L}_{tri}$ and $\mathcal{L}_{info}$, respectively. In addition, we further introduce a sparsity regularization objective to learn sparse and compact change representations, noted as $\mathcal{L}_{spa}$. The joint training objective is:
\begin{equation}\label{eq.losses}
    \mathcal{L} = \mathcal{L}_{tri} + \alpha \mathcal{L}_{info} + \beta \mathcal{L}_{spa}
\end{equation}
where $\alpha$ and $\beta$ are two weighting parameters.

In the inference phase, the semantic latent $\mathbf{y}_1$ and $\mathbf{y}_2$ are first mapped to a coarse change map, then refined using the VFM decoder and an IoU matching function. The details are elaborated in \ref{sc3.chg_map}.

\subsection{Contrastive Change Learning} \label{sc3.CL}

\begin{figure*}[t]
\centering
    \includegraphics[width=0.95\linewidth]{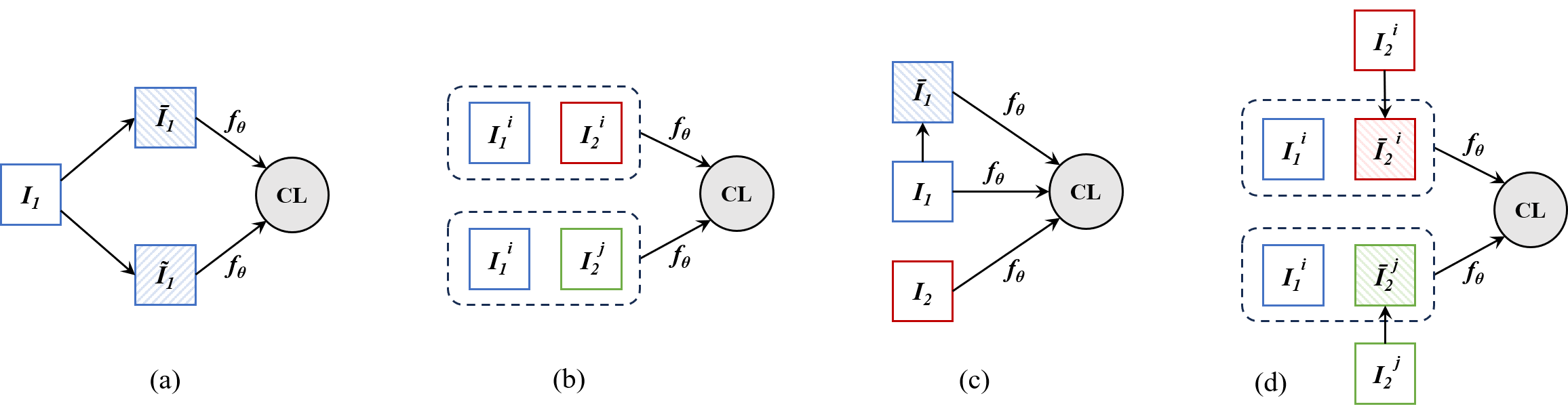}
    \caption{Comparison of CL paradigms in CD. (a) \textbf{Consistency regularization}: $f_\theta$ extracts stable representations across weak/strong perturbations; (b) \textbf{Spatial contrast}: $f_\theta$ distinguishes same/different regions; (c) Proposed \textbf{Consistency-regularized Temporal Contrast} (CTC): $f_\theta$ identifies temporal differences independent of spectral or seasonal variations, and (d) Proposed \textbf{Consistency-regularized Spatial Contrast} (CSC): $f_\theta$ distinguishes same/different regions despite perturbations.}
\label{fig.CL_paradigms}
\end{figure*}

Before introducing the proposed CL paradigms, let us first review the two typical CL paradigms in CD, and analyze their usage and limitations.

\textit{1) Consistency Regularization (CR).} As depicted in Fig.\ref{fig.CL_paradigms}(a), a DNN $f_{\theta}$ learns to improve the robustness and generalization of feature embeddings. An image $I$ is first augmented with weak and strong transformations, resulting in two copies $\Tilde{I}$ and $\Bar{I}$. Then a distance loss is calculated between the two copies to ensure consistency across perturbations.

Since this learning paradigm does not explicitly model differences/similarities, it is often adopted in semi-supervised \cite{bandara2022revisiting} or weakly-supervised \cite{zhao2024pixellevel} learning settings to extend the CD insights learned with limited samples.

\textit{2) Spatial Contrast (SC).} As illustrated in Fig.\ref{fig.CL_paradigms}(b), $f_{\theta}$ learns to differentiate between bitemporal image pairs $[I_1^i, I_2^i]$ of the same region $i$ and $[I_1^i, I_2^j]$ of different regions $i$ and $j$. This drives $f_{\theta}$ to learn consistent embeddings against temporal variations. Areas with high similarity are identified as $unchanged$, whereas their opposites are detected as $changed$ \cite{chen2022selfsupervised}.

However, we identify that there are several limitations in this paradigm: i) changes are identified through negative embedding of the similarities rather than through explicit modeling. This often causes sensitivity to noise. ii) $f_{\theta}$ focuses on the discriminative elements within a region, such as certain edges or corners, rather than effectively exploiting the local semantic context. 

Considering these limitations, we introduce two novel CL paradigms specifically tailored to the context of CD.

\textbf{Consistency-regularized Temporal Contrast} (CTC). An RS image $I_1$ is first augmented with a transform function $\phi(\cdot)$, producing a copy $\Bar{I}_1$. Subsequently, $I_1$ is employed as an anchor for comparison with both a positive sample $\Bar{I}_1$ and a negative sample $I_2$. $\phi(\cdot)$ simulates spectral and spatial noise between multi-temporal observations, as illustrated in Fig.\ref{fig.flowchar}. Consequently, $f_\theta$ learns to exploit noise-invariant difference representations, i.e., the semantic changes. With greater details, Fig.\ref{fig.flowchar} illustrates the CTC paradigm with bi-directional comparisons within $[I_1, \Bar{I}_1, I_2]$ and $[I_2, \Bar{I}_2, I_1]$. 

A triplet training objective $\mathcal{L}_{tri}$ using cosine distance is utilized for comparisons within the triplets. This is to align with the cosine difference embedding during the inference stage. The calculations are as follows:
\begin{equation}
\begin{aligned}
    \mathcal{L}_{tri} = max[cos(\mathbf{y_1}, \mathbf{y_2})-cos(\mathbf{y_1}, \mathbf{\Bar{y}_1})+m, 0]\\
    + max[cos(\mathbf{y_2}, \mathbf{y_1})-cos(\mathbf{y_2}, \mathbf{\Bar{y}_2})+m, 0]
\end{aligned}
\end{equation}
where $m=1$ is a margin parameter to promote separation between the anchor and positive.

\textbf{Consistency-regularized Spatial Contrast} (CSC). This contrastive learning paradigm integrates CR into typical SC learning, thereby enhancing the embedding of spatial consistency against perturbations. CSC alleviates the vulnerability to noise inherent in the SC paradigm by incorporating transformation $\phi(\cdot)$. The transformations, particularly the spatial transformations, reduce dependence on high-frequency spatial details, thereby prompting the exploitation of local semantic patterns such as color and texture.

We have introduced an additional variation in CSC, i.e., the calculation of consistency at each spatial position. Given a batch consisting of $N$ paired RS images $\{[I_1^i, I_2^i], [I_1^j, I_2^j], ..., [I_1^k, I_2^k]\}$, we first apply $\phi(\cdot)$ on each of the temporal images, thus getting two sets of augmented images. These images are further encoded with $f_{\theta+w}$, resulting in 4 sets of features: $[\mathbf{y_1^i, y_1^j, ..., y_1^k}]$, $[\mathbf{y_2^i, y_2^j, ..., y_2^k}]$, $[\mathbf{\Bar{y}_1^i, \Bar{y}_1^j, ..., \Bar{y}_1^k}]$ and $[\mathbf{\Bar{y}_2^i, \Bar{y}_2^j, ..., \Bar{y}_2^k}]$. We then calculate the co-occurrences between them, resulting in two matrices each with $N \times N$ dimensions, as illustrated in Fig.\ref{fig.CL_paradigms}(d). We utilize an infoNCE loss function to effectively train $f_{\theta+w}$ for the differentiation of genuine image pairs. It is calculated across both temporal phases, represented as:
\begin{equation} \label{eq.info_loss}
\begin{aligned}
    \mathcal{L}_{info} = -\frac{1}{N} \sum_{u=1}^{N} \log \left[ \frac{\exp \left( \mathbf{y}_1^u \odot \mathbf{\Bar{y}}_2^u \right)}{\sum_{v=1}^{N} \exp \left( \mathbf{y}_1^u \odot \mathbf{\Bar{y}}_2^v \right)} \right]\\
    -\frac{1}{N} \sum_{u=1}^{N} \log \left[ \frac{\exp \left( \mathbf{y}_2^u \odot \mathbf{\Bar{y}}_1^u \right)}{\sum_{v=1}^{N} \exp \left( \mathbf{y}_2^u \odot \mathbf{\Bar{y}}_1^v \right)} \right]
\end{aligned}
\end{equation}
where $\odot$ denotes a spatial similarity function that we introduce in this study. Instead of pooling the spatial features into single vectors for similarity calculation \cite{chen2022selfsupervised}, we compute the similarity at each spatial patch $p$, denoted as:
\begin{equation} \label{eq.sim_calc}
    \mathbf{y} \odot \mathbf{\Bar{y}} = \frac{1}{w \times h} \sum_{p} \left( \frac{\mathbf{y}^p \cdot \mathbf{\Bar{y}}^p}{|\mathbf{y}^p||\mathbf{\Bar{y}}^p|} \right)
\end{equation}
Both $\mathcal{L}_{tri}$ and $\mathcal{L}_{info}$ are calculated based on cosine similarity. While $\mathcal{L}_{tri}$ embeds appearance-invariant temporal differences, $\mathcal{L}_{info}$ embeds noise-resilient temporal consistencies. Therefore, when a certain temporal consistency pattern is captured in CSC, it suppresses the difference representations of the same area in CTC.

\subsection{Grid Sparsity loss} \label{sc3.loss_sparse}

Changed objects are commonly sparsely distributed in RS images and each present as compact regions. In contrast, edges and points are often associated with noise. Although training objectives that promote sparse representations have been explored in the literature, they typically calculate and penalize the average value of $\mathbf{y_c}$ \cite{bandara2023deep}. However, this approach does not guarantee sparsity, as there exists a trivial solution of learning an additional bias term on $\mathbf{y_c}$.

Differently, we propose a novel grid sparsity loss where sparsity is assessed at the level of each local grid rather than at each pixel. Considering the frequency of changes along with spatial resolution in an RS image, first, we predefine a sparsity threshold $T$ as well as a grid size $d$. Subsequently, the average density of each grid $\mathbf{g}$ is calculated and ranked, while a {\footnotesize$1-T$} ratio of grids with the lowest density are selected for loss calculation as follows:
\begin{equation} \label{eq.sparse_loss}
\begin{aligned}
    \mathcal{L}_{spa} = max\{ \frac{1}{n} \sum^{n} [sort\uparrow(y^\mathbf{g})], 0\},\\
    y^\mathbf{g} = \frac{1}{d*d} \sum_{p \in \mathbf{g}} \mathbf{y}_c^p, 
    n = wh*(1-T)/d^2
\end{aligned}
\end{equation}

We empirically set $d=16$ for HR RS images. This regularization objective ensures that a proportion of less than {\footnotesize$1-T$} potential changes exhibit high values, whereas the insignificant change representations in other areas are minimized.

\subsection{Change mapping} \label{sc3.chg_map}

\begin{algorithm}[t]
    \caption{Algorithm for IoU Matching and Refinement}
    \label{AlgorithmD}  \label{Algorithm.IoU_matching}
    \begin{algorithmic}[1]
        \renewcommand{\algorithmicrequire}{\textbf{Input:}}
        \renewcommand{\algorithmicensure}{\textbf{Output:}}
        \REQUIRE Change probability map $\mathbf{y}_c$,\\
                 VFM-extracted bi-temporal object masks $M_1$, $M_2$,\\
                 Parameter: IoU threshold $T_{IoU}$;
        \ENSURE Refined Change Map $M_c$;
        \FORALL{$m_i \in M_1, m_j \in M_2$}
        \IF {$(m_i \cap m_j)/(m_i \cup m_j)>T_{IoU}$}
            \STATE $M_{1} \leftarrow M_{1} \setminus \{m_i\}$
            \STATE $M_{2} \leftarrow M_{2} \setminus \{m_j\}$
        \ENDIF
        \ENDFOR
        \STATE $M_{12} \leftarrow M_{1} \cup M_{2}$
        \FORALL{$m \in M_{12}$}
        \IF {$m \odot \mathbf{y}_c / \sum_{p} m_p >T_{IoU}$}
            \STATE $M_c \leftarrow M_{c} \cup \{m\}$
        \ENDIF
        \ENDFOR
 \RETURN $M_c$ 
    \end{algorithmic}
\end{algorithm}

The use of VFM and CL methodologies aims to enhance the effective exploitation of semantic contexts across multi-temporal image domains. In the inference phase, the major challenge lies in accurately mapping fine-grained changes. We employ a coarse-to-fine refinement strategy. First, a coarse change probability map $\mathbf{y}_c$ is derived by projecting the negative cosine embedding of the bi-temporal semantic embeddings:
\begin{equation}
    \mathbf{y}_c = \sigma[-cos(\mathbf{y}_1, \mathbf{y}_2)*\eta]
\end{equation}
where $\sigma$ is a $sigmoid$ function and $\eta=ln(1/0.07)$ is a scaling factor defined following literature practice \cite{deuser2023sample4geo}.

Then, we employ a pretrained VFM decoder $g_{\gamma}$ to segment two groups of bi-temporal masks $M_1=\{m_1^1, m_1^2, ..., m_1^k\}$ and $\{M_2=m_2^1, m_2^2, ..., m_2^k\}$ using the spatial prompts generated on high-response regions in $\mathbf{y}_c$. Given the logic implication of \textit{change}, high-overlap objects in $M_1$ and $M_2$ can be inferred as false alarms. Therefore, we use an XOR-alike matching algorithm (denoted $\ominus$) to merge $M_1$ and $M_2$ while eliminating the objects with large overlaps:
\begin{equation}
    M_{12} = M_1 \ominus M_2
\end{equation}
We further implement an Intersection-over-Union (IoU) analysis between $\mathbf{y}_c$ and $M_{12}$ to match the VFM-generated masks with the high-confidence regions in $\mathbf{y}_c$. The matched objects replace their counterparts in $\mathbf{y}_c$ as the changed items. For more details, a pseudo-code of this IoU analysis and matching algorithm is provided in Algorithm \ref{Algorithm.IoU_matching}.

Using this change mapping algorithm, the coarse predictions derived from the DNN are refined into detailed CD results, aligning with the spatial details present in the HR imagery.

\subsection{S2C for Unsupervised MMCD}

\begin{figure}[t]
\centering
    \includegraphics[width=1\linewidth]{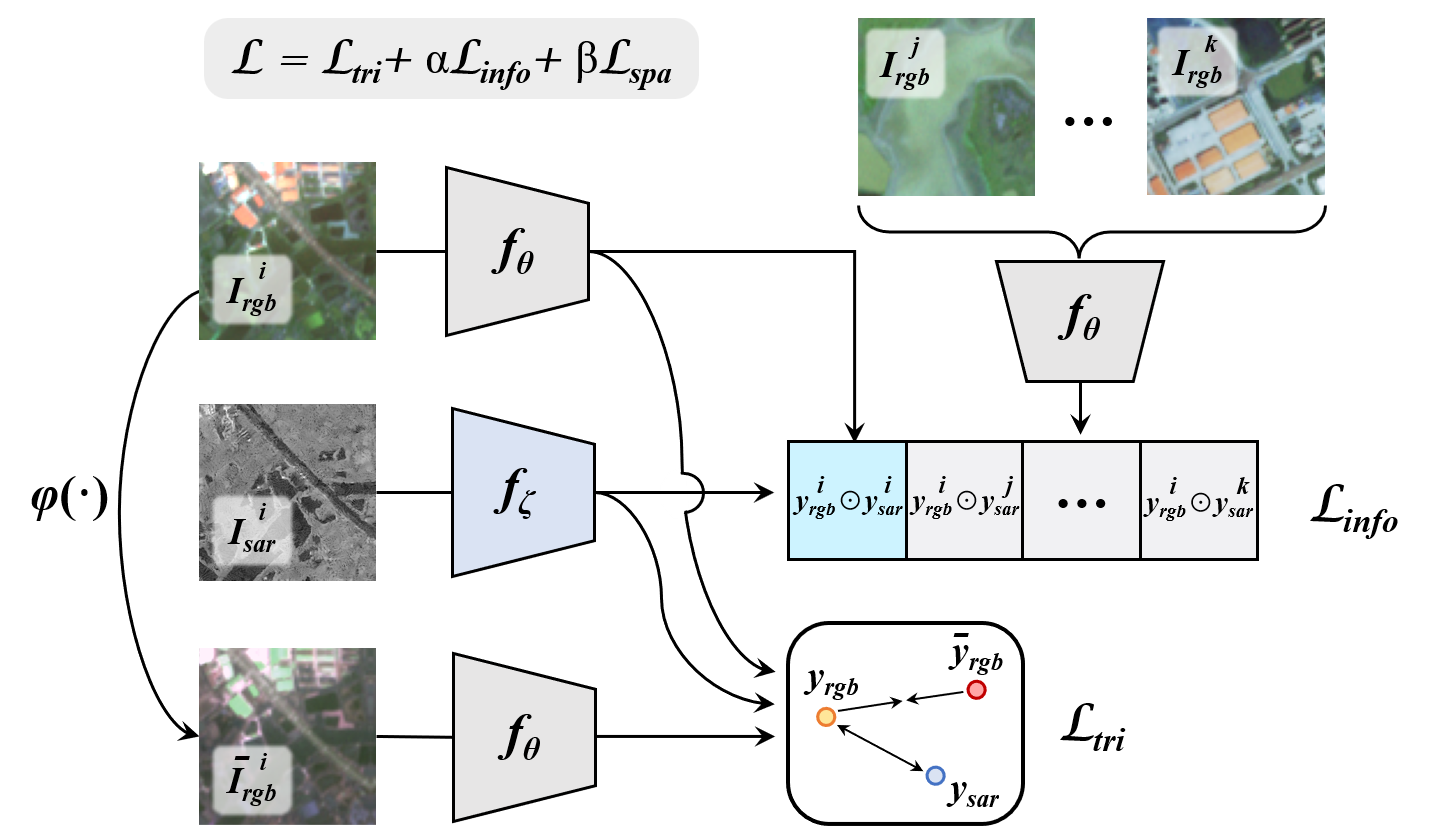}
    \caption{Illustration of the application of the proposed S2C for UCD in multimodal RS images. This learning framework applies to not only optical and SAR data, but also other image modalities.}
\label{fig.Het_flowchart}
\end{figure}

RS images observed by specialized sensors such as Synthetic Aperture Radar (SAR) and Infrared (IR) scanners demonstrate a notable modality difference compared to standard optical images. This significant domain gap precludes VFMs from extracting meaningful semantic representations \cite{ji2024segment}. Leveraging the joint modeling of consistency and differences, the S2C framework is capable of modeling modality-invariant change representations over multi-temporal observations.

Fig.\ref{fig.Het_flowchart} depicts the adapted S2C training pipeline for multimodal RS images, where the UCD process is exemplified through a pair of optical and SAR images. The semantic-to-change mapping is achieved through alignment of latent semantics, making it invariant to specific imaging modalities. Let us denote $I_{rgb}^i$ and $I_{sar}^i$ as a pair of multimodal images pertaining to region $i$. Two independent encoders with parameters $\theta$ and $\zeta$ are trained to extract features from optical and SAR images, respectively. It is worth noting that the VFMs are typically not applicable to SAR data and, therefore are not utilized in $f_\zeta$. However, with the continuous development of RS VFMs \cite{guo2024skysense, hong2024spectralgpt}, certain models may demonstrate the efficacy of semantic embedding within a specific domain. While we did not identify an optimal VFM for common SAR data, readers are encouraged to explore emerging VFMs within the S2C framework.

After feature embedding, the CTC and CSC paradigms are utilized to learn domain-invariant differences and consistencies. An adjustment in implementing the CTC is that it is asymmetrically applied, unlike its use with dual optical images. Given that optical data possess more comprehensive spatial information, the calculation of $\mathcal{L}_{tri}$ is centric to the optical data, i.e.:

\begin{equation} \label{eq.het_triplet_loss}
    \mathcal{L}_{tri} = max[cos(\mathbf{y}_{rgb}, \mathbf{y}_{sar})-cos(\mathbf{y}_{rgb}, \mathbf{\Bar{y}}_{rgb})+m, 0]
\end{equation}
where $\mathbf{y}_{rgb}$, $\mathbf{y}_{sar}$ and $\mathbf{\Bar{y}}_{rgb}$ are the semantic features encoded from $I_{rgb}$, $I_{sar}$ and $\phi(I_{rgb})$, respectively. The augmentation operations outlined in $\phi$ correspond to those detailed in Sec.\ref{sc3.CL}.

The presence of a domain gap poses more challenges to learning temporal consistency. Consequently, CSC learning is conducted directly on the original image sets $[\mathbf{y}_{rgb}^i, \mathbf{y}_{rgb}^j, ..., \mathbf{y}_{rgb}^k]$ and $[\mathbf{y}_{sar}^i, \mathbf{y}_{sar}^j, ..., \mathbf{y}_{sar}^k]$, rather than using their augmented counterparts. The loss function is computed as:
\begin{equation} \label{eq.het_info_loss}
\begin{aligned}
    \mathcal{L}_{info} = -\frac{1}{N} \sum_{u=1}^{N} \log \left[ \frac{\exp \left( \mathbf{y}_{rgb}^u \odot \mathbf{y}_{sar}^u \right)}{\sum_{v=1}^{N} \exp \left( \mathbf{y}_{rgb}^u \odot \mathbf{y}_{sar}^v \right)} \right]\\
    -\frac{1}{N} \sum_{u=1}^{N} \log \left[ \frac{\exp \left( \mathbf{y}_{sar}^u \odot \mathbf{y}_{rgb}^u \right)}{\sum_{v=1}^{N} \exp \left( \mathbf{y}_{sar}^u \odot \mathbf{y}_{rgb}^v \right)} \right]
\end{aligned}
\end{equation}

The comprehensive training objective is consistent with that in Eq.\ref{eq.losses}. The simultaneous application of CTC and CSC learning drives $f_\theta$ and $f_\zeta$ to align semantic representations across different domains, thus mapping domain-invariant changes. In the inference stage, since typical VFM decoders are unable to segment SAR objects, the refining algorithm is omitted. The change probability maps $\textbf{y}_c$ is directly binarized to map the multimodal changes.

\section{Experiments}\label{sc4}

\subsection{Datasets and Evaluation Metrics}
To test the effectiveness of the proposed method, experiments are conducted on three RS datasets with varied data distributions and semantic annotations. The experimental datasets include CLCD \cite{liu2022cnntransformer}, SECOND (binary) \cite{yang2022asymmetric} and Levir \cite{chen2020spatialtemporal}. In addition, MMCD experiments are conducted on the Wuhan dataset \cite{zhang2022domain}. Table \ref{Table.Datasets} presents an overview of each dataset. Since the experimental methods are unsupervised, we do not use any labels within the training set. Notably, the Levir dataset focuses solely on building changes, while CLCD, SECOND and Wuhan encompass various change types. Its larger spatial size and sparse change instances make it more challenging in the context of UCD. 

\begin{table}[ht]
    \centering
    \caption{Summary of the main characteristics of the experimental datasets.} \label{Table.Datasets}
    \resizebox{1\linewidth}{!}{%
    \begin{tabular}{c|c|c|c|c|c}
    \toprule
        \multirow{2}*{Datasets} & \multirow{2}*{Platform} & \multirow{2}*{Resolution} & Image & Dataset & Change \\
        & & & Size & Size & Type \\
        \hline
        CLCD & satellite & 0.5-2m & 512×512 & 600 & agricultural \\
        SECOND & aerial & 0.5-3m & 512×512 & 4,662 & land cover \\
        Levir & satellite & 0.5m & 1024×1024 & 637 & building \\
        \hline
        \multirow{2}*{Wuhan} & satellite & vis: 10m & \multirow{2}*{256×256} & \multirow{2}*{600} & \multirow{2}*{land cover} \\
        & (RGB\&SAR) & sar: 3m & & & \\
    \bottomrule
    \end{tabular}}
\end{table}

We adopt the most commonly used accuracy metrics in CD \cite{ding2024samcd, chen2023exchange} and binary segmentation tasks \cite{ding2021adversarial}, including Overall accuracy (\textit{OA}), Precision (\textit{Pre}), Recall (\textit{Rec}), and $F_1$ score. In these metrics, \textit{Pre} indicates the proportion of true positives among classified positives, while \textit{Rec} is the measure of identifying true positives. $F_1$ is the harmonic mean of \textit{Pre} and \textit{Rec}, therefore is more comprehensive in assessing the accuracy.

\begin{table}[t]
\centering
    \caption{Quantitative results of ablation study (tested on CLCD).}
    \resizebox{1\linewidth}{!}{%
        \begin{tabular}{l|c|cccc}
        \toprule
            \multirow{2}*{Methods} & \multirow{2}*{Backbone} & \multicolumn{4}{c}{Accuracy (\%)}\\
            \cline{3-6}
            & & $OA$ & $Pre$ & $Rec$ & $F_1$ \\
            \hline
            effi.SAM + CVA & effi.SAM (vit-t) & 61.24 & 13.90 & 81.01 & 23.73 \\
            effi.SAM + SC & effi.SAM (vit-t) & 92.23 & 45.76 & 24.07 & 31.55 \\
            \hline
            S2C (CSC only) & effi.SAM (vit-t) & 90.93 & 36.55 & 29.81 & 32.84 \\
            S2C (CTC only) & effi.SAM (vit-t) & 85.98 & 28.25 & 57.40 & 37.86 \\
            S2C (CSC + $\mathcal{L}_{spa}$) & effi.SAM (vit-t) & 89.71 & 33.35 & 38.33 & 35.67\\ 
            S2C (CTC + $\mathcal{L}_{spa}$) & effi.SAM (vit-t) & 91.06 & 39.52 & 38.03 & 38.76 \\
            S2C (CSC + CTC) & effi.SAM (vit-t) & 87.35 & 31.01 & 57.12 & 40.19 \\
            S2C & effi.SAM (vit-t) & 90.57 & 39.04 & 47.65 & 42.92 \\
            S2C & effi.SAM (vit-s) & 89.58 & 37.51 & \textbf{60.21} & 46.22  \\
            S2C + \textit{IoU refine} & effi.SAM (vit-s) & \textbf{91.47} & \textbf{43.85} & 52.28 & \textbf{47.69} \\
            \hline
            S2C (w/o. VFM) & ResNet18 & 87.75 & 26.80 & 37.34 & 31.21 \\
            S2C (w/o. VFM) & ResNet34 & 86.28 & 28.41 & 55.56 & 37.60 \\
            S2C & fastSAM & 86.65 & 24.71 & 38.81 & 17.78 \\
            S2C & SAM (vit-b) & 85.41 & 22.23 & 38.44 & 28.17  \\
            S2C & effi.SAM (vit-t) & 90.57 & 39.04 & 47.65 & 42.92  \\
            S2C & effi.SAM (vit-s) & 89.58 & 37.51 & 60.21 & 46.22  \\
            S2C & Dino-v2 (vit-b) & 93.95 & 59.04 & \textbf{61.24} & 60.12 \\
            S2C + \textit{IoU refine} & Dino-v2 (vit-b) & \textbf{94.46} & \textbf{63.82} & 59.12 & \textbf{61.38} \\
        \bottomrule
        \end{tabular} \label{Table.Ablation} }
\end{table}

\subsection{Implementation Details}\label{sc4.implement}
The training of S2C is performed using cropped images of $512 \times 512$ pixels over 20 epochs. The trained weights with the highest accuracy on the validation set are saved for subsequent evaluation on the test set. The training batch size depends on the backbone to fit in GPU memory, usually exceeding 12 in our implementation. The learning rate $lr$ is initially set to 0.01, and is updated at each iteration as: $0.01*(1-iterations/total\_iterations)^{1.5}$. The optimization algorithm is the Stochastic Gradient Descent with Nesterov momentum. The weighting parameters in eq.\ref{eq.losses} are set to $\alpha=0.2, \beta=1$, while $\alpha$ can be adjusted across datasets to balance $\mathcal{L}_{tri}$ and $\mathcal{L}_{info}$ in training dynamics. An sensitivity analysis on these paramenters are provided in Sec.\ref{sc.ablation}. , The sparsity threshold $T$ in Eq.(\ref{eq.sparse_loss}) is set according to the change sparsity in different datasets: $T=0.2$ for the CLCD and Wuhan datasets, $T=0.2$ for the Levir dataset with very sparse changes, and $T=0.4$ for the SECOND dataset with more change instances.

Apart from \textit{strong augmentations} as detailed in Sec.\ref{sc3.CL}, random flipping operations are also introduced as \textit{weak augmentations}. These operations are executed on each of the two temporal images to enhance sample diversity, rather than induce spatial or spectral perturbations.


\subsection{Ablation Study}\label{sc.ablation}

\textbf{Quantitative Evaluation.} An ablation study is conducted through cumulative integration of the proposed methodologies, including the CTC, CSC, grid sparse loss ($\mathcal{L}_{spa}$), and IoU matching and refinement (\textit{IoU refine}). Given that the proposed method employs both VFM and CL, an intuitive strategy is to combine these two techniques as a baseline approach. However, VFM alone is not capable of UCD, and there is no existing literature approach (to the best of our knowledge) that integrates CL with VFM for CD. Therefore, we implement these two approaches as the baseline: i) applying CVA and clustering on the VFM-encoded semantic features for CD following the practice in \cite{zheng2024segment}, and ii) conducting CS-based CL using the VFM features. Due to computational constraints, we employ \textit{efficient-SAM}(vit-t) \cite{xiong2024efficientsam} as a frozen encoder ($\theta$) in the initial baseline, which is an efficient variant of SAM \cite{Kirillov2023Segment}.

The quantitative results are presented in Table \ref{Table.Ablation}. As indicated by the results of \textit{eff.SAM + CVA}, direct change analysis on VFM features leads to suboptimal accuracy. By contrast, the foundational method that integrates VFM with SC-based CL demonstrates significant accuracy. The proposed CTC and CSC further surpass the conventional SC-based CL paradigm. It is worth noting that the CTC alone outperforms both SC and CSC by a large margin, establishing it as the leading single CL paradigm for CD. Meanwhile, the proposed grid sparsity regularization also exhibits notable effectiveness. Its addition to each CL paradigm results in an average enhancement of $~2\%$ in $F_1$. Adding CSC improves the robustness of CTC against temporal noise, leading to an increase of over $3\%$ in $F_1$. The refining algorithm substantially enhances the precision of the results, providing an increase of $1.47\%$ in $F_1$ and approximately $2\%$ in $OA$.

\begin{figure}[t]
\centering
    \includegraphics[width=0.8\linewidth]{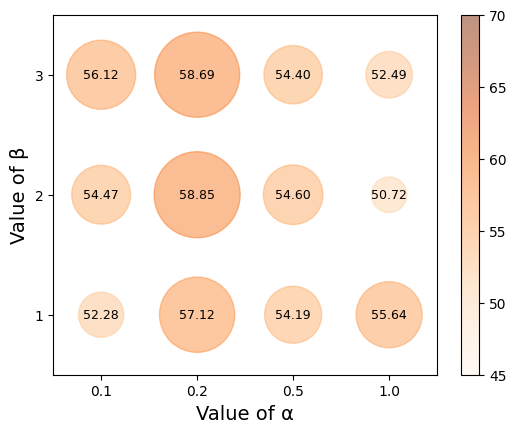}
    \caption{$F_1$ (\%) obtained by S2C with different weighting parameters.}
\label{fig.param_acc}
\end{figure}

\textbf{S2C with different backbones.} While S2C is introduced as a methodology that integrates CL and VFM, the core technique employed is a UCD framework, which is adaptable to other types of DNNs. Table \ref{Table.Ablation} also presents an evaluation of S2C utilizing various different backbones, including a vanilla ResNet \cite{he2016resnet} and several other VFMs. Surprisingly, the implementation of S2C with a simple ResNet34 backbone still yields considerably high accuracy. This further confirms its efficacy as a general framework for UCD.

Compared to using conventional DNNs, employing VFMs as backbone greatly improves the $Pre$ of S2C. This can be attributed to the rich semantic contexts inherent to VFMs, which significantly facilitate the discrimination of semantic changes. The implemented VFMs include SAM \cite{Kirillov2023Segment}, fastSAM \cite{zhao2023fast}, efficient SAM \cite{xiong2024efficientsam} and Dino-v2\cite{oquab2024dinov2}. Among SAM and its variants, efficient SAM obtains the highest accuracy. Its parameter size is considerably reduced compared to the original SAM, thereby improving its convergence in the context of UCD. Employing Dino-v2 results in the highest accuracy, with an advantage of approximately $12\%$ compared to the other backbones. Therefore, Dino-v2 is selected as the backbone in the subsequent experiments.

\begin{figure*}[t]
\centering
    \setlength{\tabcolsep}{1pt}
    \begin{tabular}{>{\centering\arraybackslash}m{0.4cm}>{\centering\arraybackslash}m{1.8cm}>{\centering\arraybackslash}m{1.8cm}>{\centering\arraybackslash}m{1.8cm}>{\centering\arraybackslash}m{1.8cm}>{\centering\arraybackslash}m{1.8cm}>{\centering\arraybackslash}m{1.8cm}>{\centering\arraybackslash}m{1.8cm}>{\centering\arraybackslash}m{1.8cm}>{\centering\arraybackslash}m{1.8cm}}
        (a)&
        \includegraphics[width=1.8cm]{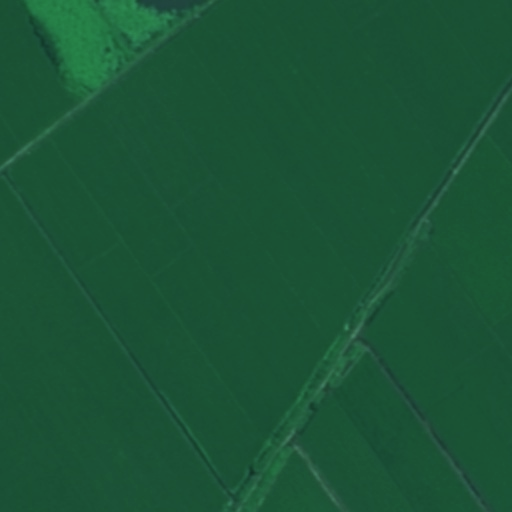} &
        \includegraphics[width=1.8cm]{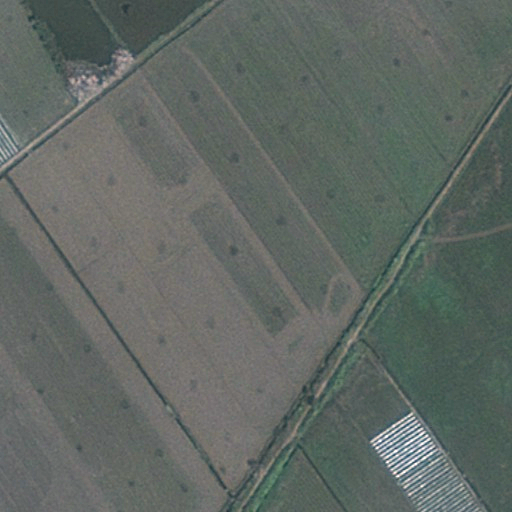} &
        \includegraphics[width=1.8cm]{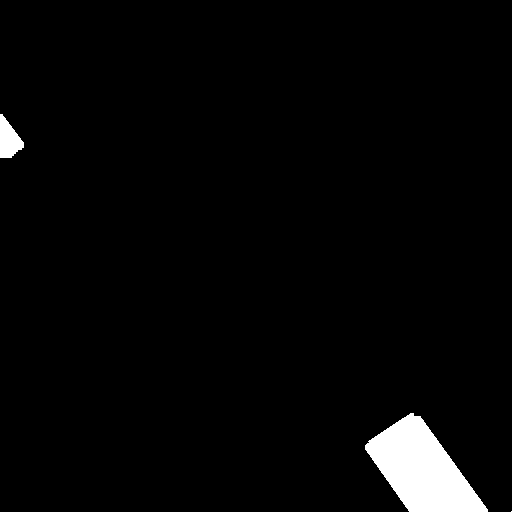} &
        \includegraphics[width=1.8cm]{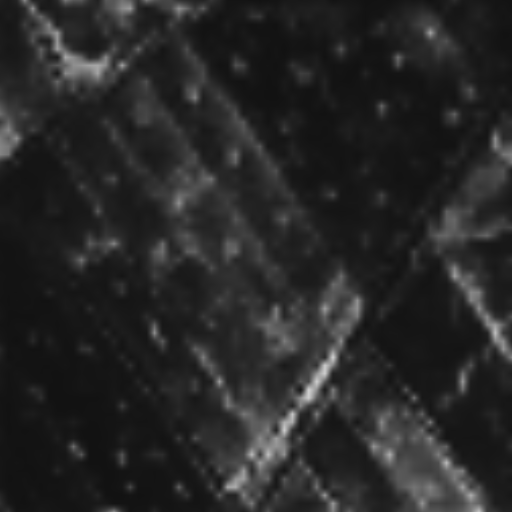} &
        \includegraphics[width=1.8cm]{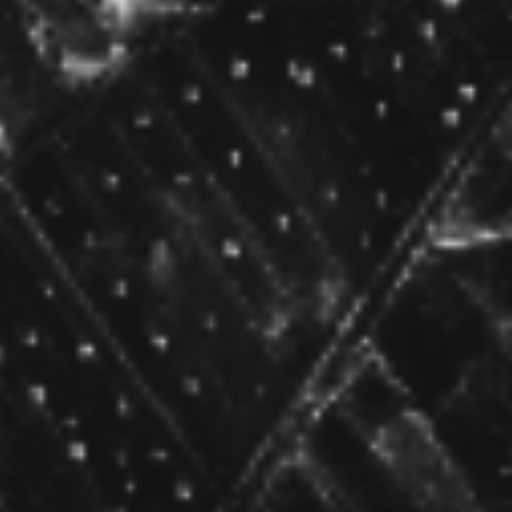} &
        \includegraphics[width=1.8cm]{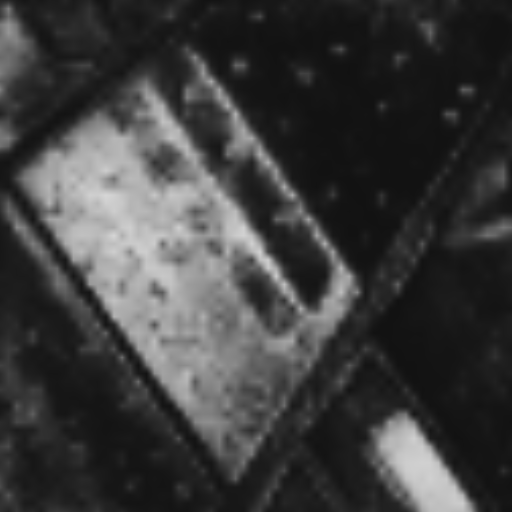} &
        \includegraphics[width=1.8cm]{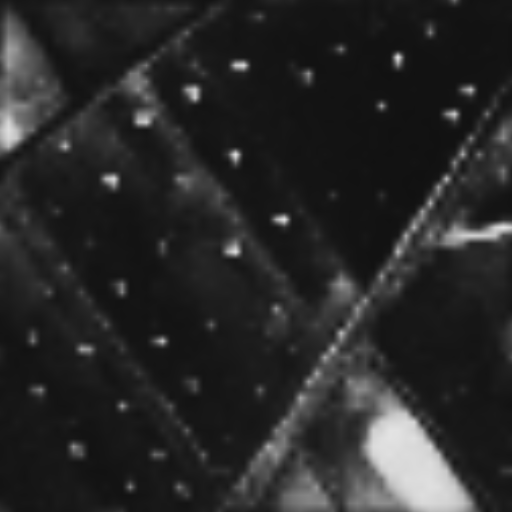}&
        \includegraphics[width=1.8cm]{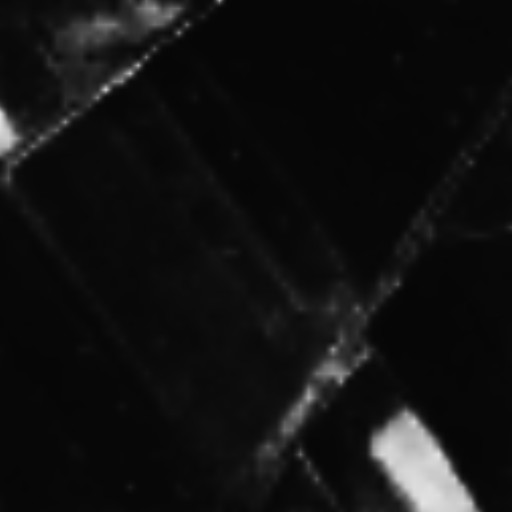}&
        \includegraphics[width=1.8cm]{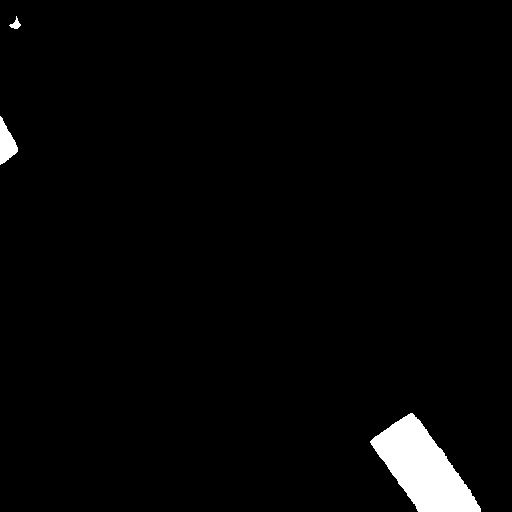}\\
        (b)&
        \includegraphics[width=1.8cm]{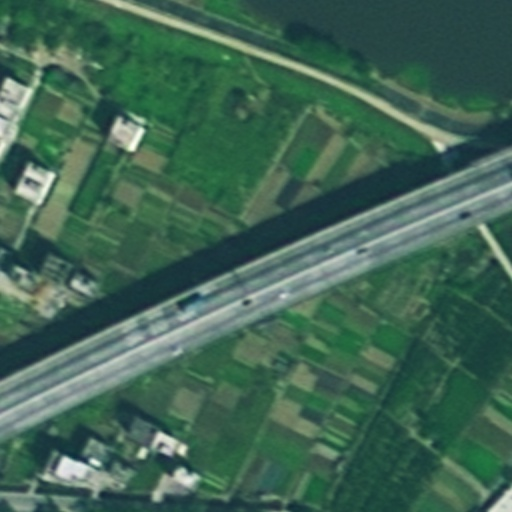} &
        \includegraphics[width=1.8cm]{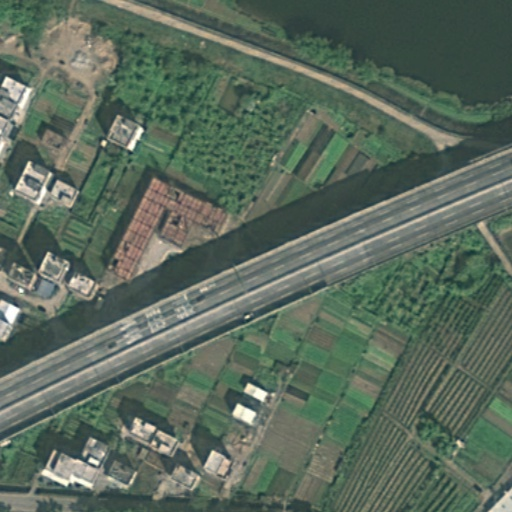} &
        \includegraphics[width=1.8cm]{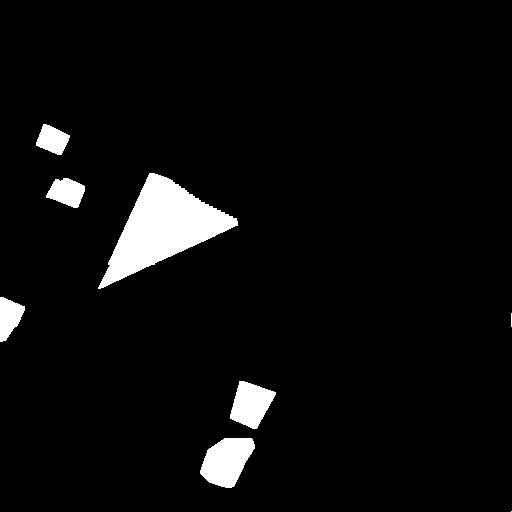} &
        \includegraphics[width=1.8cm]{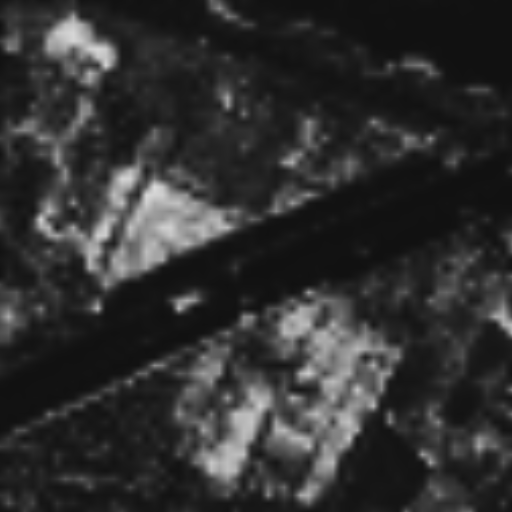} &
        \includegraphics[width=1.8cm]{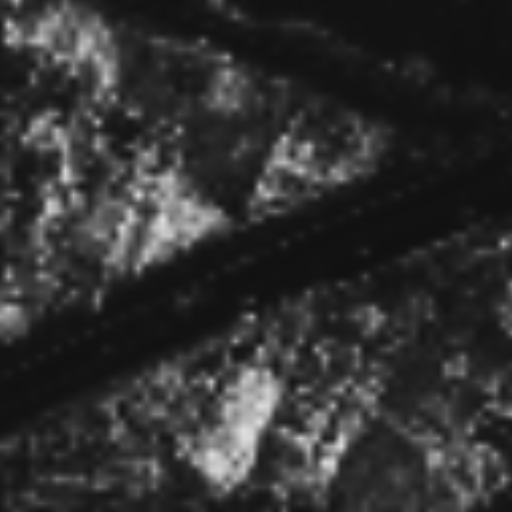} &
        \includegraphics[width=1.8cm]{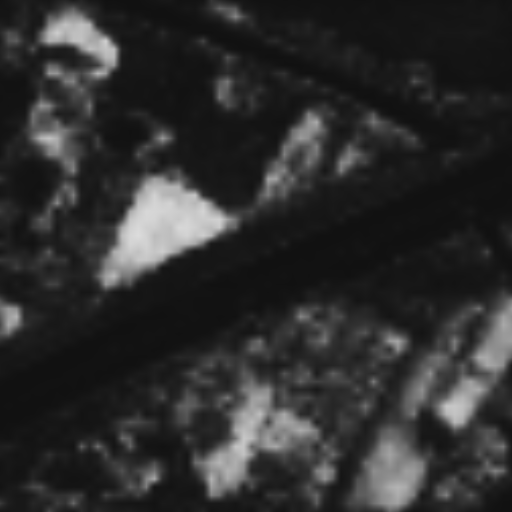} &
        \includegraphics[width=1.8cm]{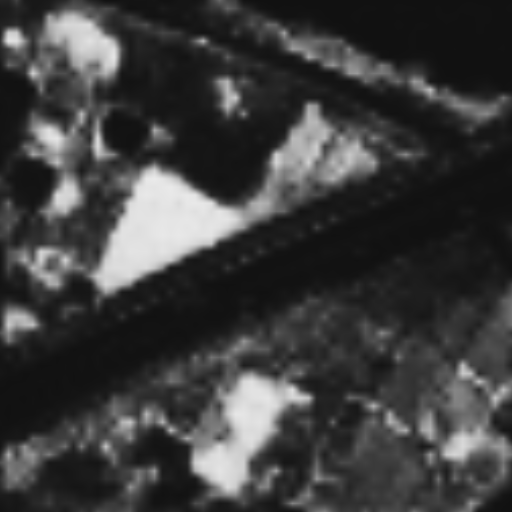}&
        \includegraphics[width=1.8cm]{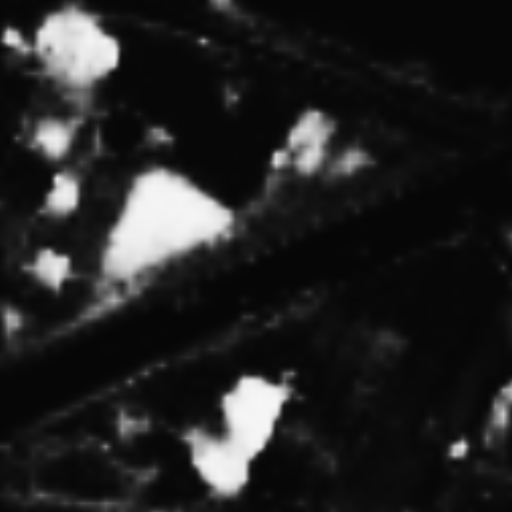}&
        \includegraphics[width=1.8cm]{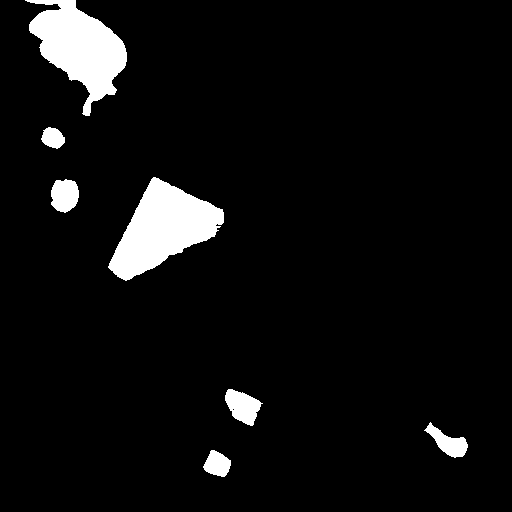}\\
        (c)&
        \includegraphics[width=1.8cm]{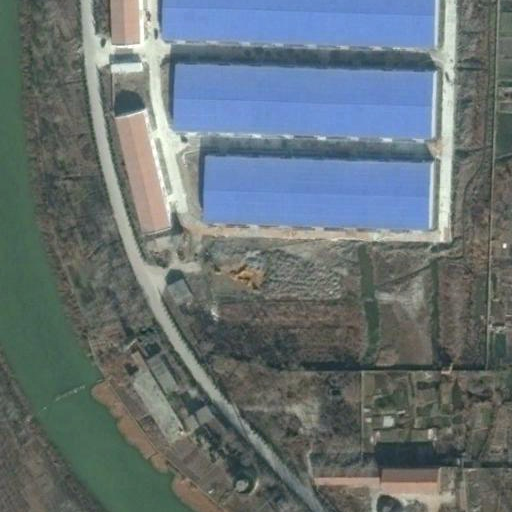} &
        \includegraphics[width=1.8cm]{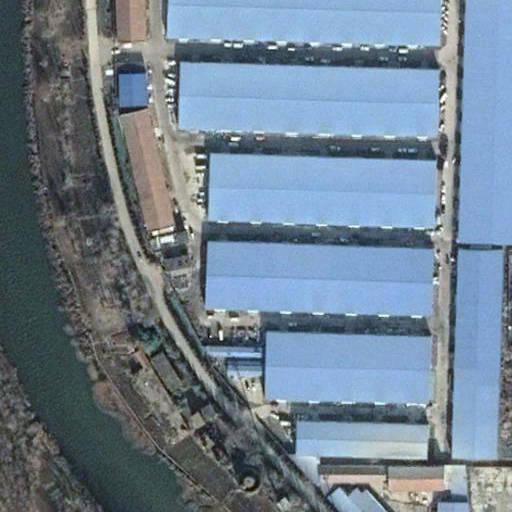} &
        \includegraphics[width=1.8cm]{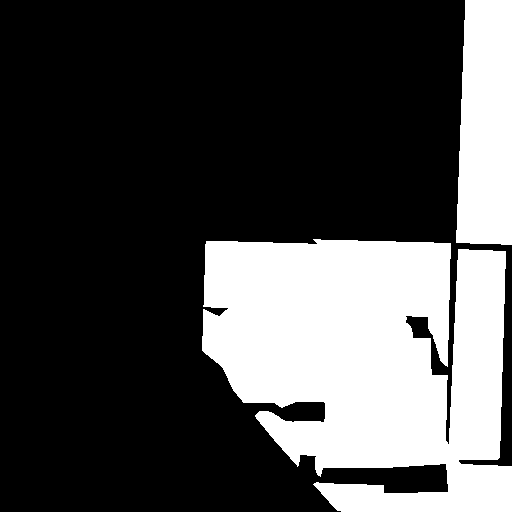} &
        \includegraphics[width=1.8cm]{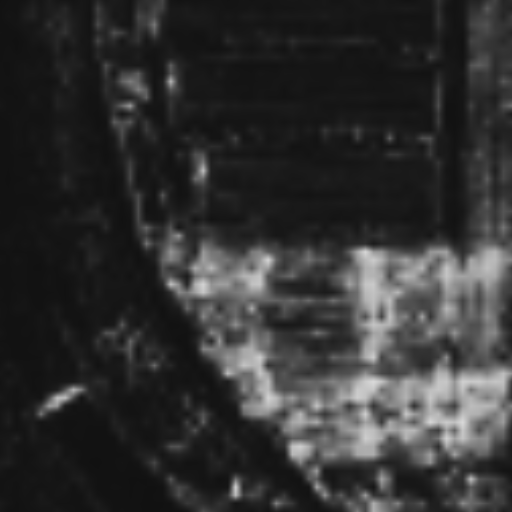} &
        \includegraphics[width=1.8cm]{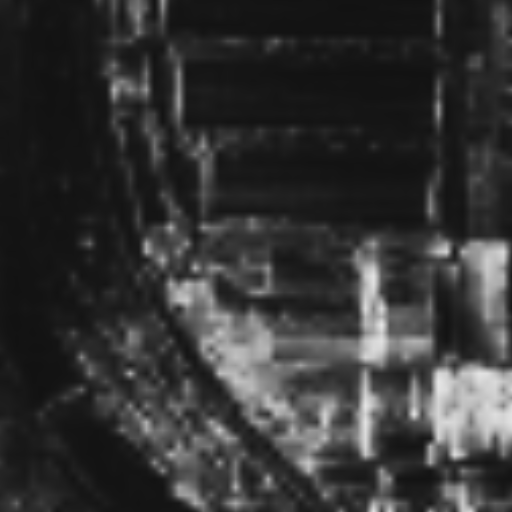} &
        \includegraphics[width=1.8cm]{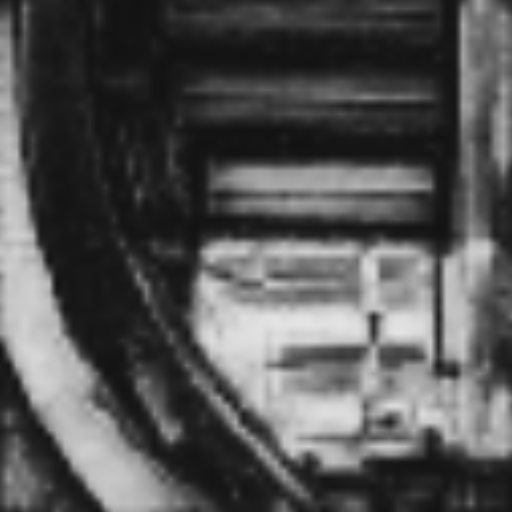} &
        \includegraphics[width=1.8cm]{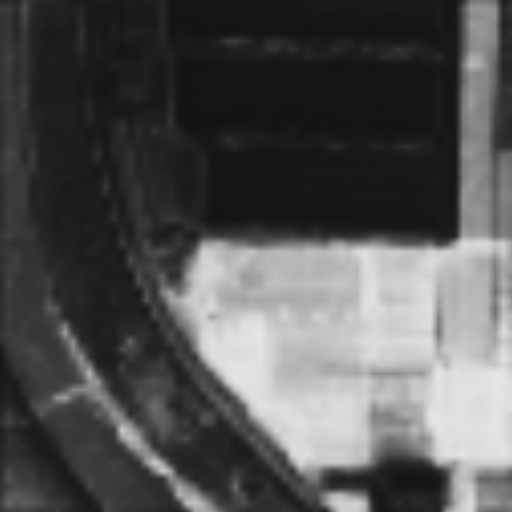}&
        \includegraphics[width=1.8cm]{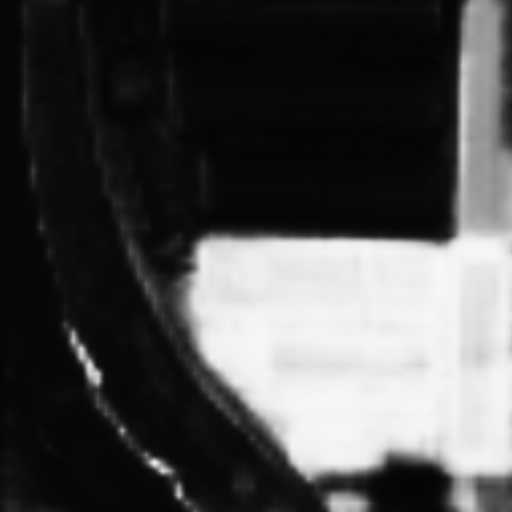}&
        \includegraphics[width=1.8cm]{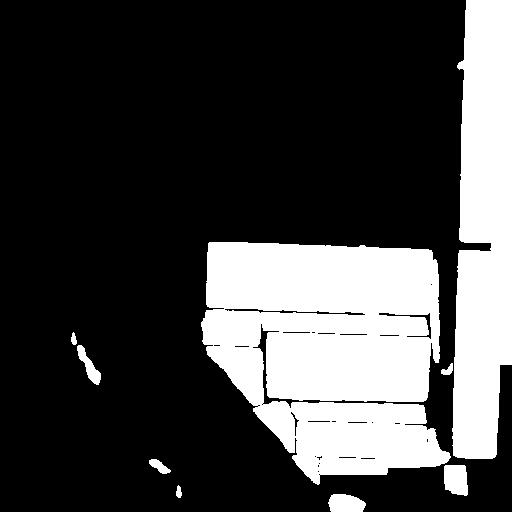}\\
        (d)&
        \includegraphics[width=1.8cm]{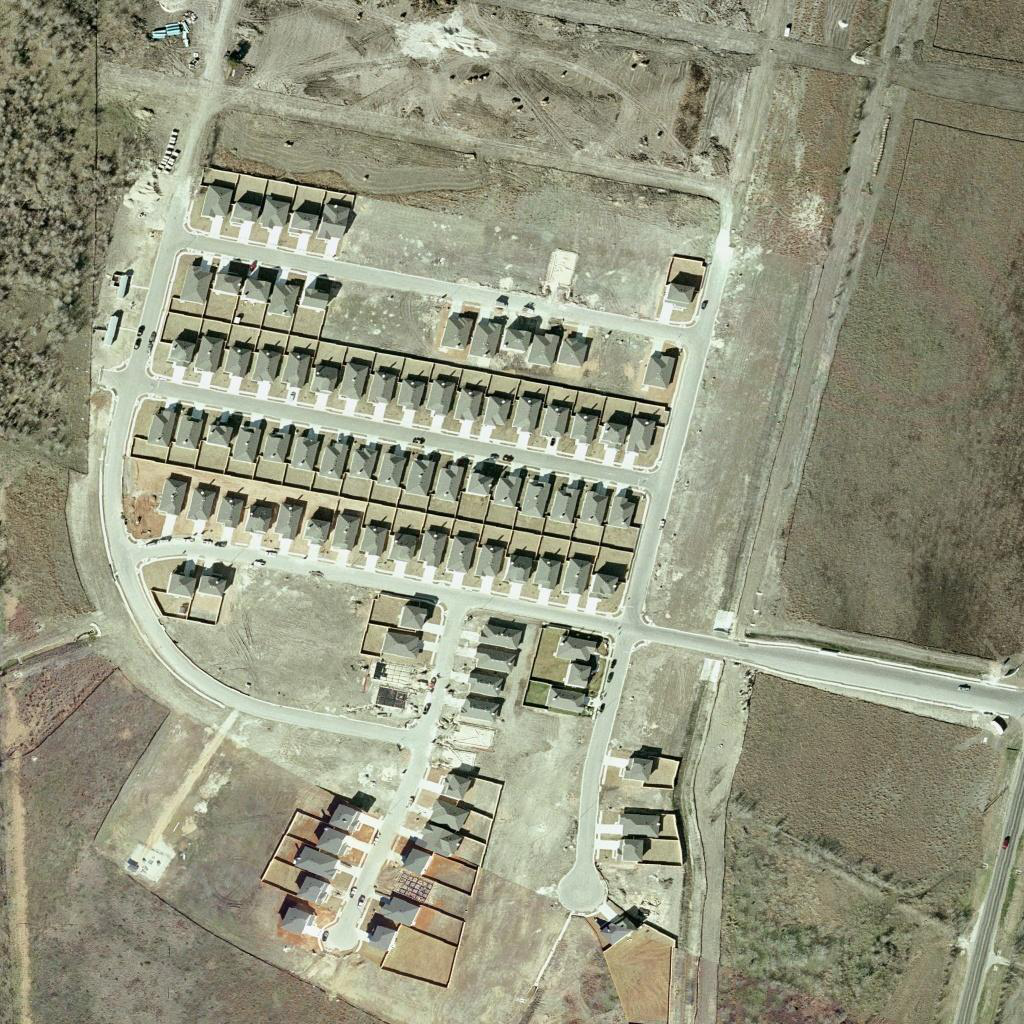} &
        \includegraphics[width=1.8cm]{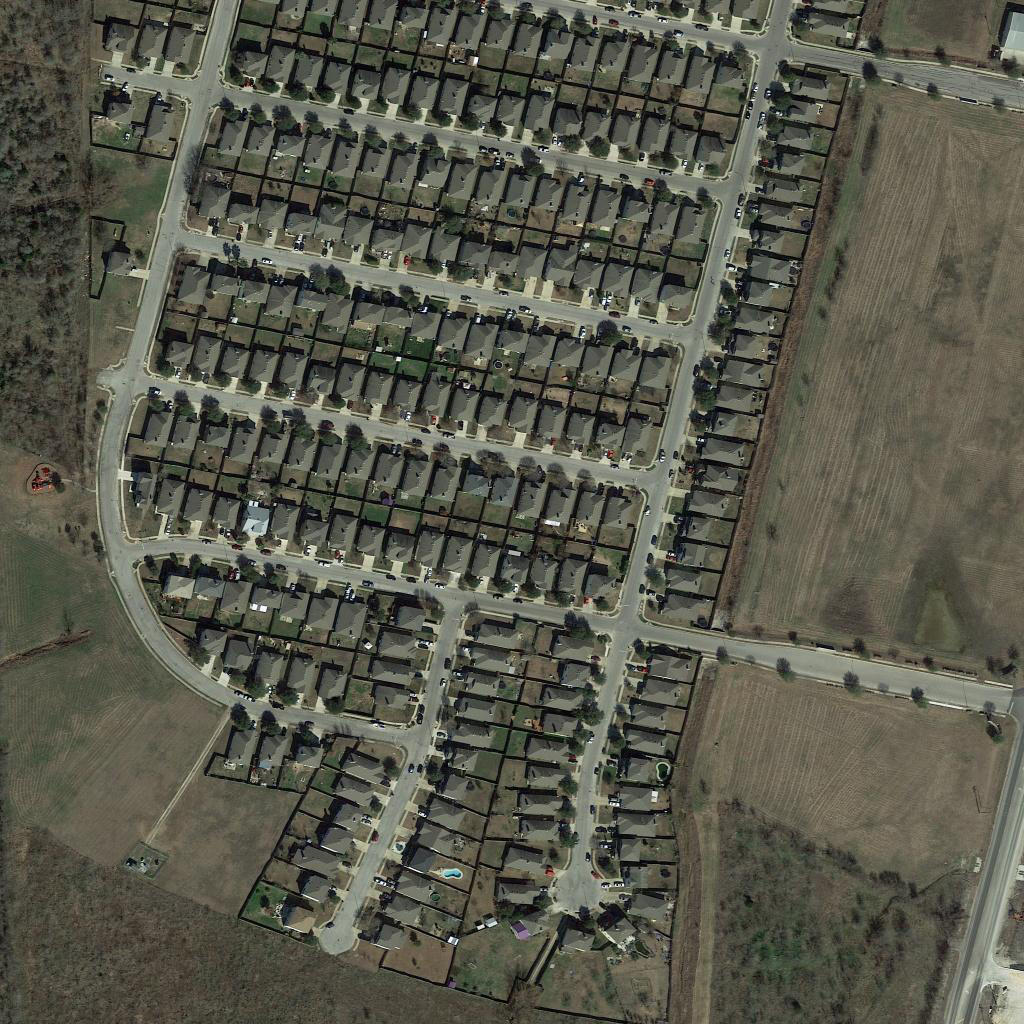} &
        \includegraphics[width=1.8cm]{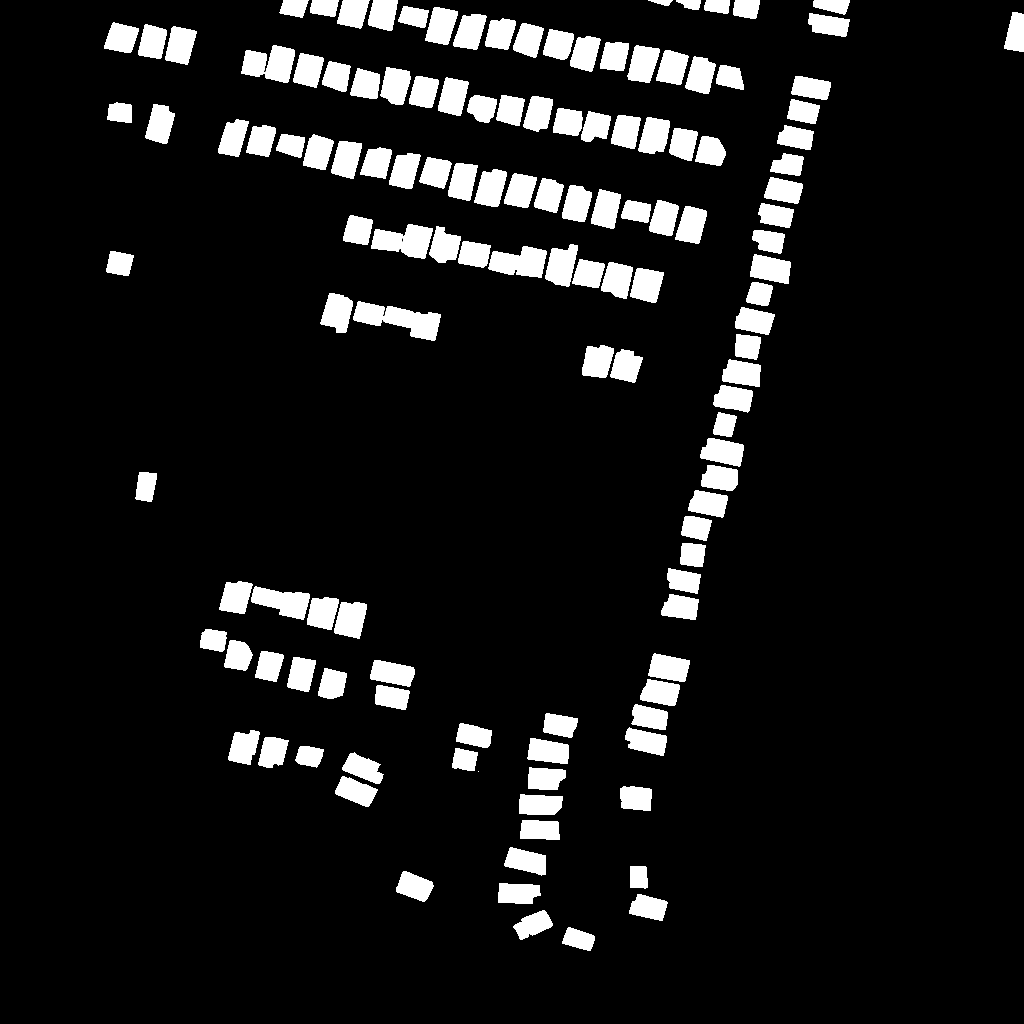} &
        \includegraphics[width=1.8cm]{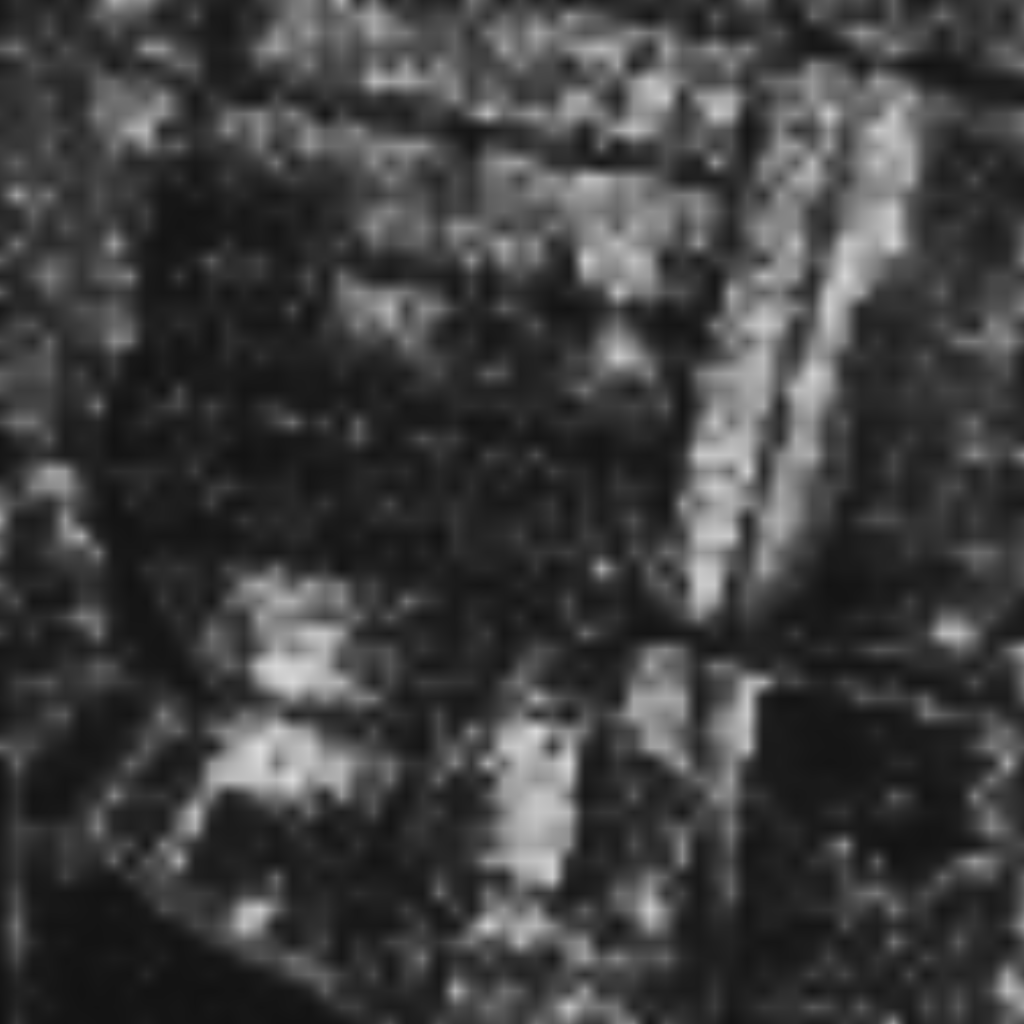} &
        \includegraphics[width=1.8cm]{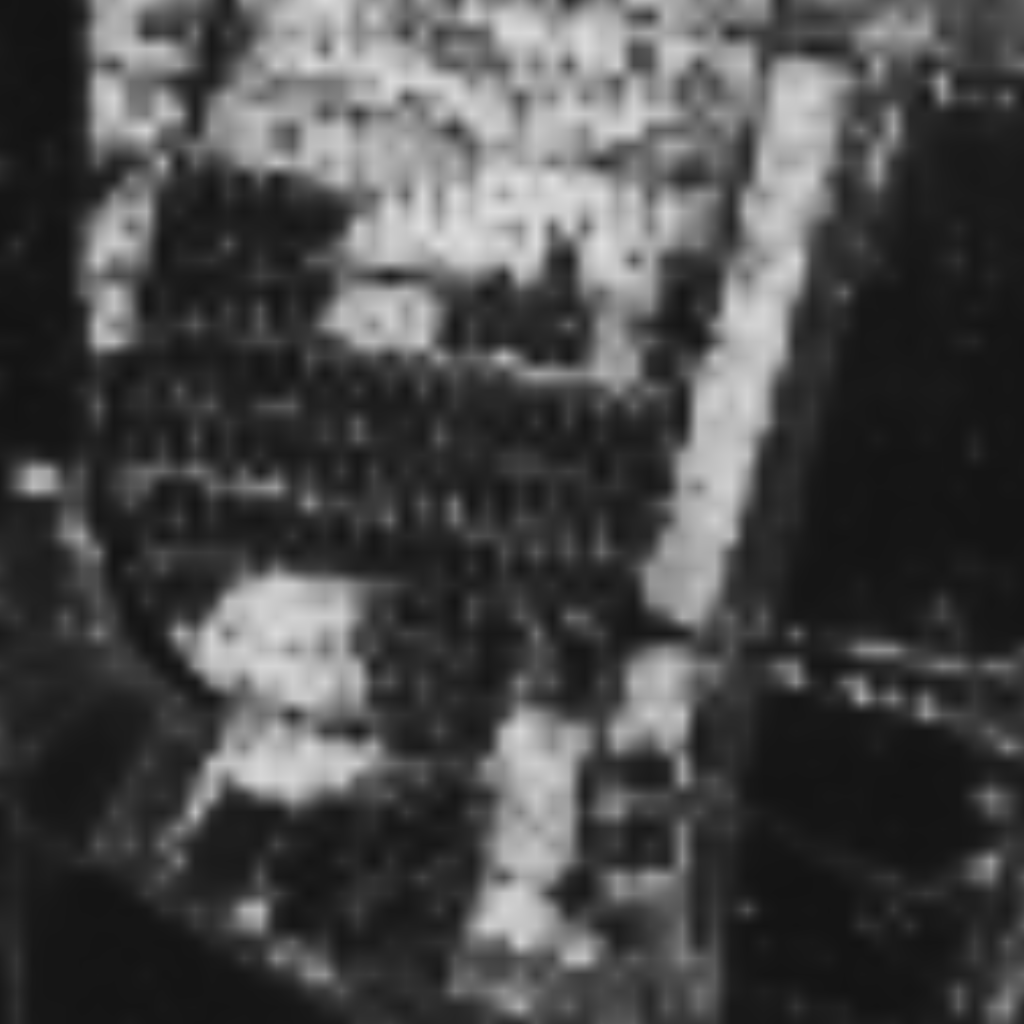} &
        \includegraphics[width=1.8cm]{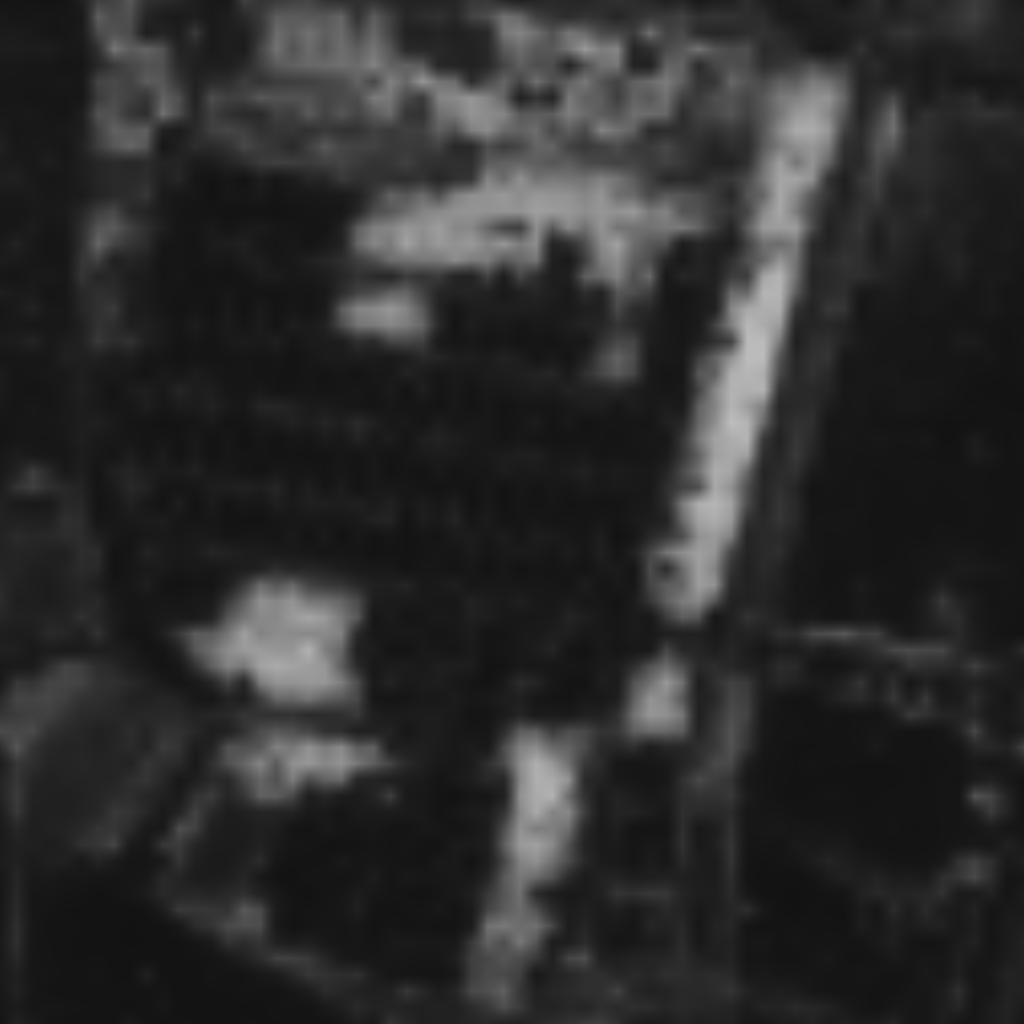} &
        \includegraphics[width=1.8cm]{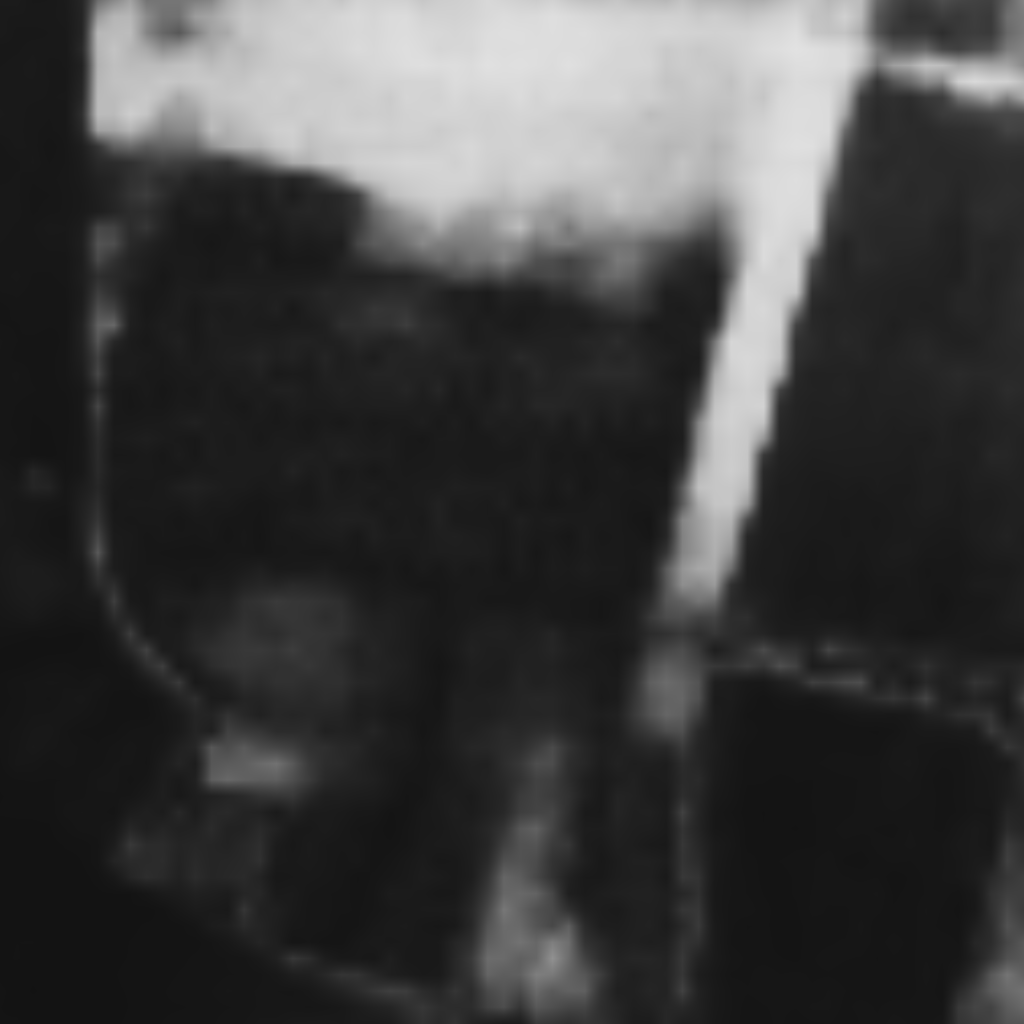}&
        \includegraphics[width=1.8cm]{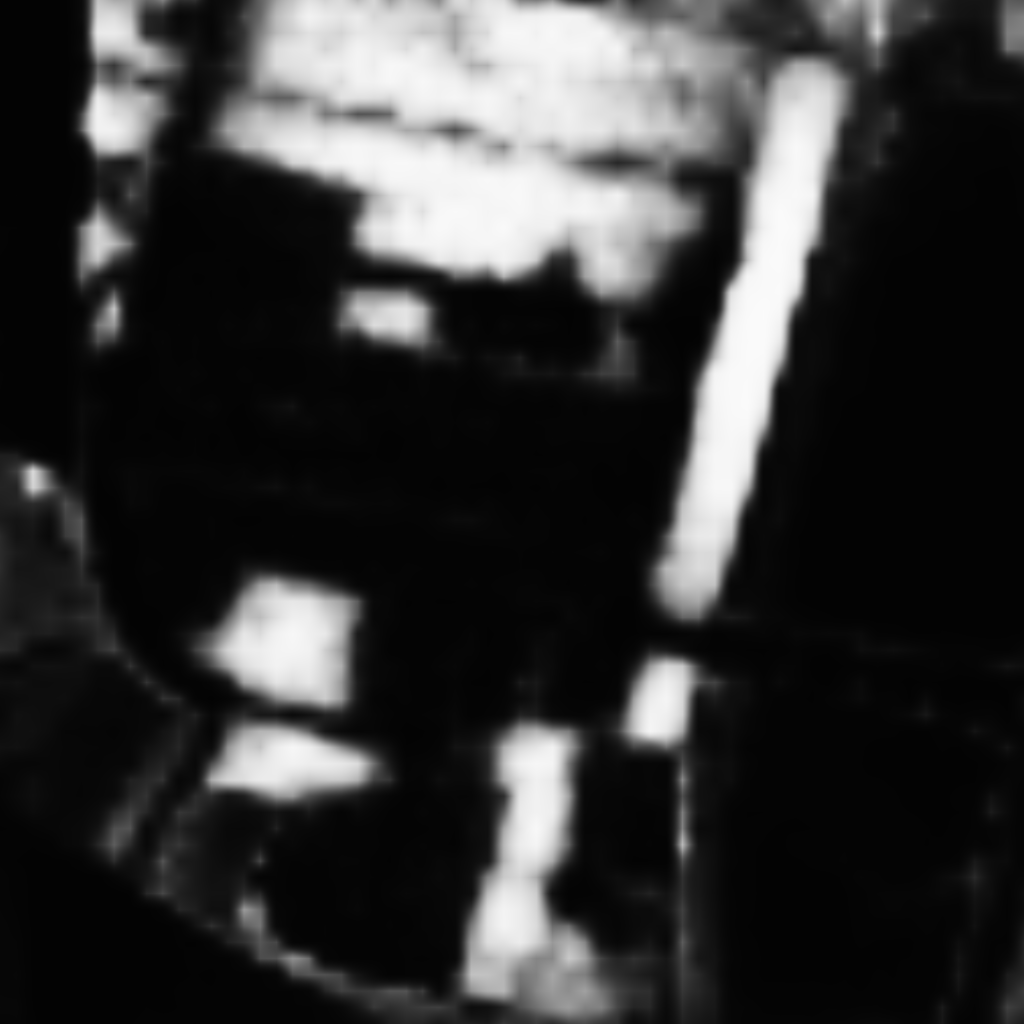}&
        \includegraphics[width=1.8cm]{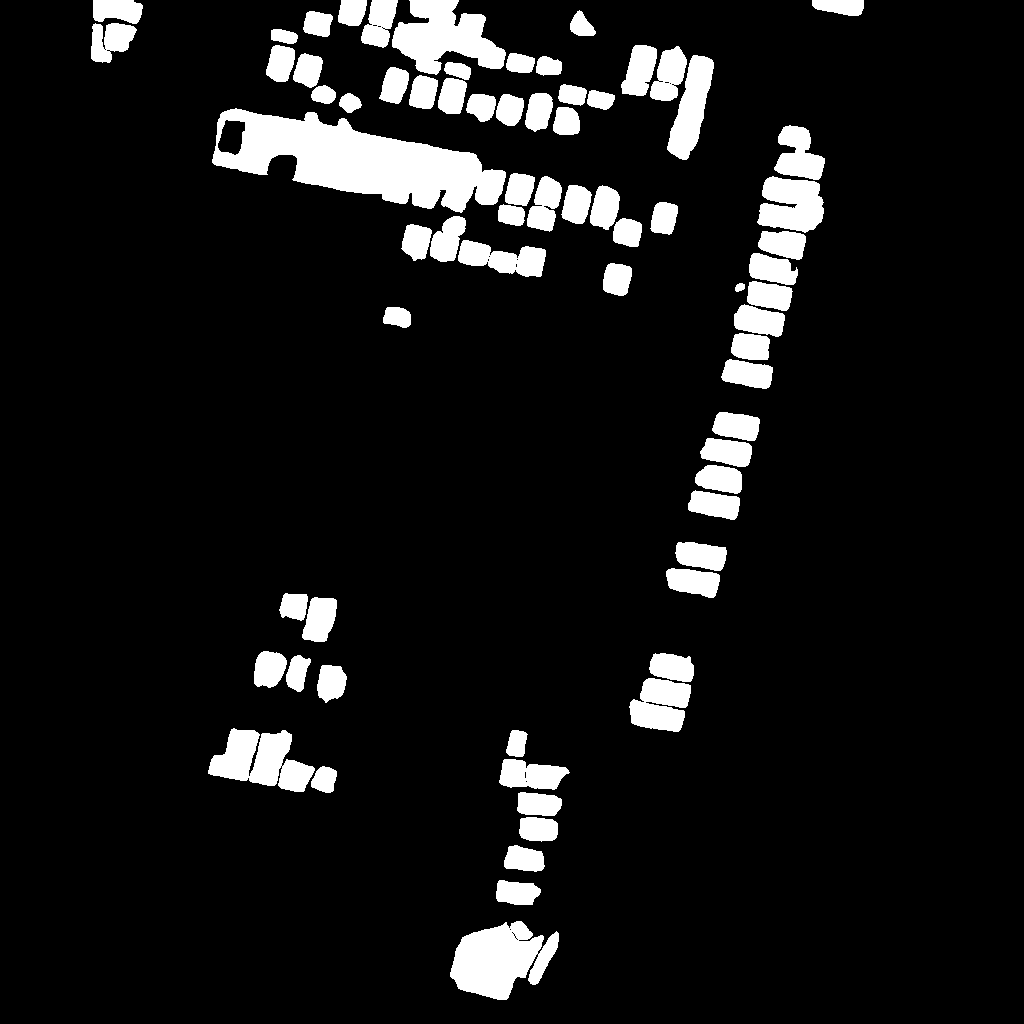}\\
        & $T_1$ image & $T_2$ image & GT & SC (baseline) & S2C {\small(CSC only)} & S2C {\small(CTC only)} & S2C {\small(CSC+CTC)} & S2C & S2C {\small(Refined)}\\
    \end{tabular}
    \caption{Example of results from different methods in the ablation study. CTC highlights differences while introducing much noise, CSC enhances change representations and reduces certain false alarms, while $\mathcal{L}_{spa}$ further suppresses insignificant changes.}
    \label{Fig.ablation_vis}
\end{figure*}

\textbf{Parameter Analysis.} In Eq.\ref{eq.losses} two weighting parameters $\alpha$ and $\beta$ are introduced to balance the different training objectives. The optimal weighting parameters are determined by a sensitivity analysis. Considering that $\mathcal{L}_{info}$ poses considerable optimization challenges and commonly exhibits high values, $\alpha$ is assigned small values within the range of $[0, 1]$. In contrast, $\mathcal{L}_{spa}$ is straightforward to optimize and has relatively low values, thus $\beta$ is assigned with values over 1. Fig.\ref{fig.param_acc} reports the accuracy in $F_1$ obtained with different values of $\alpha$ and $\beta$. To mitigate random variability, the reported accuracy is the mean value across three trials. The results show a strong correlation between $\alpha$ and accuracy, whereas the effect of $\beta$ appears inconsistent. This finding aligns with the underlying learning mechanism within S2C, where $\alpha$ balances between $\mathcal{L}_{tri}$ and $\mathcal{L}_{info}$, affecting the focus on mapping differences or similarities. The allocation of $\alpha =0.2, \beta =2$ yields the highest observed accuracy.

\begin{figure}[t]
\centering
    \includegraphics[width=1\linewidth]{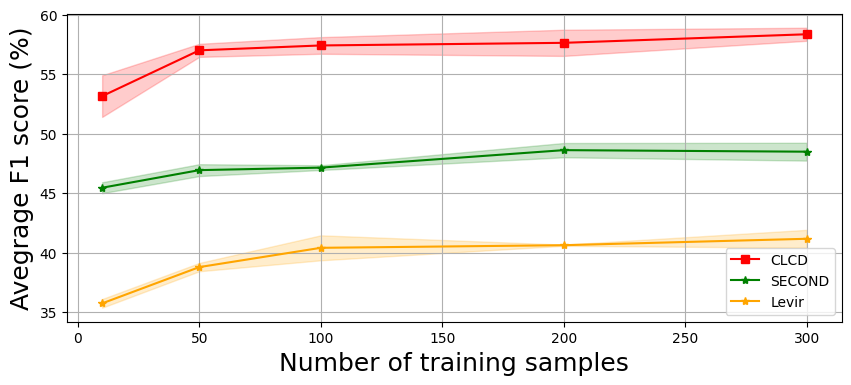}
    \caption{Average $F_1$ (\%) of S2C versus sample volume.}
\label{fig.acc_curve}
\end{figure}

\begin{table}[t]
    \centering
    \caption{Average $F_1$ (\%) of S2C obtained with varying amount of unlabeled samples.}
    \resizebox{1\linewidth}{!}{%
        \begin{tabular}{r|cccccc}
        \toprule
            \multirow{2}*{Dataset} & \multicolumn{6}{c}{Number of samples}\\
            \cline{2-7}
            & 5 & 10 & 50 & 100 & 200 & 300 \\
            \hline
            CLCD & 53.39\textcolor{gray}{$\pm$4.9} & 53.16\textcolor{gray}{$\pm$3.5} & 57.01\textcolor{gray}{$\pm$1.2} & 57.42\textcolor{gray}{$\pm$1.4} & 57.64\textcolor{gray}{$\pm$2.2} & 58.36\textcolor{gray}{$\pm$1.1} \\
            SECOND & 47.15\textcolor{gray}{$\pm$3.5} & 45.46\textcolor{gray}{$\pm$0.9} & 46.94\textcolor{gray}{$\pm$1.0} & 47.15\textcolor{gray}{$\pm$0.4} & 48.62\textcolor{gray}{$\pm$1.2} & 48.49\textcolor{gray}{$\pm$1.5} \\
            Levir & 35.98\textcolor{gray}{$\pm$3.1} & 35.74\textcolor{gray}{$\pm$0.7} & 38.79\textcolor{gray}{$\pm$0.7} & 40.41\textcolor{gray}{$\pm$2.1} & 40.63\textcolor{gray}{$\pm$0.1} & 41.17\textcolor{gray}{$\pm$1.5} \\
        \bottomrule
        \end{tabular}
        }\label{Table.FewSample}
\end{table}

\textbf{Qualitative Assessment.} Fig.\ref{Fig.ablation_vis} presents some examples of the CD results obtained using the different techniques. These results are selected in the 3 datasets covering different scenes, including (a) cropland, (b) countryside (c) factories, and (d) residential blocks. One can observe that using either the CSC or CTC alone leads to much noise, while their collaborative employment greatly reduces the false alarms. The grid sparsity generalization further removes much insignificant change representation and leads to more smoothed CD results. After the post-processing of \textit{IoU refinement}, binary CD results matching the object boundaries are produced.

\textbf{Training S2C with Few Samples.} An additional advantage of incorporating consistency regularization in contrastive learning is the enhancement of generalization ~\cite{yang2023revisiting, bandara2022revisiting}. Throughout the experimental process, we observe that the accuracy of the S2C framework is not sensitive to the number of training samples (unlabeled). Consequently, additional experiments are conducted to investigate the requisite quantity of unlabeled image samples for effective training of the S2C framework.

We train the S2C model using varying amounts of unlabeled data, ranging from 300 to as few as 5 and subsequently evaluate the resulting accuracy on the entire test set. To increase the diversity of samples, additional random cropping and enlarging augmentations are added in the preprocessing operations. This is added into the \textit{weak augmentation} process as elaborated in Sec.\ref{sc4.implement}. This operation is crucial for deploying the S2C with very few samples: when there are insufficient samples to enable spatial comparisons within the CSC paradigm, cropped sub-regions can be alternatives. To mitigate variations due to sample selection, the experiments are conducted over three trials, each randomly selecting the specified sample volume.

The average $F_1 (\%)$ results of the experimental trials are reported in Table \ref{Table.FewSample}. Compared with the results obtained using 300 training samples, utilizing only 100 samples yields an almost negligible average $F_1$ decrease of 1\% across the three datasets. Employing a minimal set of merely 5 samples results in an average $F_1$ decrease of about 3.8\%. This low requirement on training samples suggests that the S2C framework can be easily deployed in real-world applications. To intuitively illustrate the results, a graph of accuracy versus sample volume is presented in Fig.\ref{fig.acc_curve}. By analyzing the figure, one can infer that on the considered datasets, employing a minimum of 5 samples is sufficient for the effective training of the S2C, while utilizing 100 samples is adequate for attaining near-optimal accuracy.

\subsection{Comparative Experiments}

\begin{table*}[t]
    \centering
    \caption{Quantitative accuracy (\%) evaluation of the proposed S2C and SOTA UCD methods on various benchmark datasets.}
    \resizebox{1\linewidth}{!}{%
        \begin{tabular}{r|c|cccc|cccc|cccc}
        \toprule
            \multirow{2}*{Methods} & \multirow{2}*{Publication} & \multicolumn{4}{c|}{CLCD} & \multicolumn{4}{c|}{SECOND} & \multicolumn{4}{c}{Levir} \\
            \cline{3-14}
            & & $OA$ & $Pre$ & $Rec$ & $F_1$ & $OA$ & $Pre$ & $Rec$ & $F_1$ & $OA$ & $Pre$ & $Rec$ & $F_1$ \\
            \hline
            \color{gray} SAM-CD &\color{gray} \textit{TGRS 2024} \cite{ding2024samcd} & \color{gray}96.26 & \color{gray}73.01 & \color{gray}78.84 & \color{gray}75.81 & \color{gray}88.56 & \color{gray}73.32 & \color{gray}66.00 & \color{gray}69.47 & \color{gray}99.17 & \color{gray}92.62 & \color{gray}91.04 & \color{gray}91.82\\
            \hline
            CVA & \textit{TGRS 2000} \cite{Bruzzone2000diff} & 71.01 & 8.49 & 29.62 & 13.20 & 59.17 & 20.55 & 37.34 & 26.51 & 66.50 & 5.80 & 36.59 & 10.02 \\
            ISFA & \textit{TGRS 2014} \cite{Wu2014Slow} & 74.37 & 8.60 & 25.39 & 12.85 & 60.23 & 20.13 & 34.24 & 25.35 & 69.32 & 6.03 & 34.45 & 10.27\\
            DCVA & \textit{TGRS 2019} \cite{saha2019unsupervised} & 53.91 & 11.35 & \textbf{76.26} & 19.76 & 55.51 & 25.63 & \underline{66.07} & 36.93 & 48.28 & 7.20 & 76.94 & 13.16\\
            DSFA & \textit{TGRS 2019} \cite{Du2019Unsupervised} & 52.09 & 8.53 & 55.94 & 14.80 & 48.10 & 19.06 & 50.28 & 27.65 & 60.29 & 5.99 & 46.22 & 10.60\\
            KPCA & \textit{TCYB 2022} \cite{wu2021unsupervised} & 53.47 & 13.84 & 44.52 & 21.11 & 54.69 & 20.44 &  44.87 & 28.09 & 54.97 & 5.61 & 49.52 & 10.08\\
            CDRL & \textit{CVPR 2022} \cite{noh2022unsupervised} & 64.74 & 7.75 & 34.27 & 12.64 & 62.65 & 37.78 & 22.90 & 28.51 & 62.59 & 5.32 & 37.74 & 9.32\\
            SiROC & \textit{TGRS 2022} \cite{Kondmann2022Spatial} & 82.38 & 18.93 & 41.65 & 26.03 & 69.80 & 27.92 & 33.59 & 30.49 & 64.82 & 7.38 & 51.14 & 12.90 \\
            I3PE & \textit{ISPRS 2023} \cite{chen2023exchange} & 91.12 & 32.42 & 17.79 & 22.97 & 74.09 & 34.53 & 35.03 & 34.78 & \underline{91.03} & 17.39 & 20.31 & 18.74\\
            FCD-GAN & \textit{TPAMI 2023} \cite{wu2023fully} & 83.93 & 22.06 & 45.79 & 29.77 & 68.36 & 29.47 & 43.41 & 35.11 & 83.42 & 8.87 & 24.29 & 12.99 \\
            AnyChange & \textit{NeurIPS 2024} \cite{zheng2024segment} & - & - & - & - & - & 30.5 & \textbf{83.2} & 44.6 & - & 13.3 & \textbf{85.0} & 23.0 \\
            \hline    
            \multicolumn{2}{r|}{S2C-Dinov2 (proposed)} & \underline{93.95} & \underline{59.04} & \underline{61.24} & \underline{60.12} & \textbf{84.55} & \textbf{67.15} & 42.41 & \underline{51.99} & 89.28 & \underline{29.42} & \underline{78.88} & \underline{42.86} \\ 
            \multicolumn{2}{r|}{S2C + \textit{IoU Refine} (proposed)} & \textbf{94.46} & \textbf{63.82} & 59.12 & \textbf{61.38} & \underline{84.45} & \underline{64.91} & 46.02 & \textbf{53.86} & \textbf{92.84} & \textbf{34.85} & 70.69 & \textbf{46.69} \\ 
        \bottomrule
        \end{tabular}
        }\label{Table.CompareSOTA}
\end{table*}

\begin{figure*}[t]
\centering
    \setlength{\tabcolsep}{1pt}
    \begin{tabular}{>{\centering\arraybackslash}m{0.4cm}>{\centering\arraybackslash}m{1.5cm}>{\centering\arraybackslash}m{1.5cm}>{\centering\arraybackslash}m{1.5cm}>{\centering\arraybackslash}m{1.5cm}>{\centering\arraybackslash}m{1.5cm}>{\centering\arraybackslash}m{1.5cm}>{\centering\arraybackslash}m{1.5cm}>{\centering\arraybackslash}m{1.5cm}>{\centering\arraybackslash}m{1.5cm}>{\centering\arraybackslash}m{1.5cm}>{\centering\arraybackslash}m{1.5cm}}
        & & & & \multicolumn{7}{c}{\includegraphics[width=10cm]{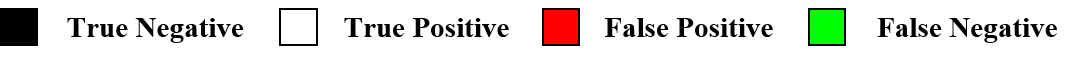}}\\
        (a)&
        \includegraphics[width=1.5cm]{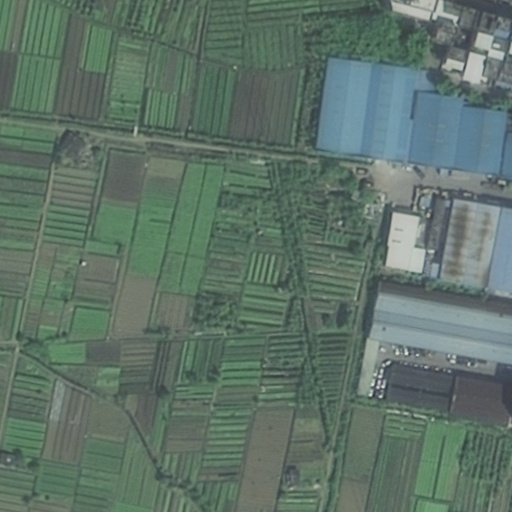} &
        \includegraphics[width=1.5cm]{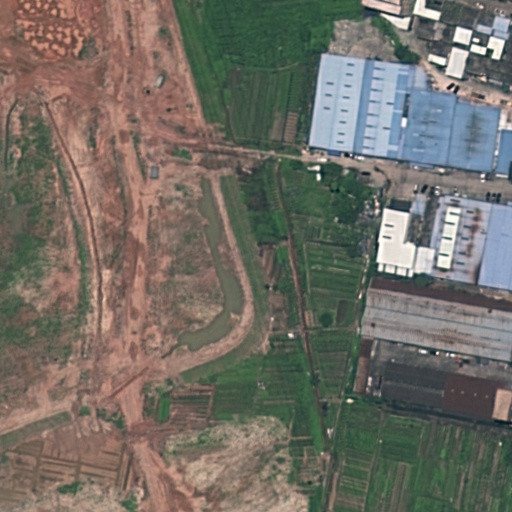} &
        \includegraphics[width=1.5cm]{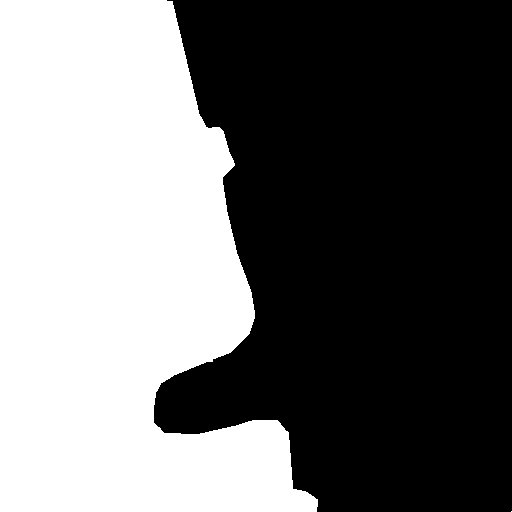} &
        \includegraphics[width=1.5cm]{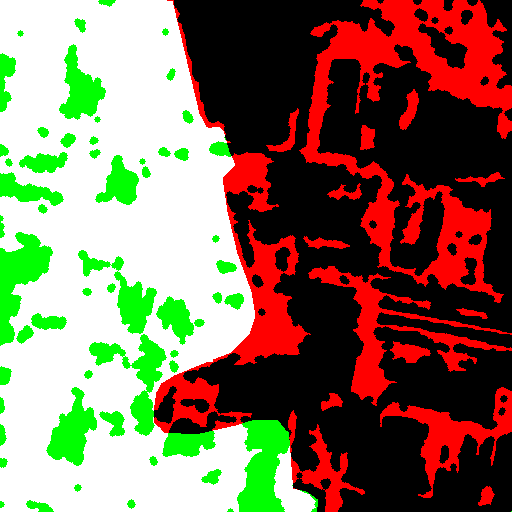} &
        \includegraphics[width=1.5cm]{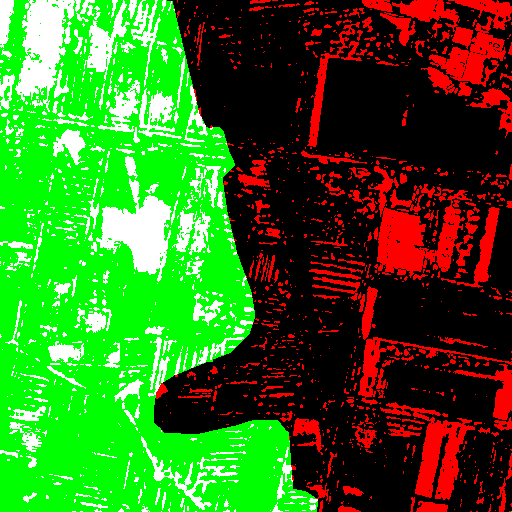} &
        \includegraphics[width=1.5cm]{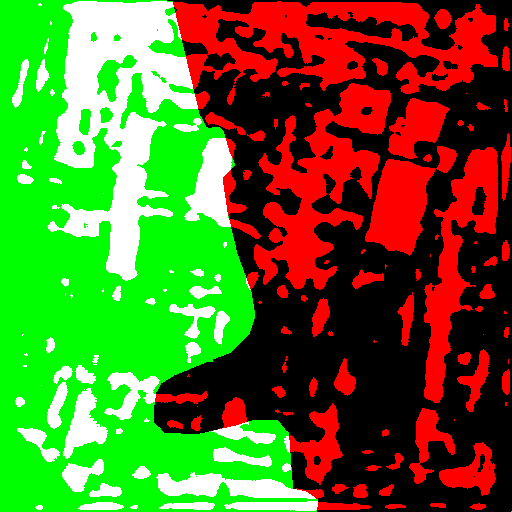} &
        \includegraphics[width=1.5cm]{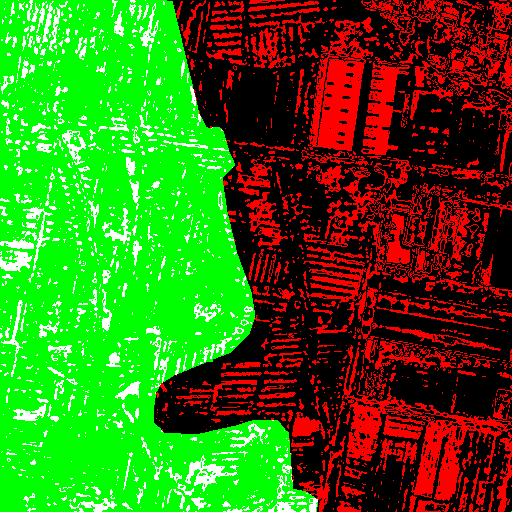} &
        \includegraphics[width=1.5cm]{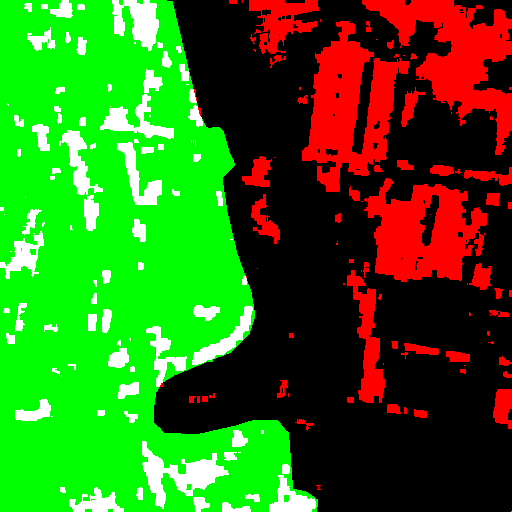}&
        \includegraphics[width=1.5cm]{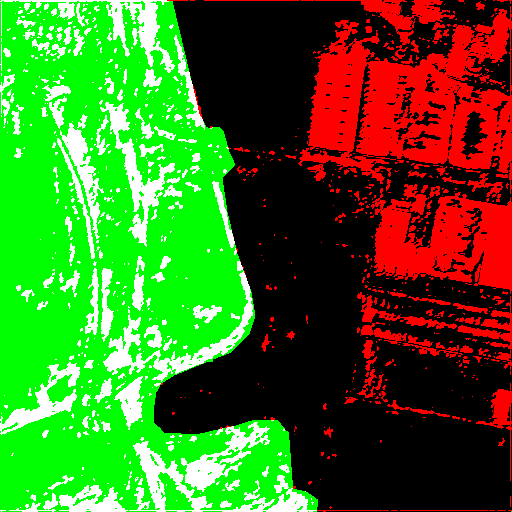}&
        \includegraphics[width=1.5cm]{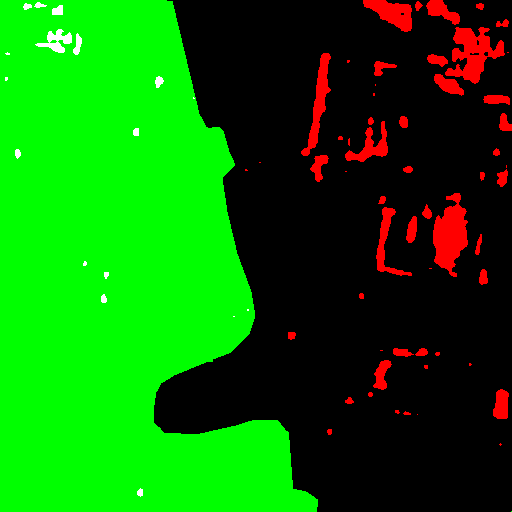}&
        \includegraphics[width=1.5cm]{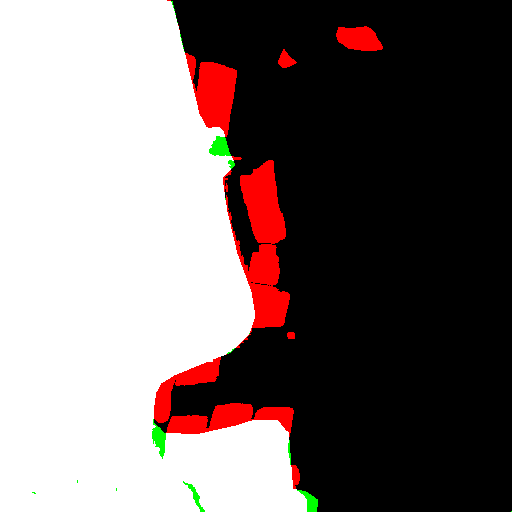}\\
        (b)&
        \includegraphics[width=1.5cm]{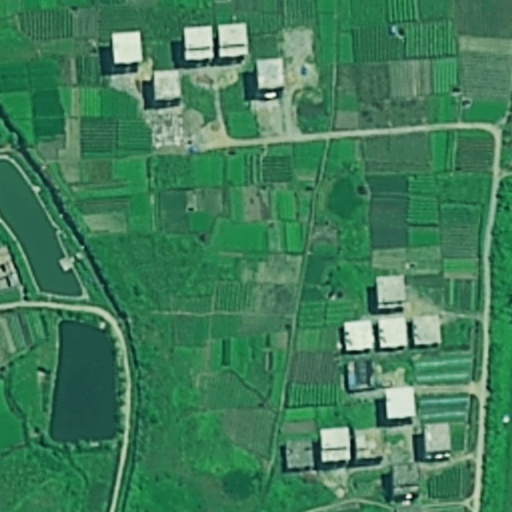} &
        \includegraphics[width=1.5cm]{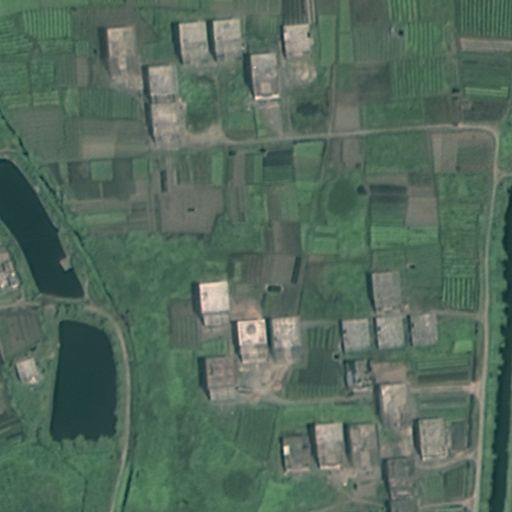} &
        \includegraphics[width=1.5cm]{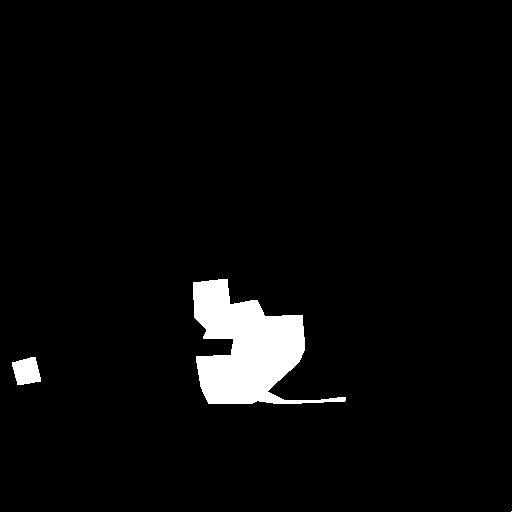} &
        \includegraphics[width=1.5cm]{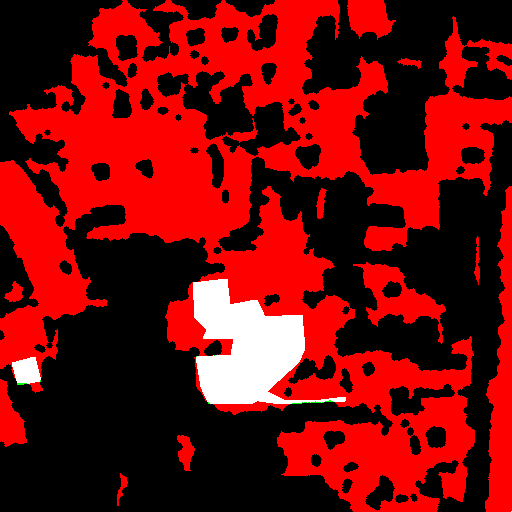} &
        \includegraphics[width=1.5cm]{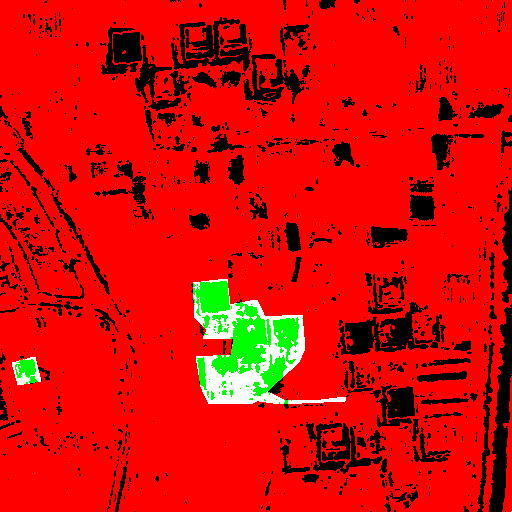} &
        \includegraphics[width=1.5cm]{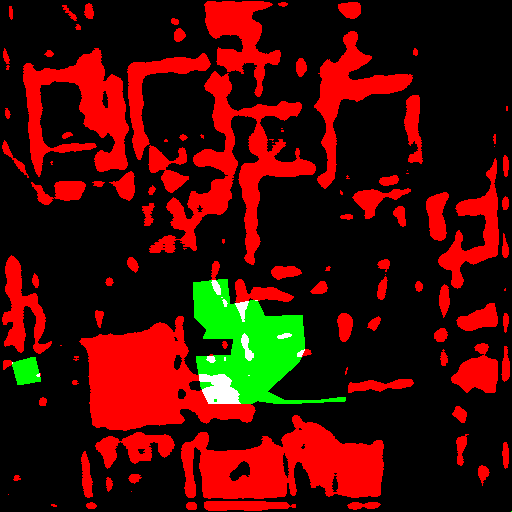} &
        \includegraphics[width=1.5cm]{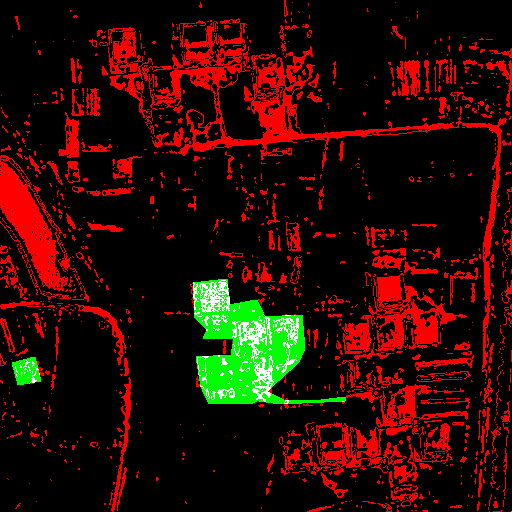} &
        \includegraphics[width=1.5cm]{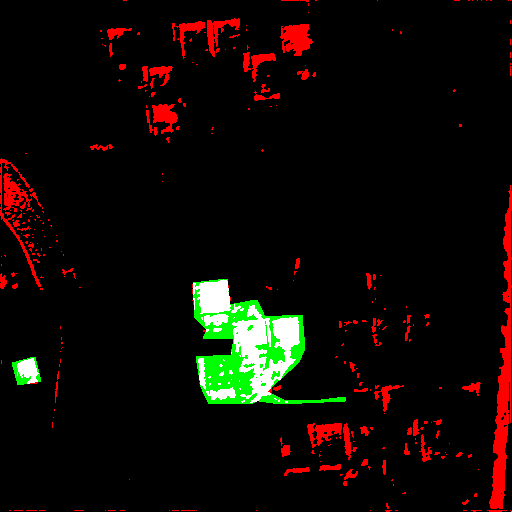}&
        \includegraphics[width=1.5cm]{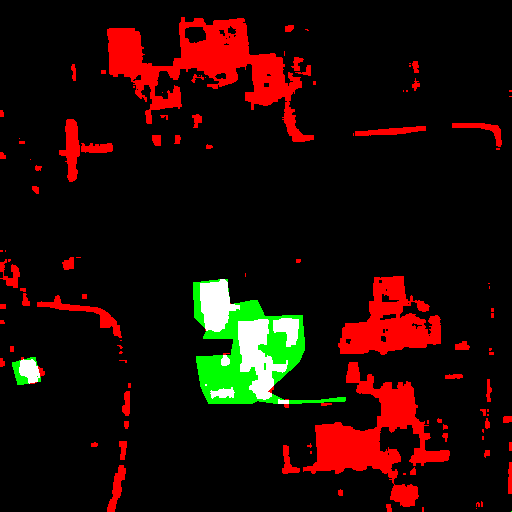}&
        \includegraphics[width=1.5cm]{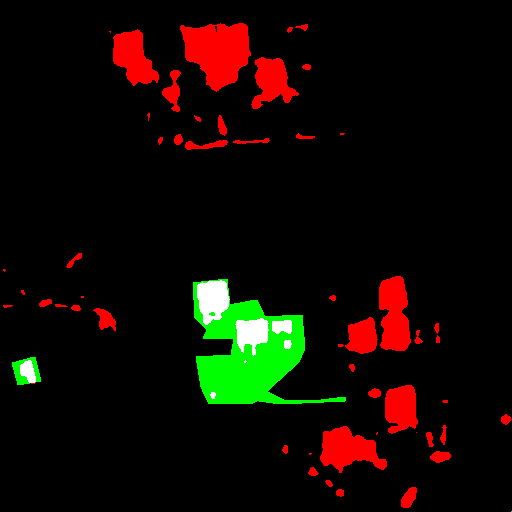}&
        \includegraphics[width=1.5cm]{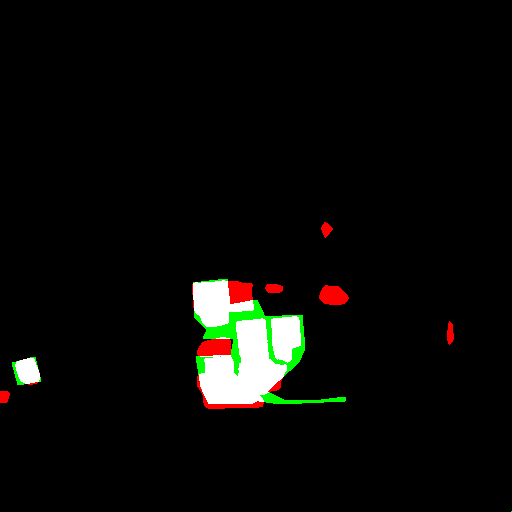}\\
        (c)&
        \includegraphics[width=1.5cm]{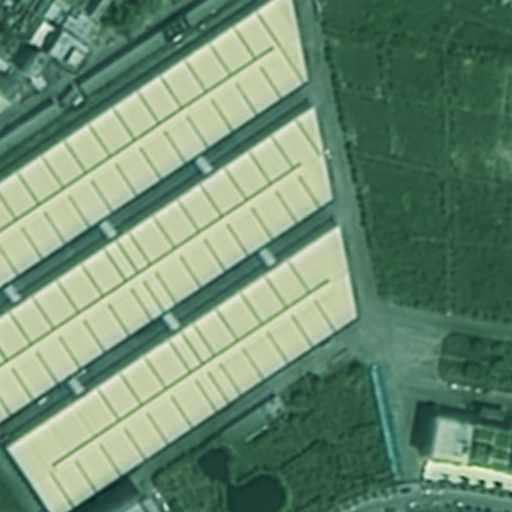} &
        \includegraphics[width=1.5cm]{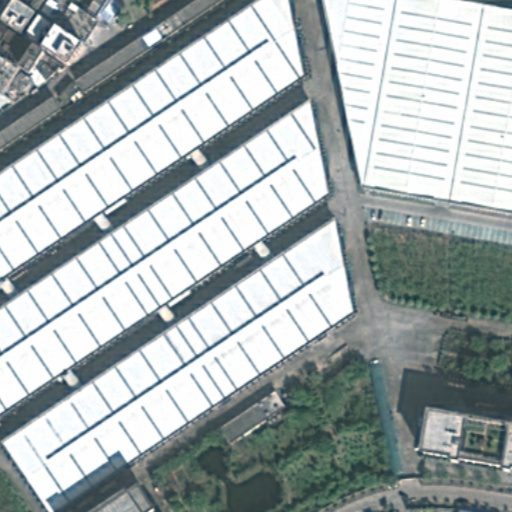} &
        \includegraphics[width=1.5cm]{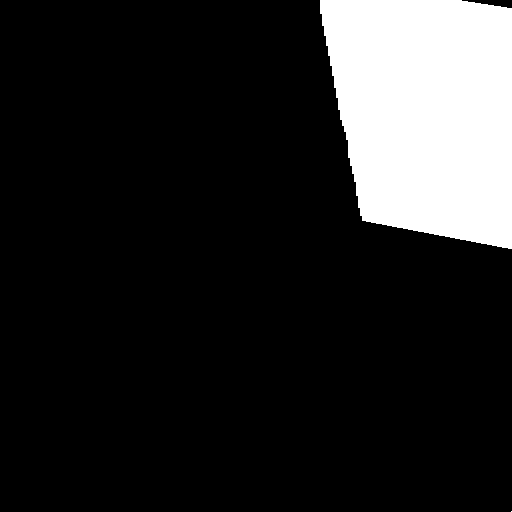} &
        \includegraphics[width=1.5cm]{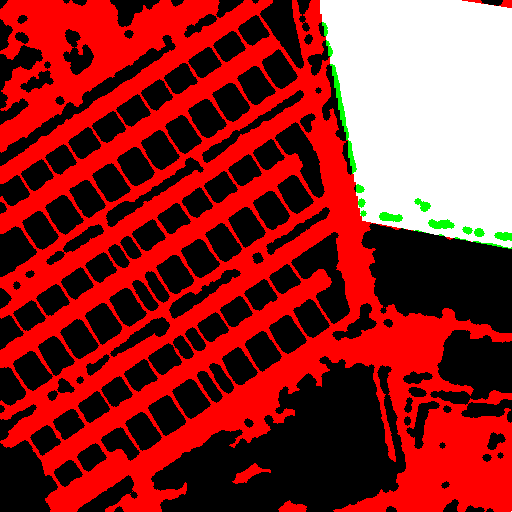} &
        \includegraphics[width=1.5cm]{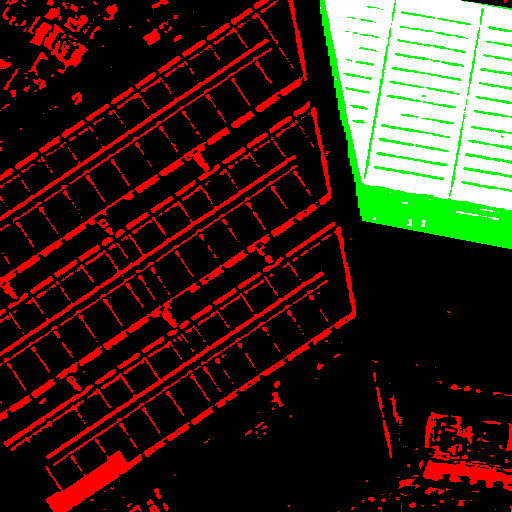} &
        \includegraphics[width=1.5cm]{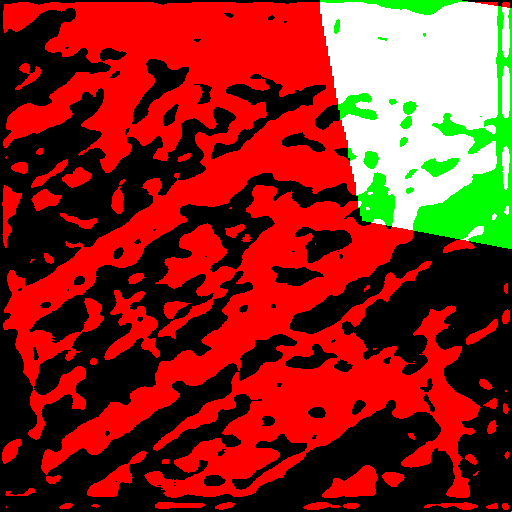} &
        \includegraphics[width=1.5cm]{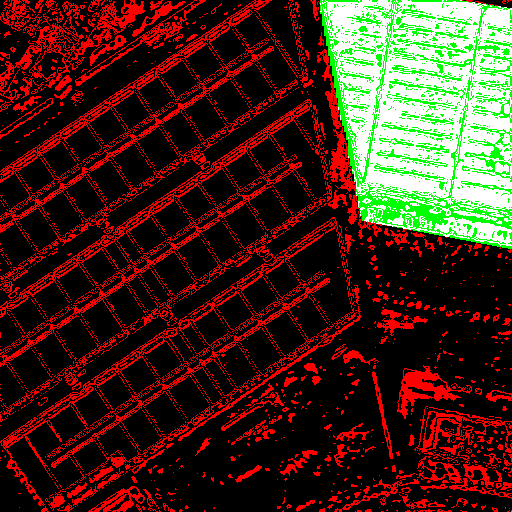} &
        \includegraphics[width=1.5cm]{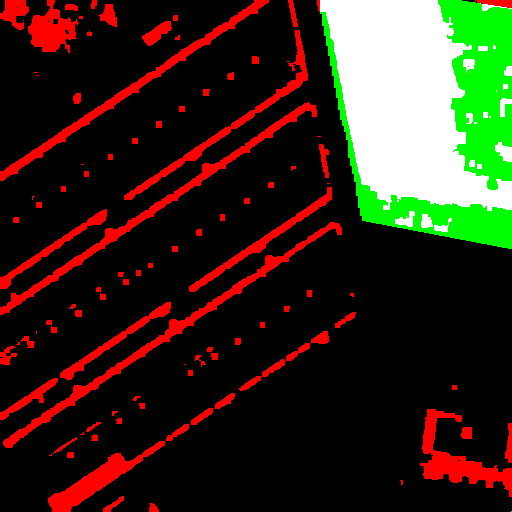}&
        \includegraphics[width=1.5cm]{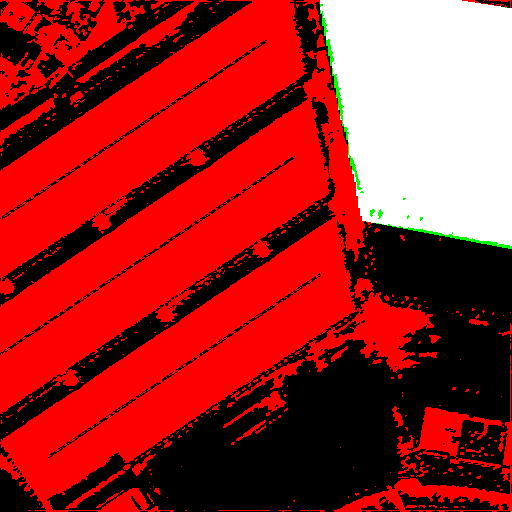}&
        \includegraphics[width=1.5cm]{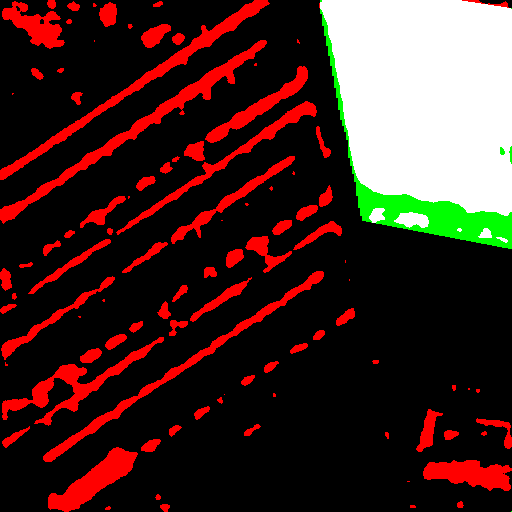}&
        \includegraphics[width=1.5cm]{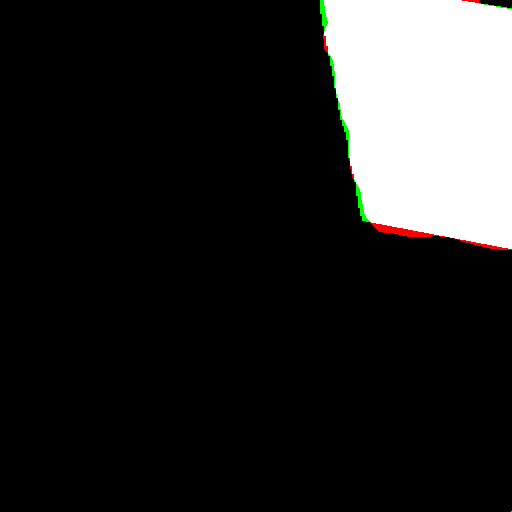}\\
        (d)&
        \includegraphics[width=1.5cm]{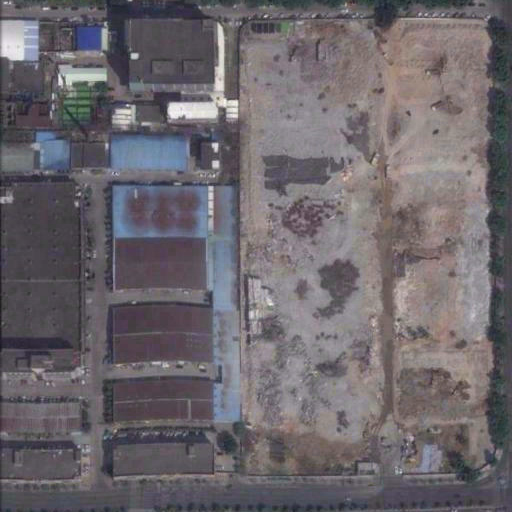} &
        \includegraphics[width=1.5cm]{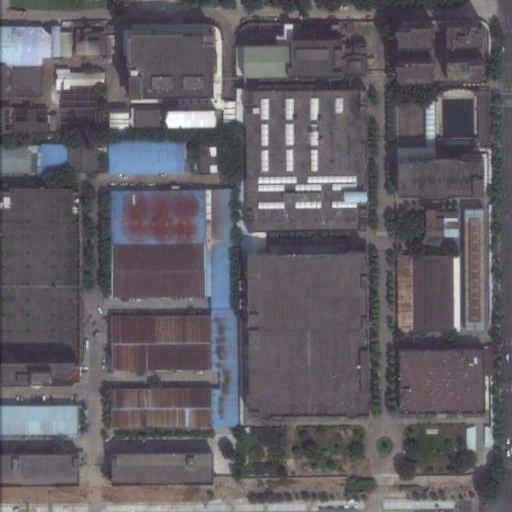} &
        \includegraphics[width=1.5cm]{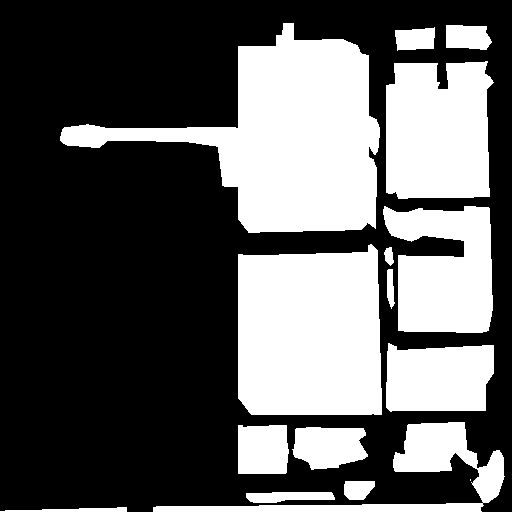} &
        \includegraphics[width=1.5cm]{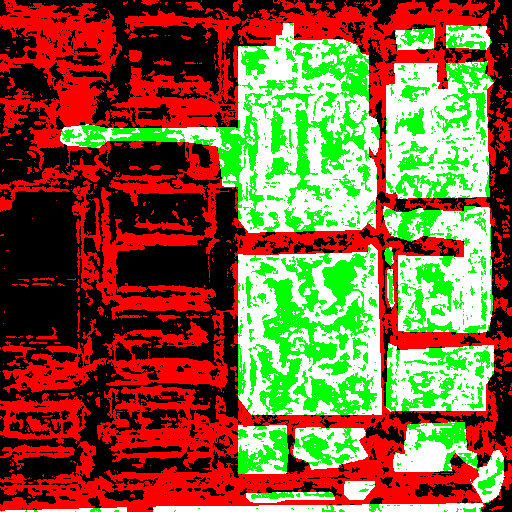} &
        \includegraphics[width=1.5cm]{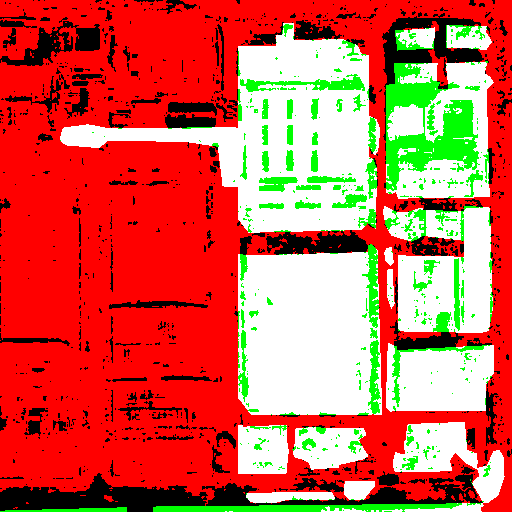} &
        \includegraphics[width=1.5cm]{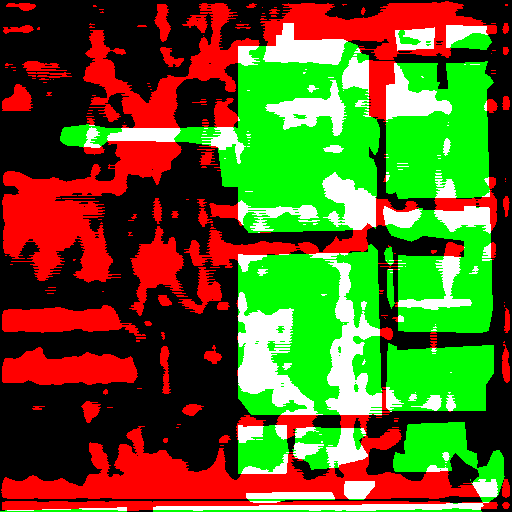} &
        \includegraphics[width=1.5cm]{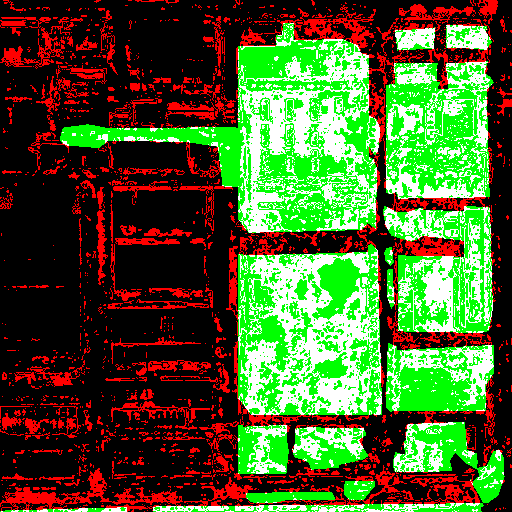} &
        \includegraphics[width=1.5cm]{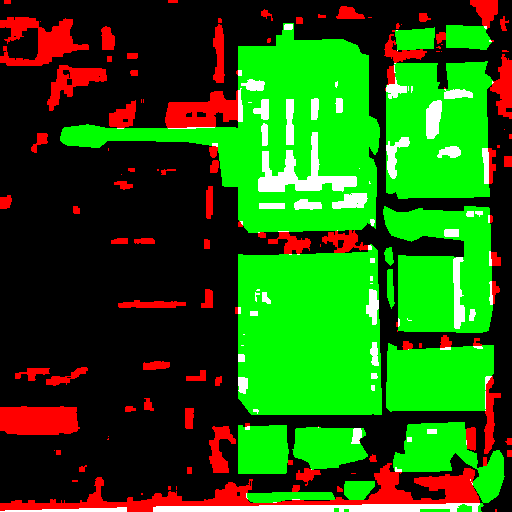}&
        \includegraphics[width=1.5cm]{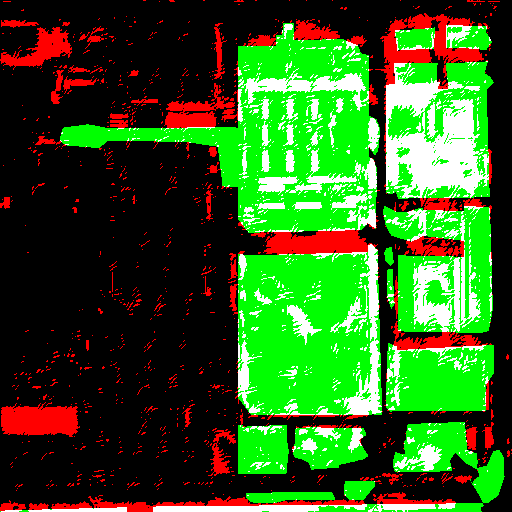}&
        \includegraphics[width=1.5cm]{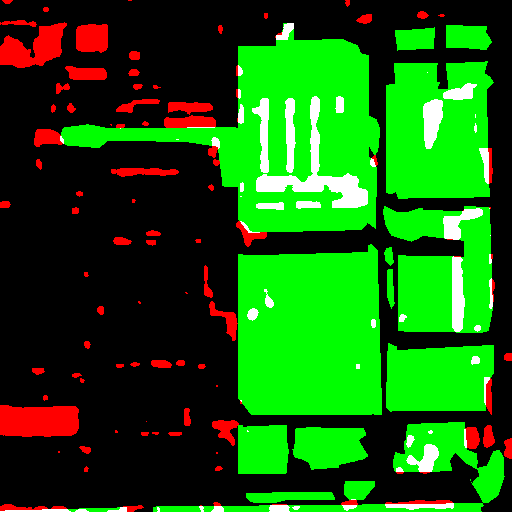}&
        \includegraphics[width=1.5cm]{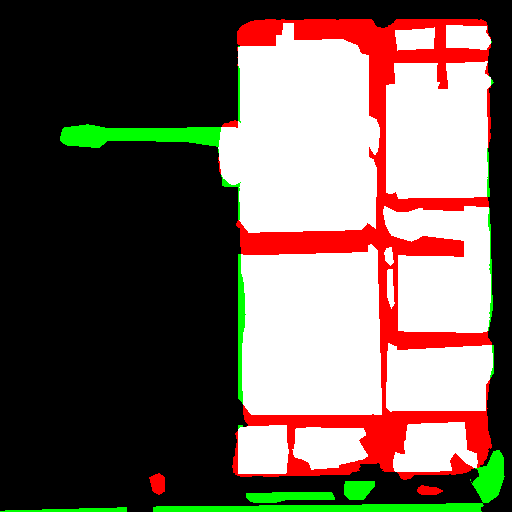}\\
        (e)&
        \includegraphics[width=1.5cm]{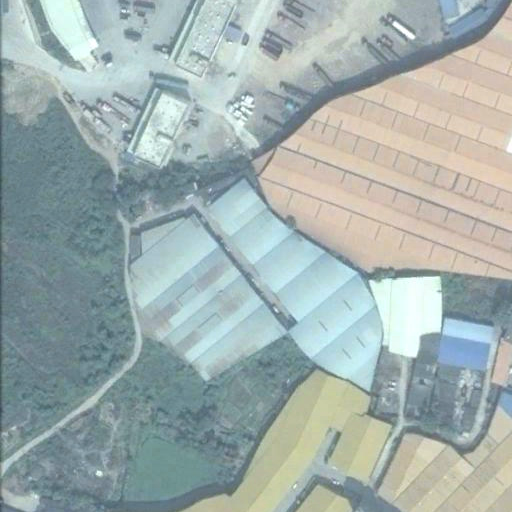} &
        \includegraphics[width=1.5cm]{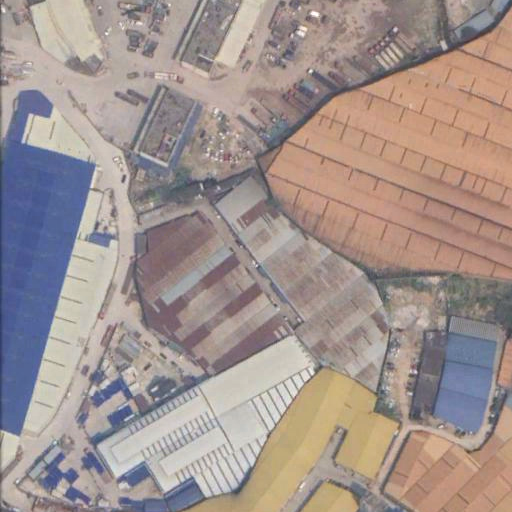} &
        \includegraphics[width=1.5cm]{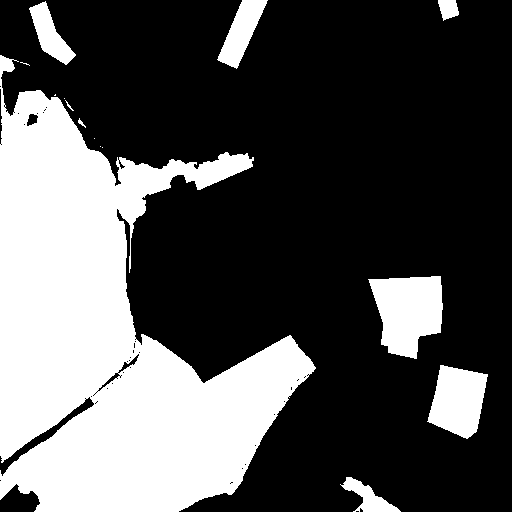} &
        \includegraphics[width=1.5cm]{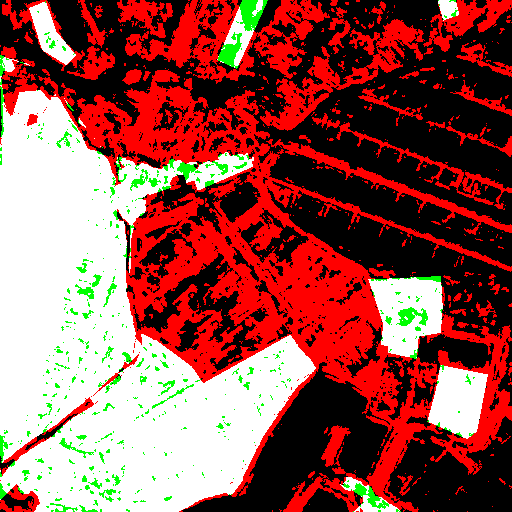} &
        \includegraphics[width=1.5cm]{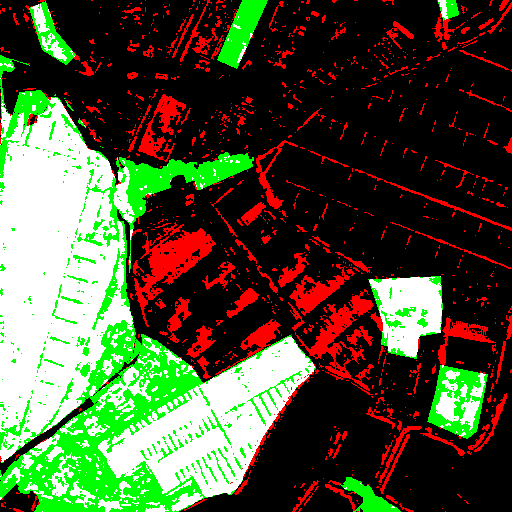} &
        \includegraphics[width=1.5cm]{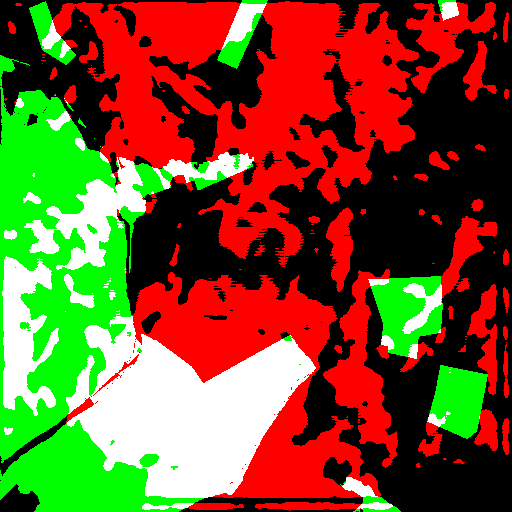} &
        \includegraphics[width=1.5cm]{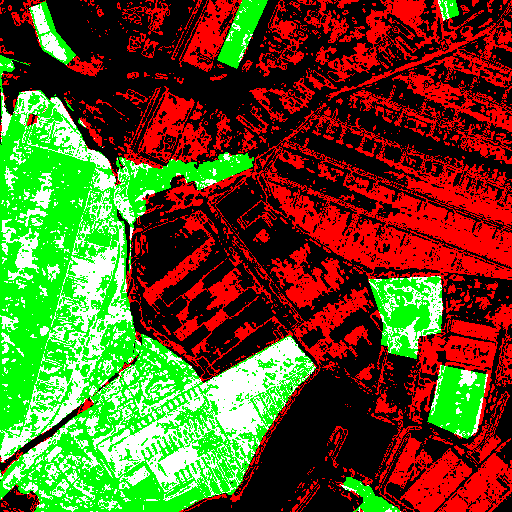} &
        \includegraphics[width=1.5cm]{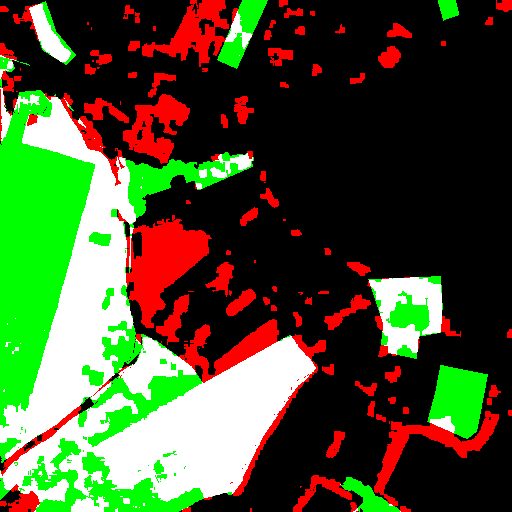}&
        \includegraphics[width=1.5cm]{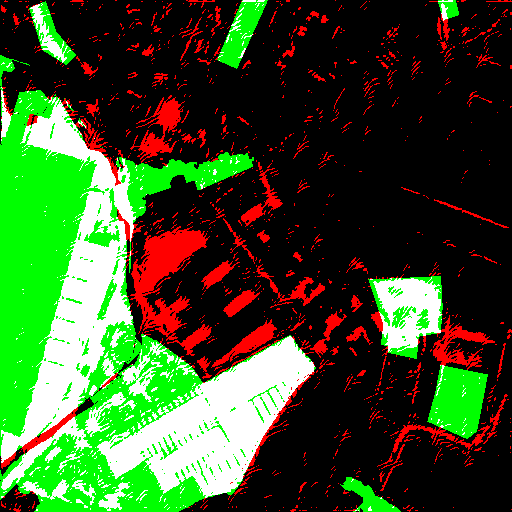}&
        \includegraphics[width=1.5cm]{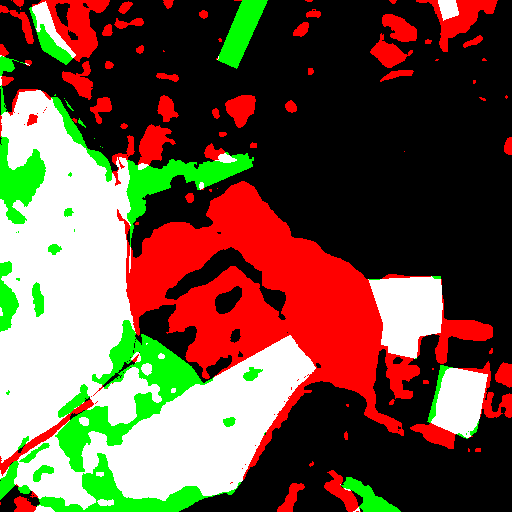}&
        \includegraphics[width=1.5cm]{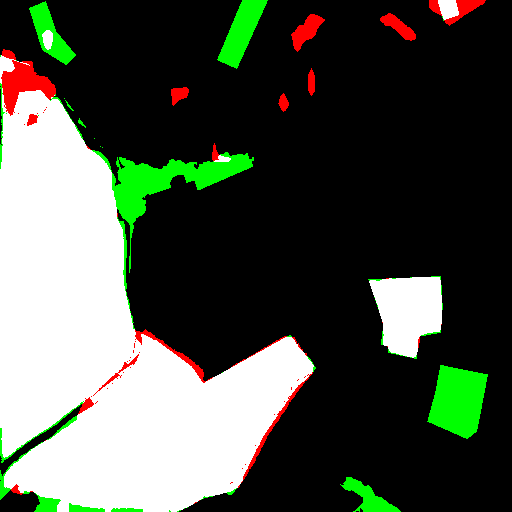}\\
        (f)&
        \includegraphics[width=1.5cm]{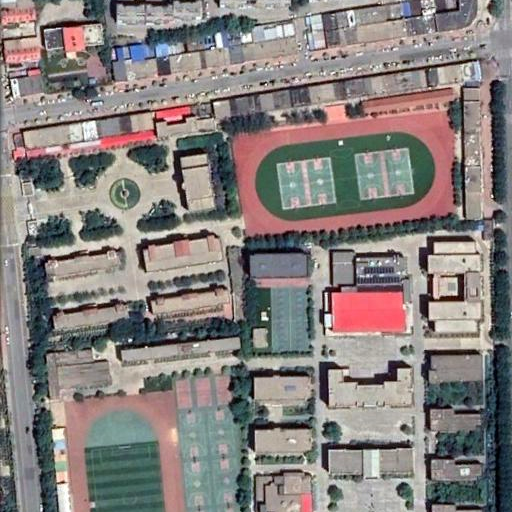} &
        \includegraphics[width=1.5cm]{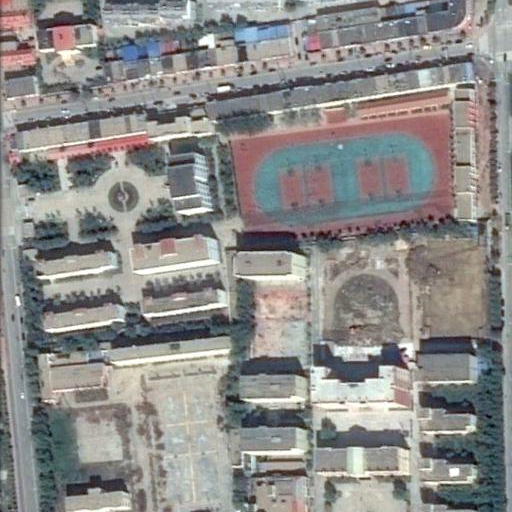} &
        \includegraphics[width=1.5cm]{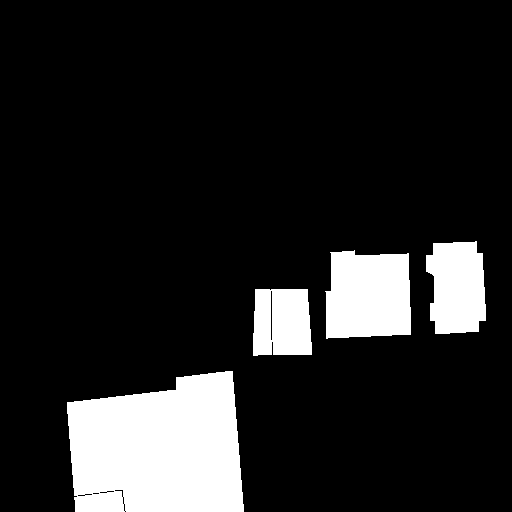} &
        \includegraphics[width=1.5cm]{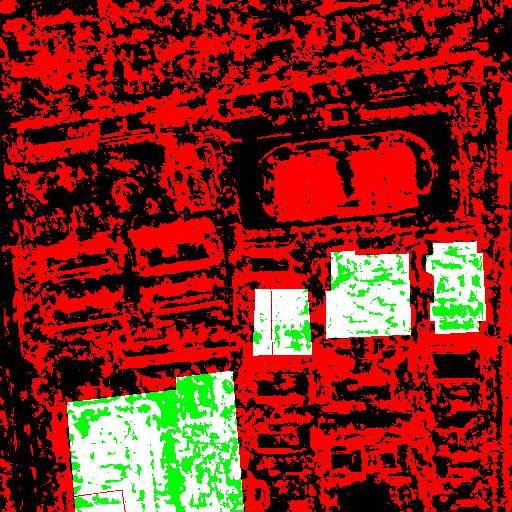} &
        \includegraphics[width=1.5cm]{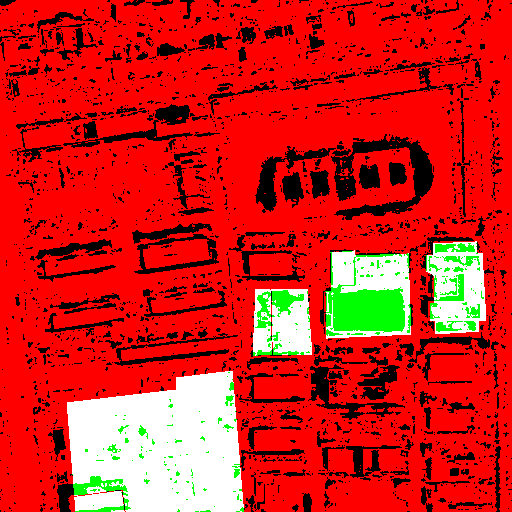} &
        \includegraphics[width=1.5cm]{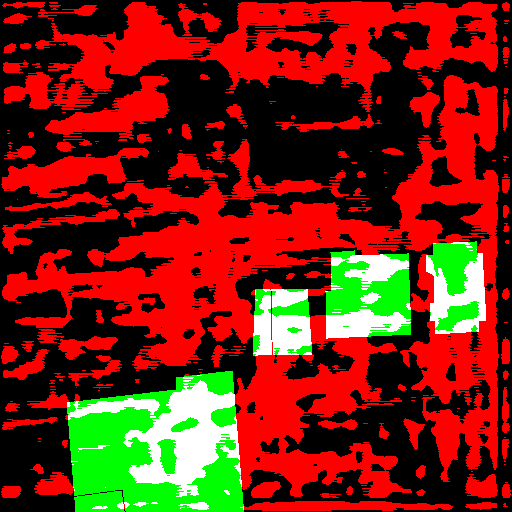} &
        \includegraphics[width=1.5cm]{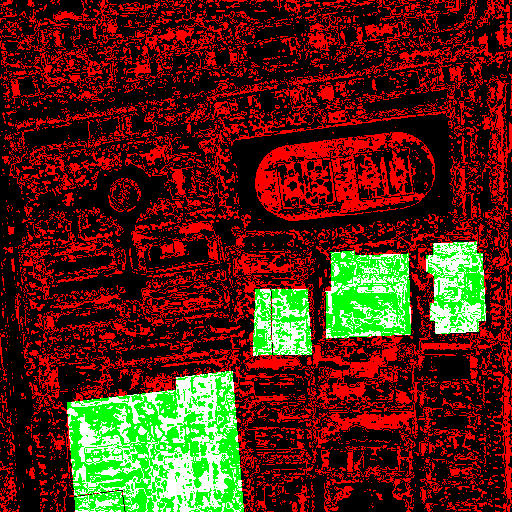} &
        \includegraphics[width=1.5cm]{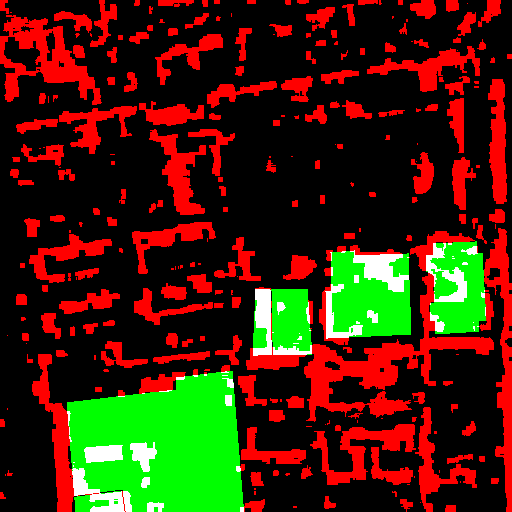}&
        \includegraphics[width=1.5cm]{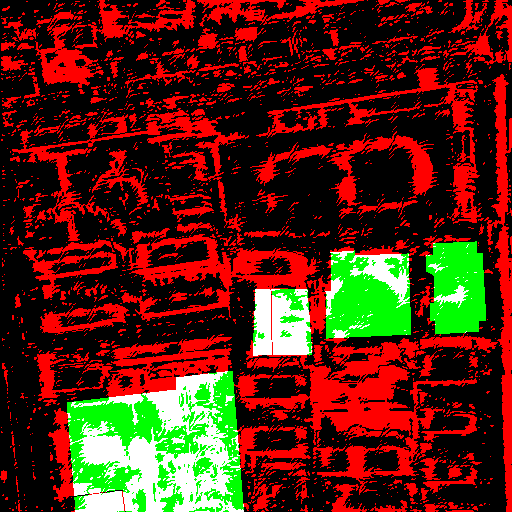}&
        \includegraphics[width=1.5cm]{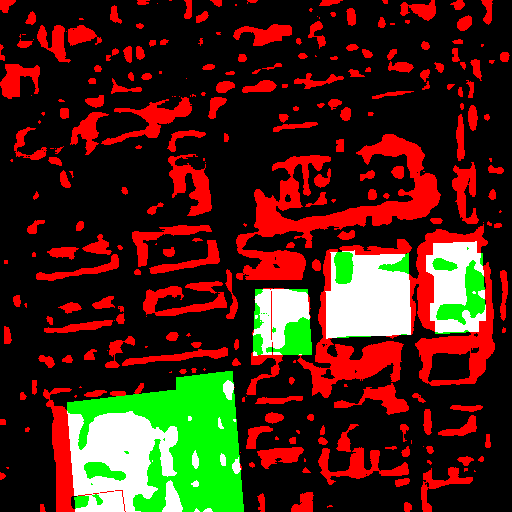}&
        \includegraphics[width=1.5cm]{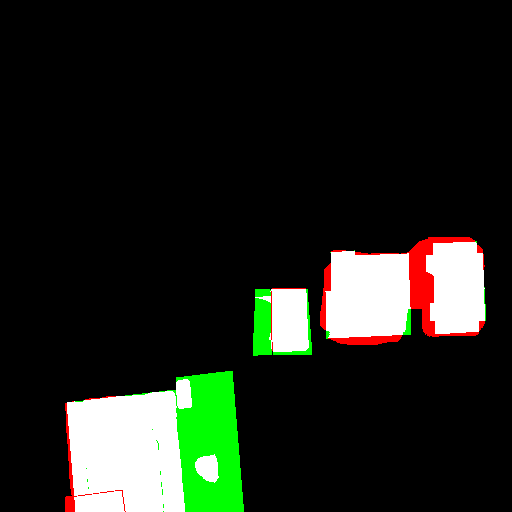}\\
        (g)&
        \includegraphics[width=1.5cm]{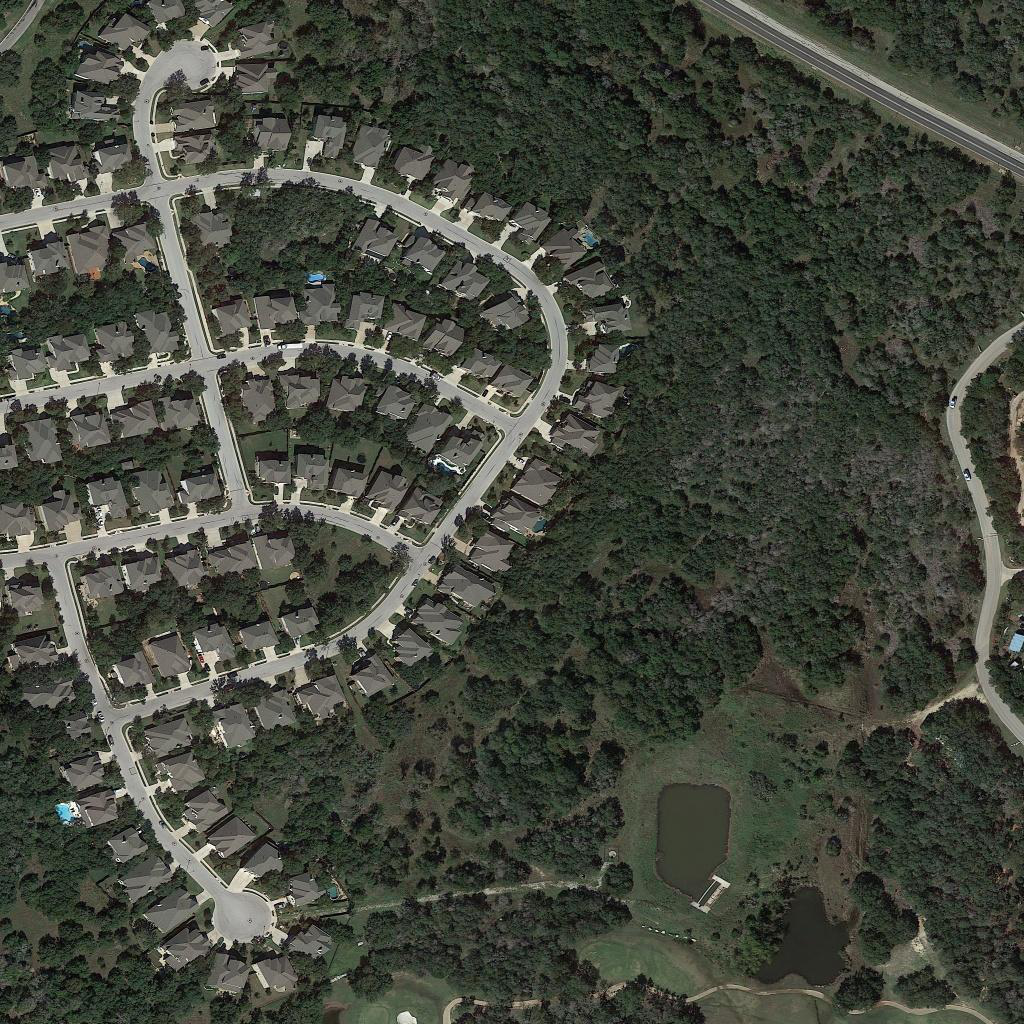} &
        \includegraphics[width=1.5cm]{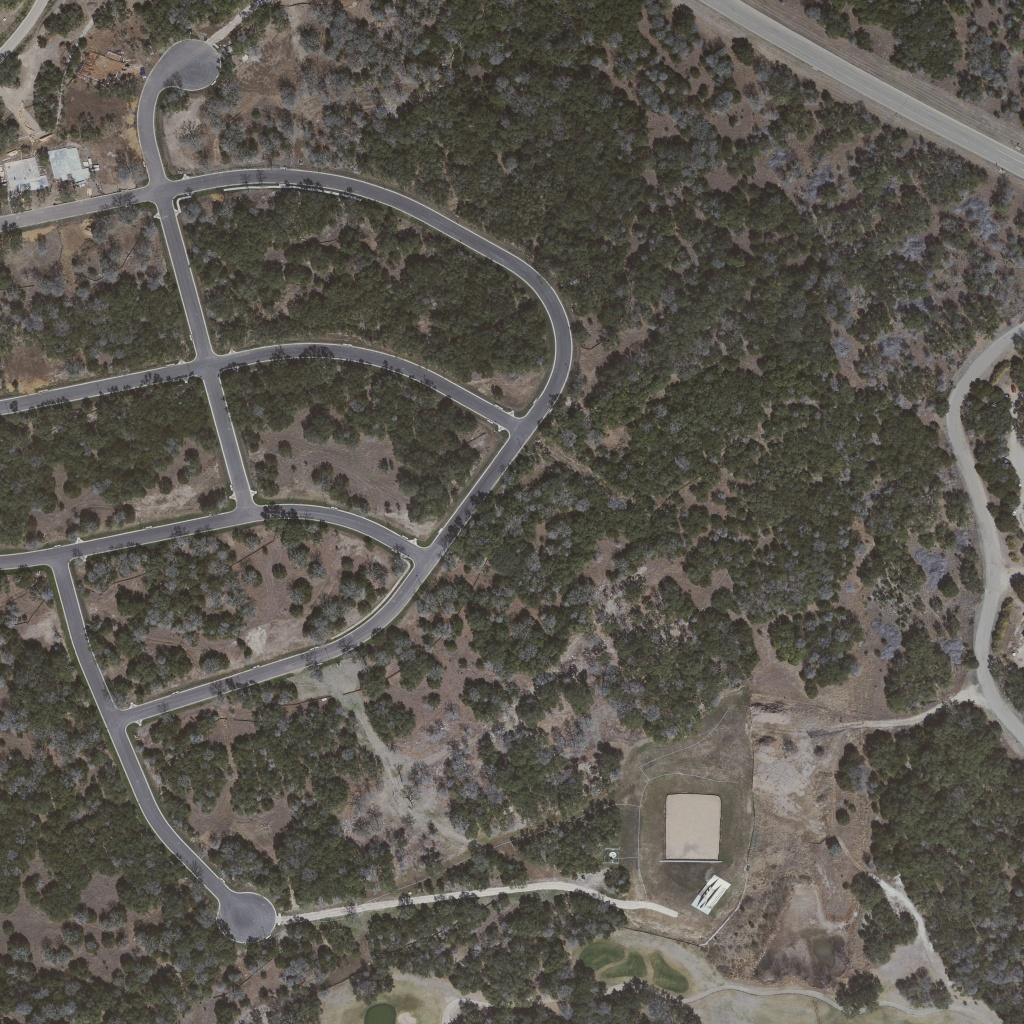} &
        \includegraphics[width=1.5cm]{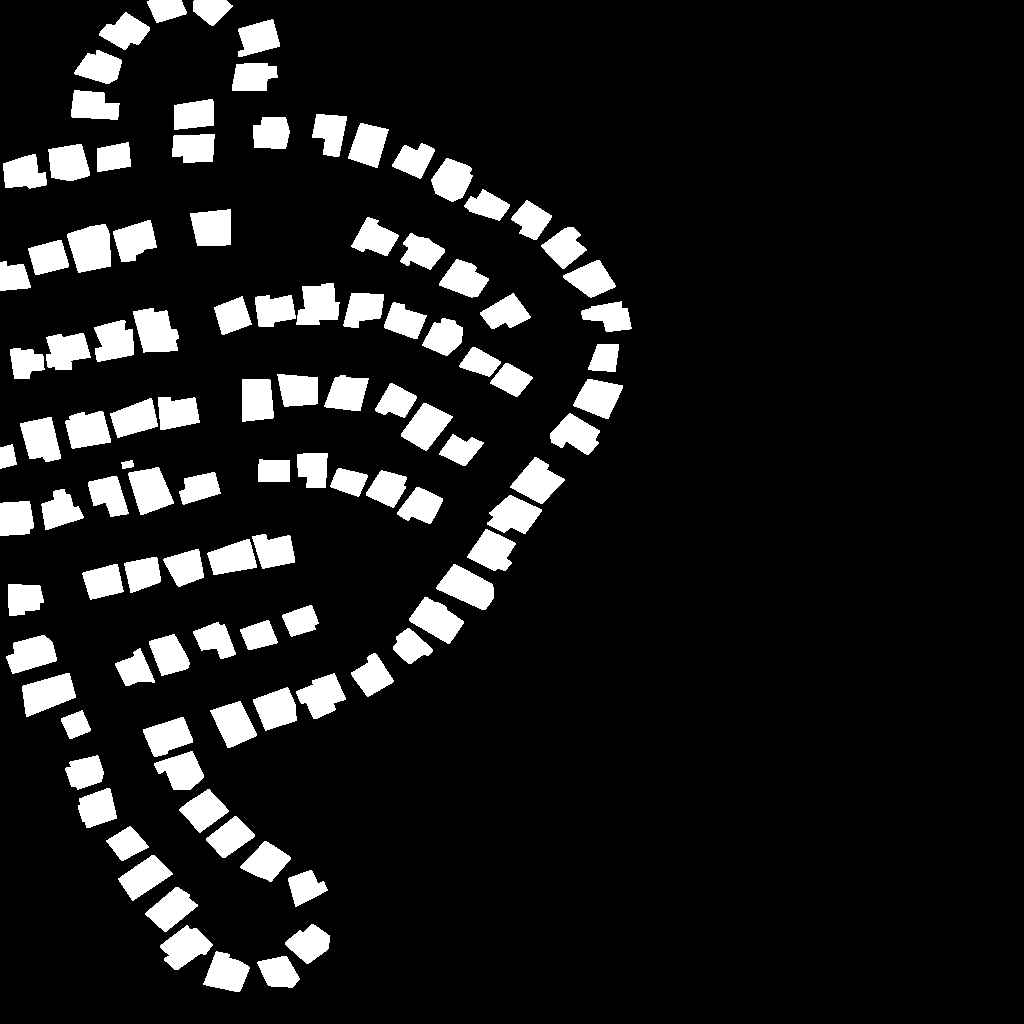} &
        \includegraphics[width=1.5cm]{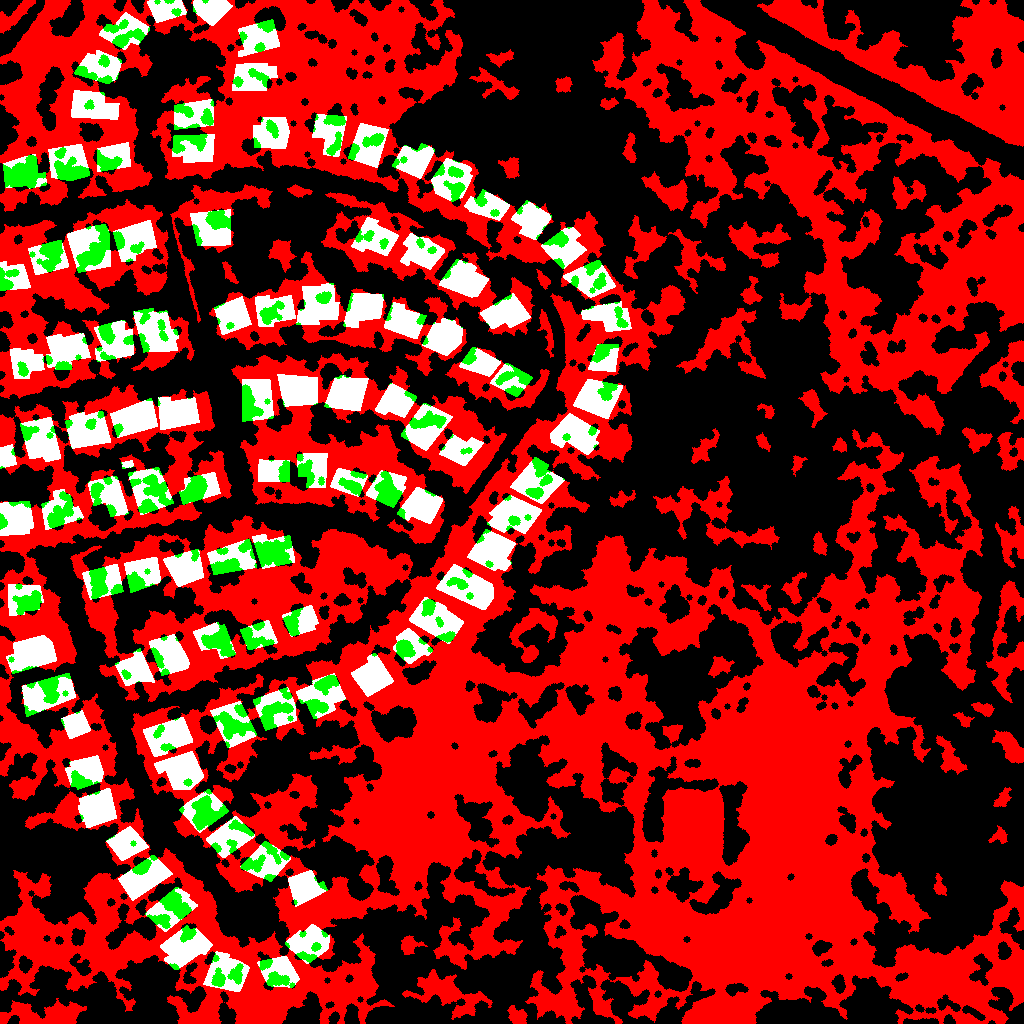} &
        \includegraphics[width=1.5cm]{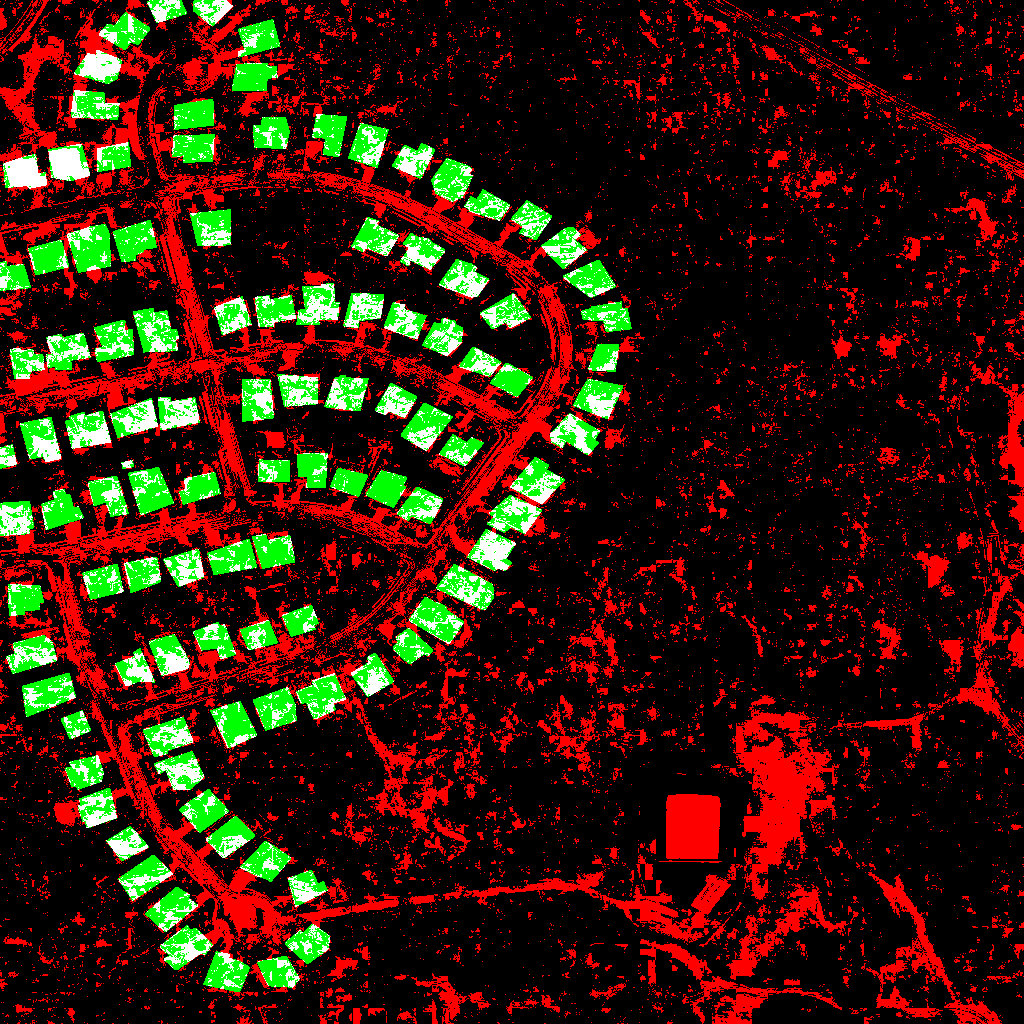} &
        \includegraphics[width=1.5cm]{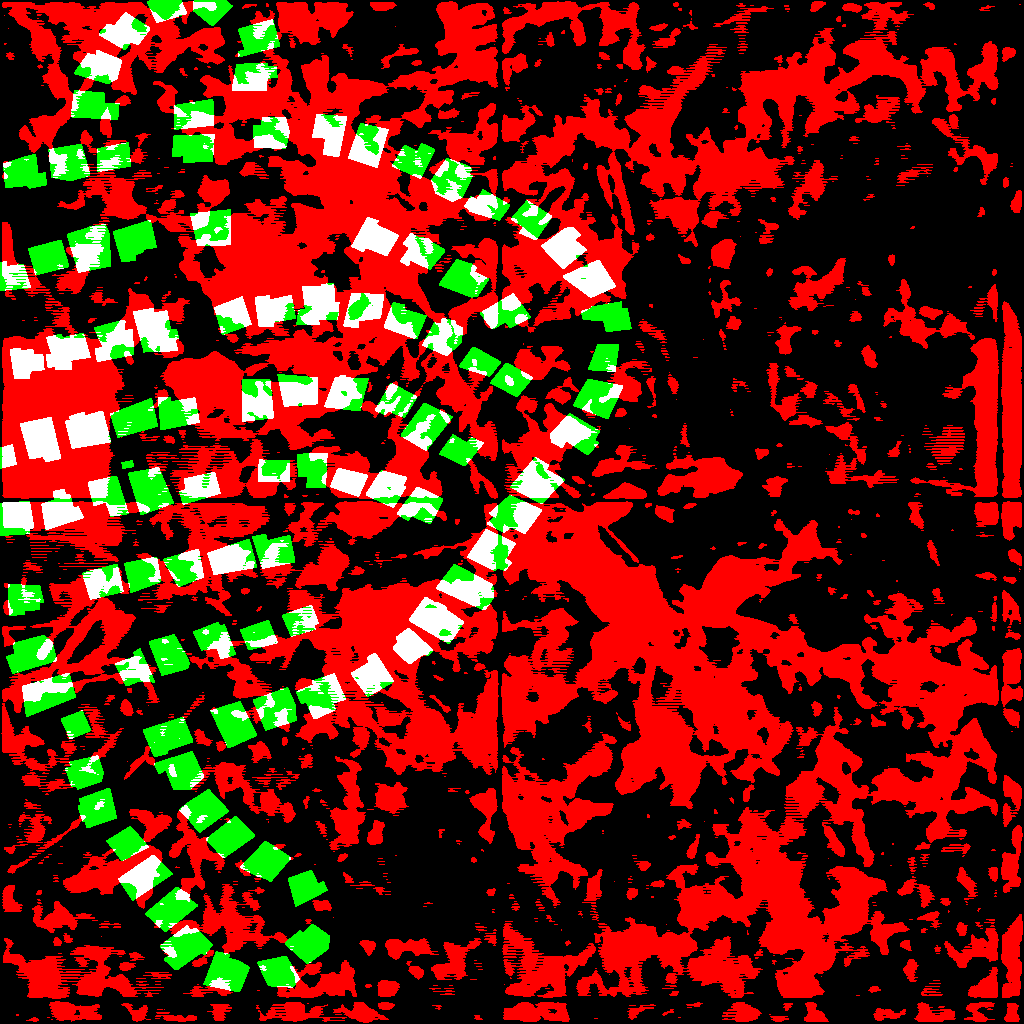} &
        \includegraphics[width=1.5cm]{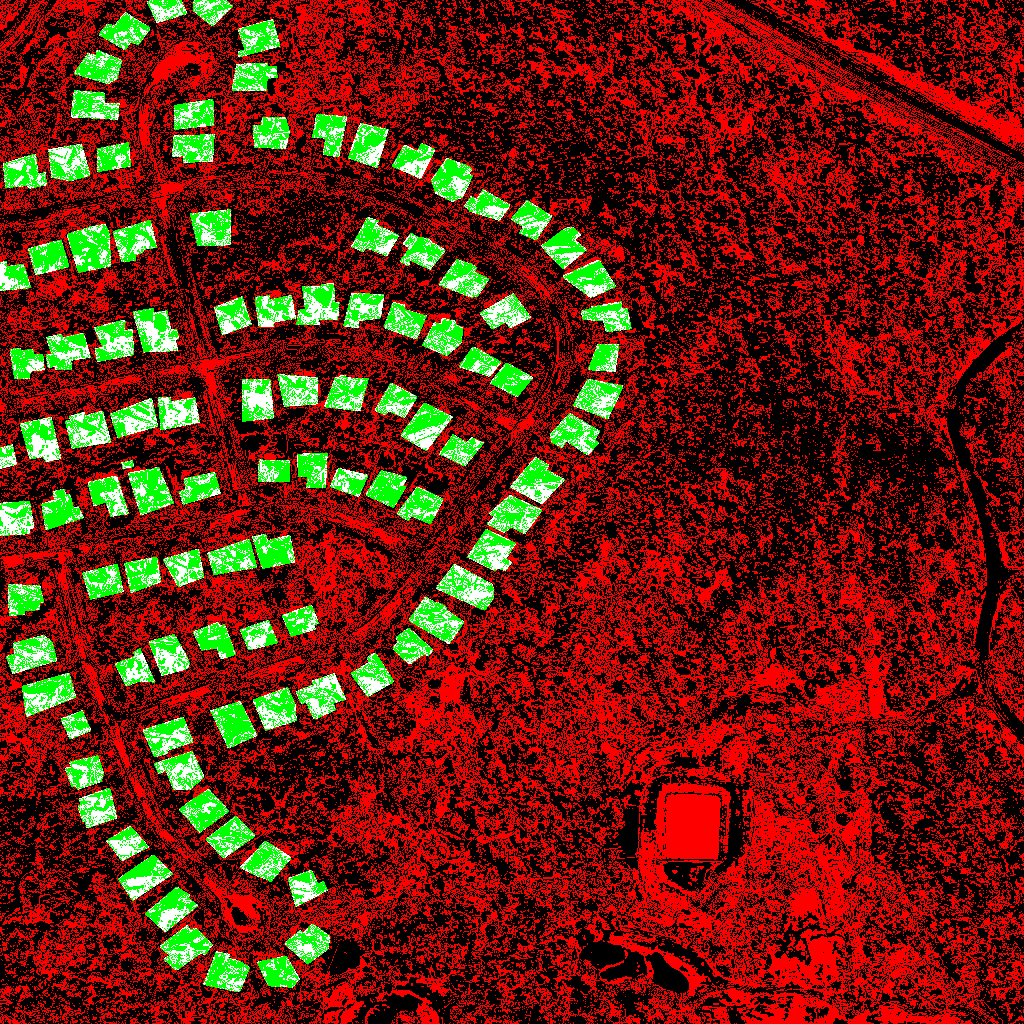} &
        \includegraphics[width=1.5cm]{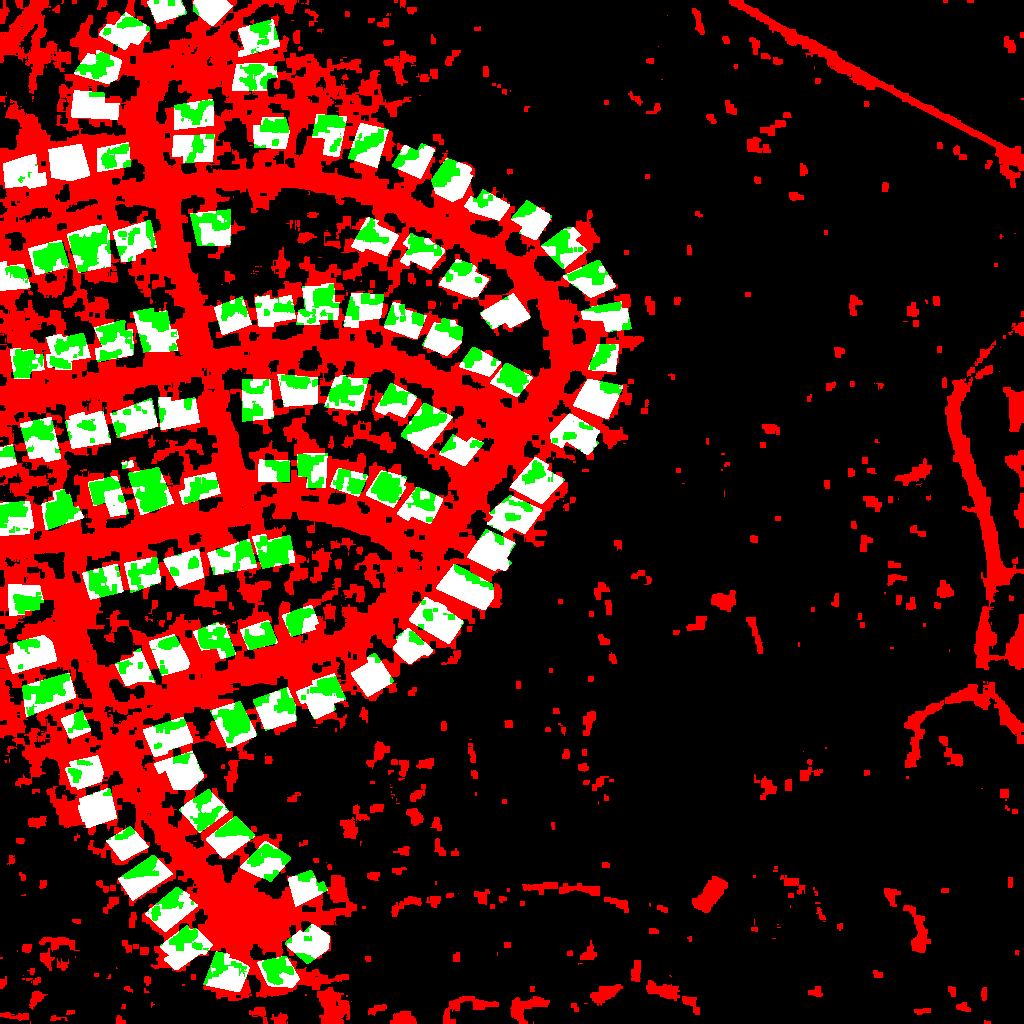}&
        \includegraphics[width=1.5cm]{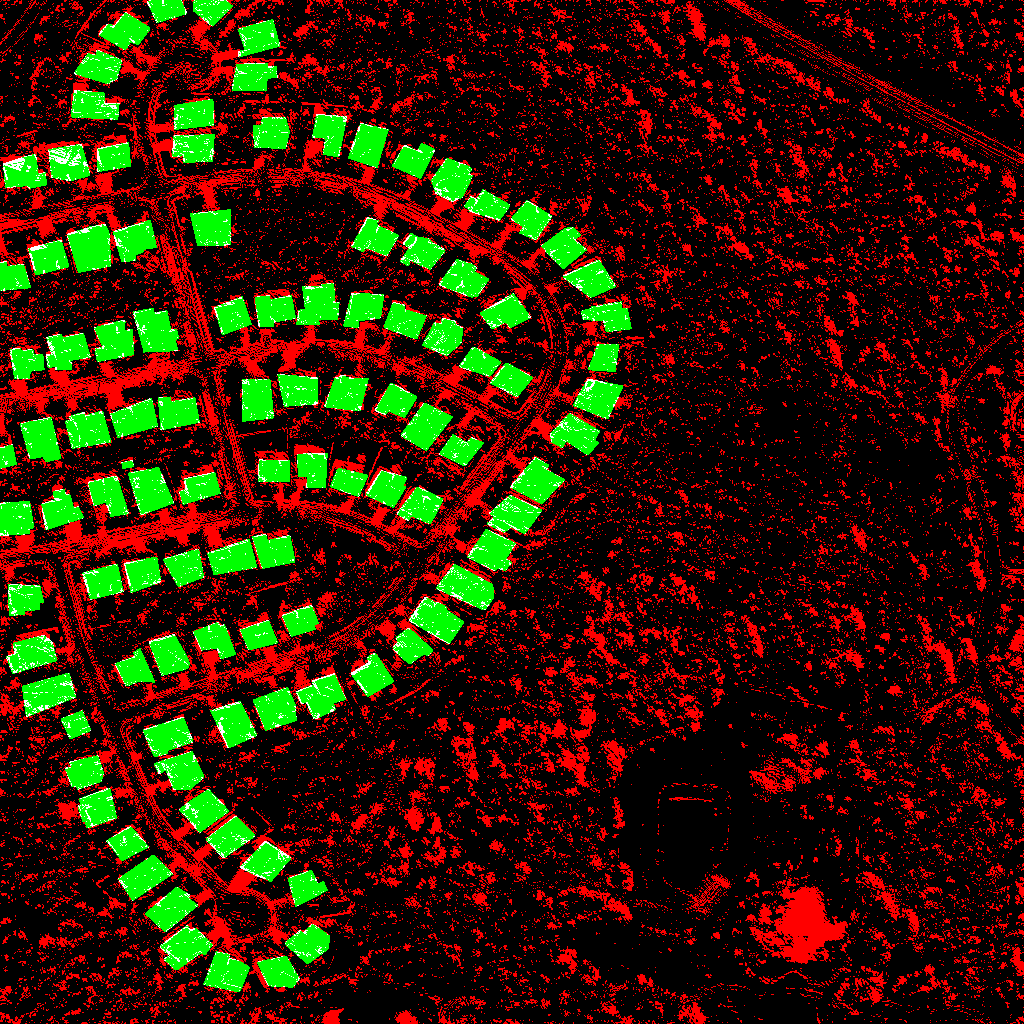}&
        \includegraphics[width=1.5cm]{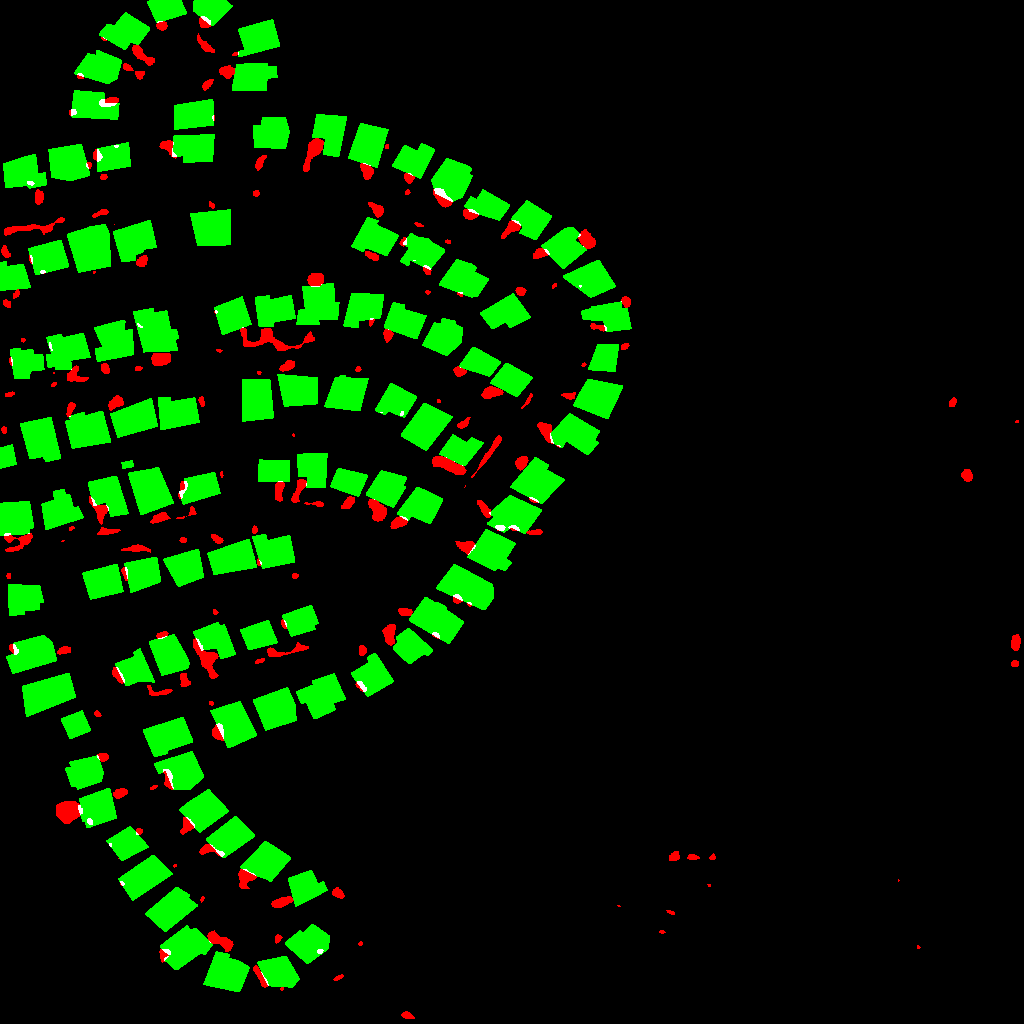}&
        \includegraphics[width=1.5cm]{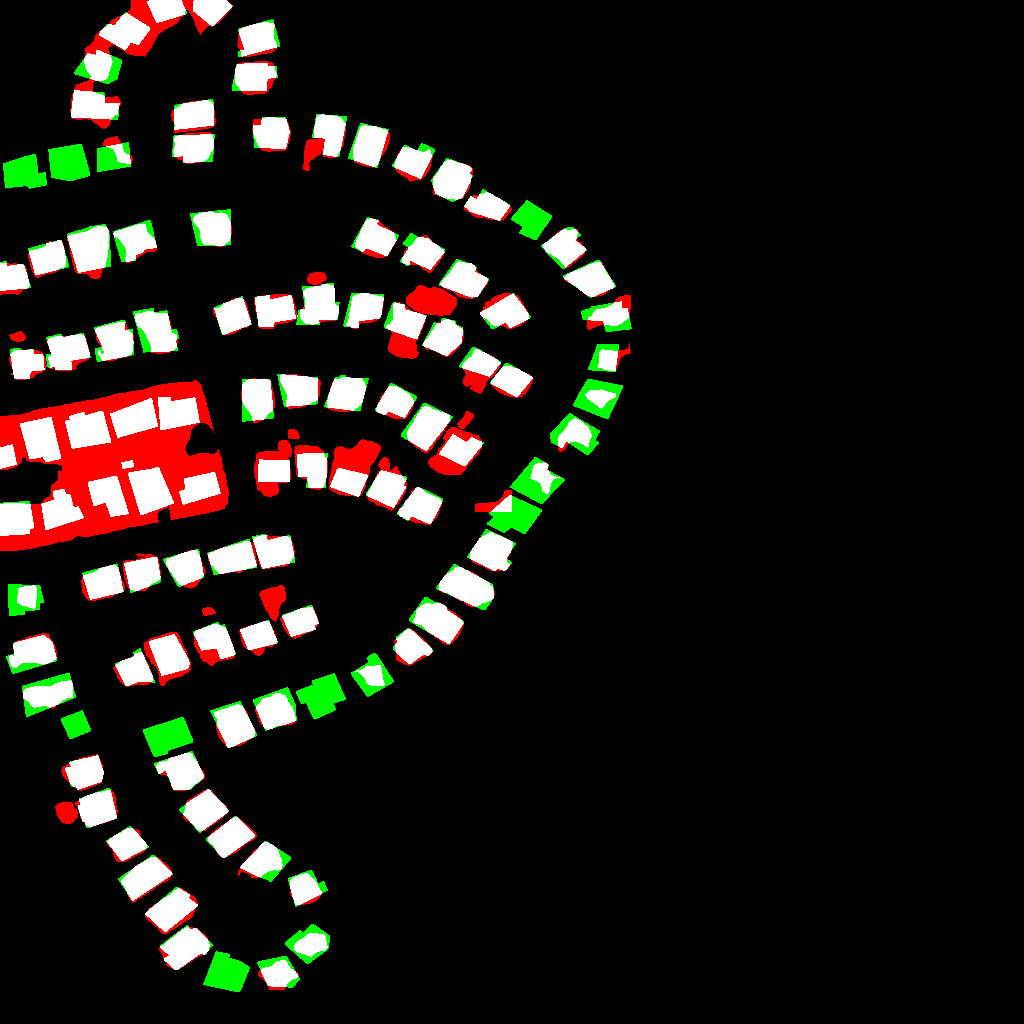}\\
        (h)&
        \includegraphics[width=1.5cm]{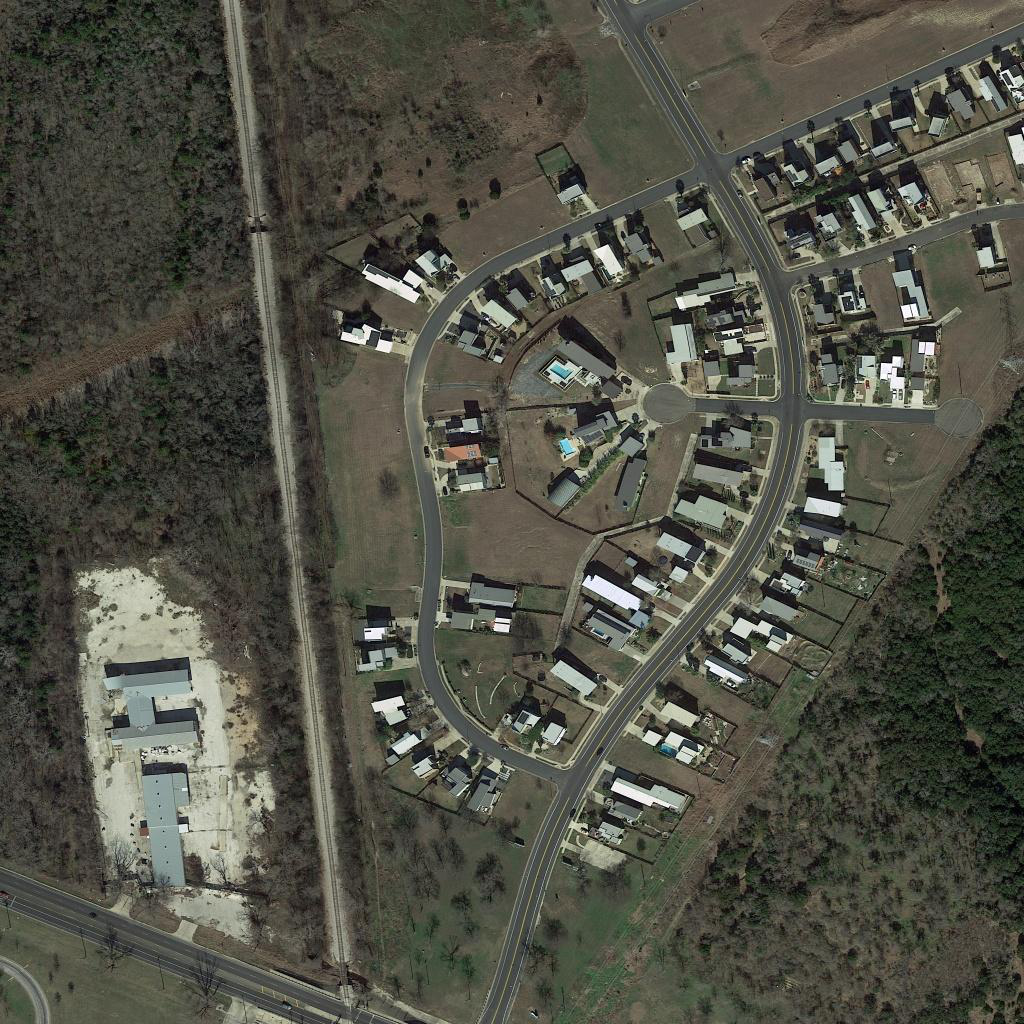} &
        \includegraphics[width=1.5cm]{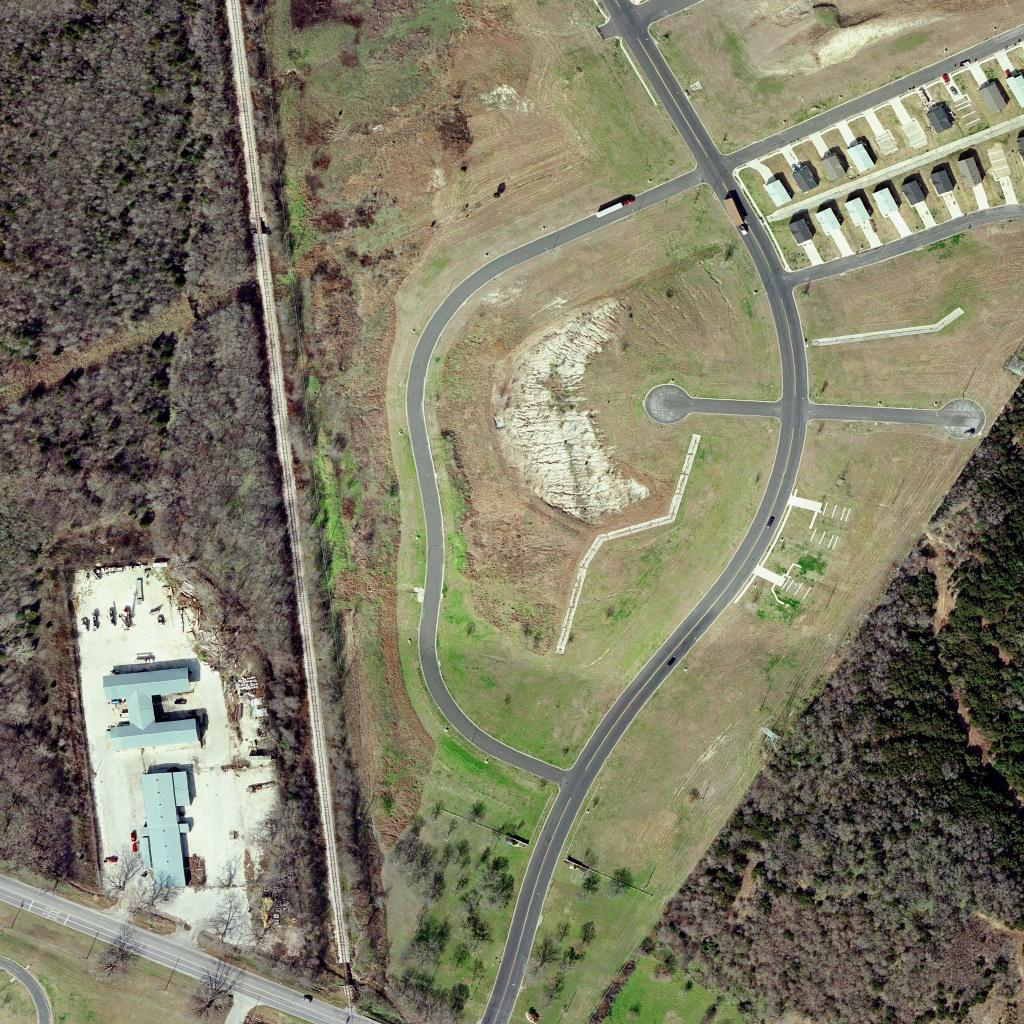} &
        \includegraphics[width=1.5cm]{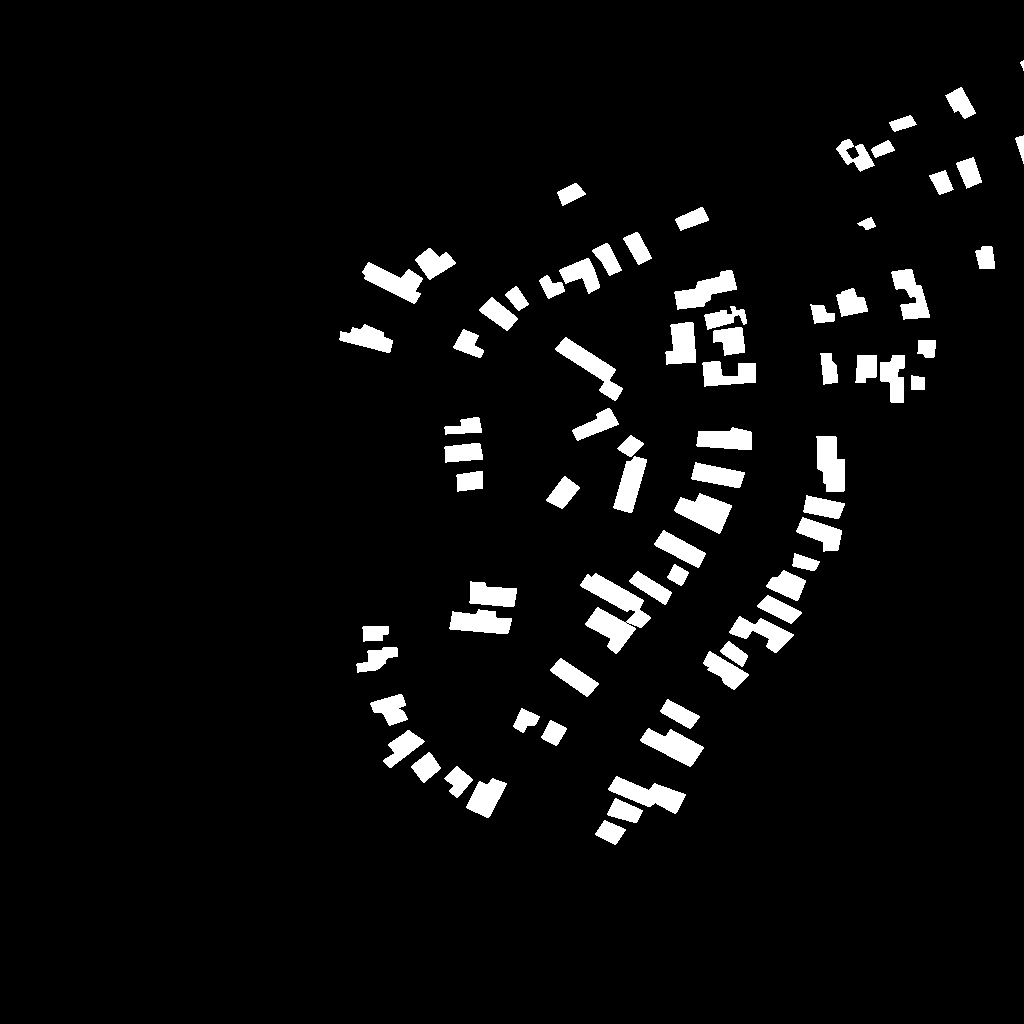} &
        \includegraphics[width=1.5cm]{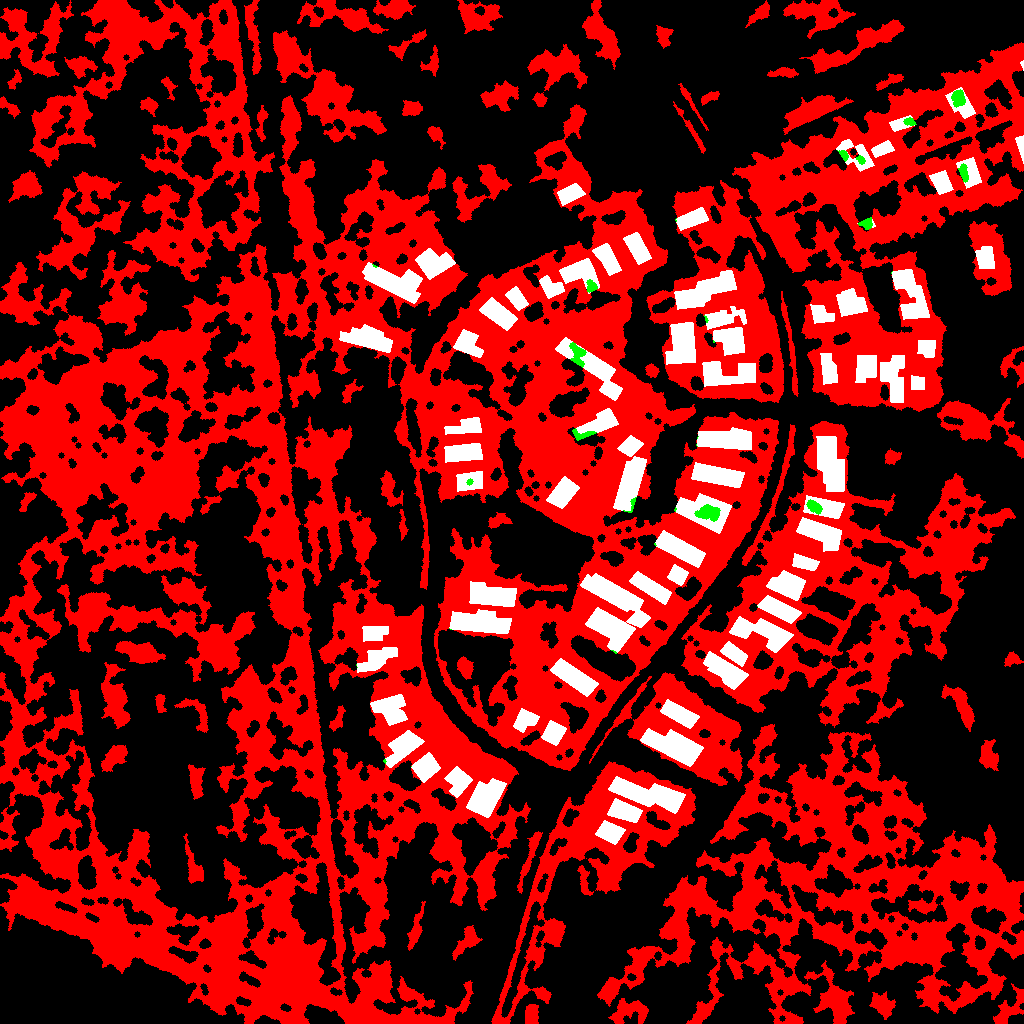} &
        \includegraphics[width=1.5cm]{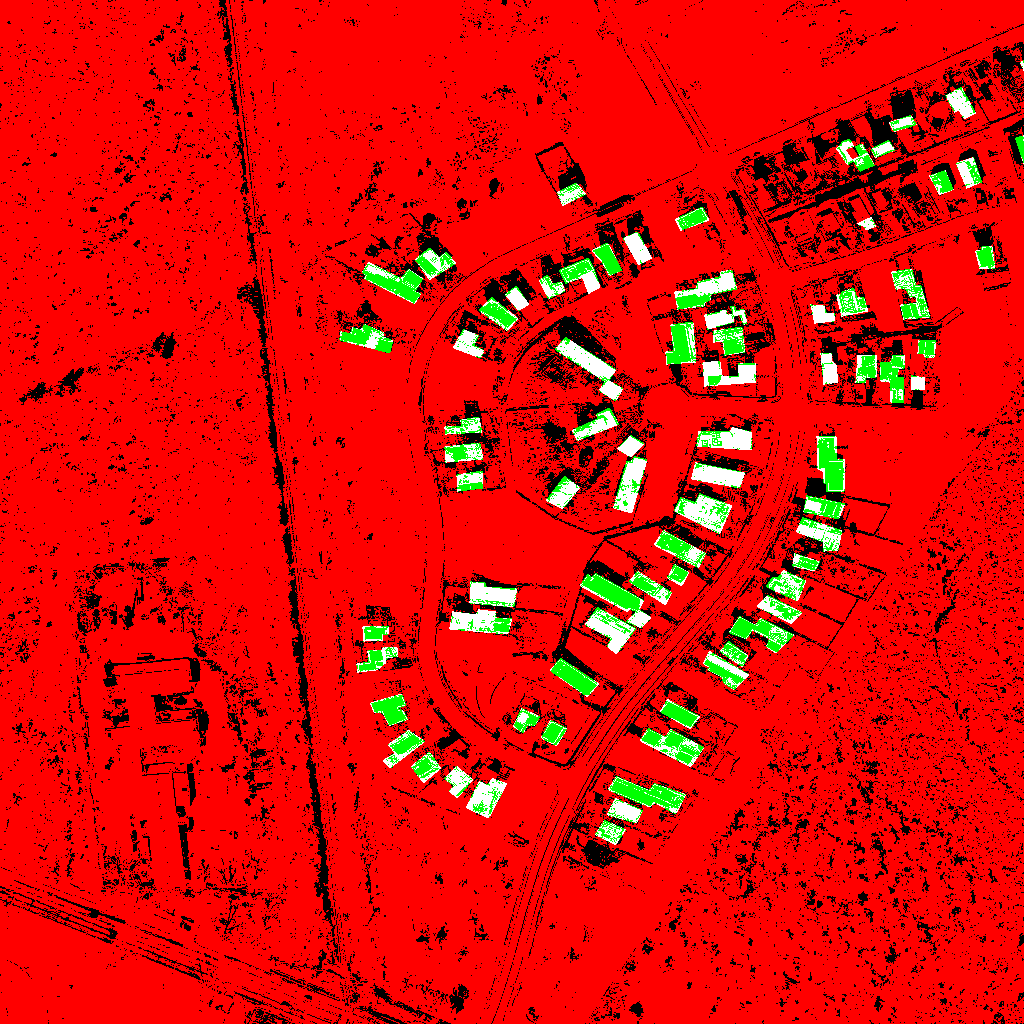} &
        \includegraphics[width=1.5cm]{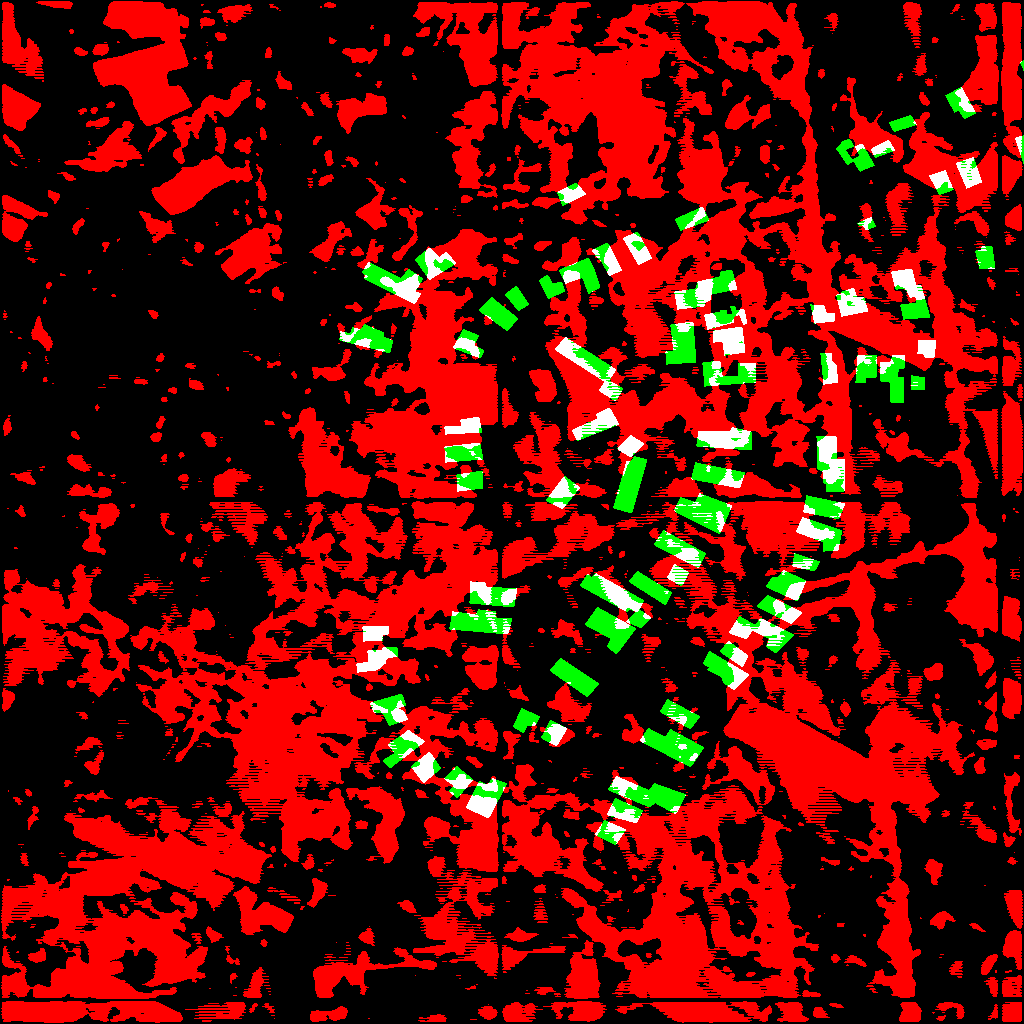} &
        \includegraphics[width=1.5cm]{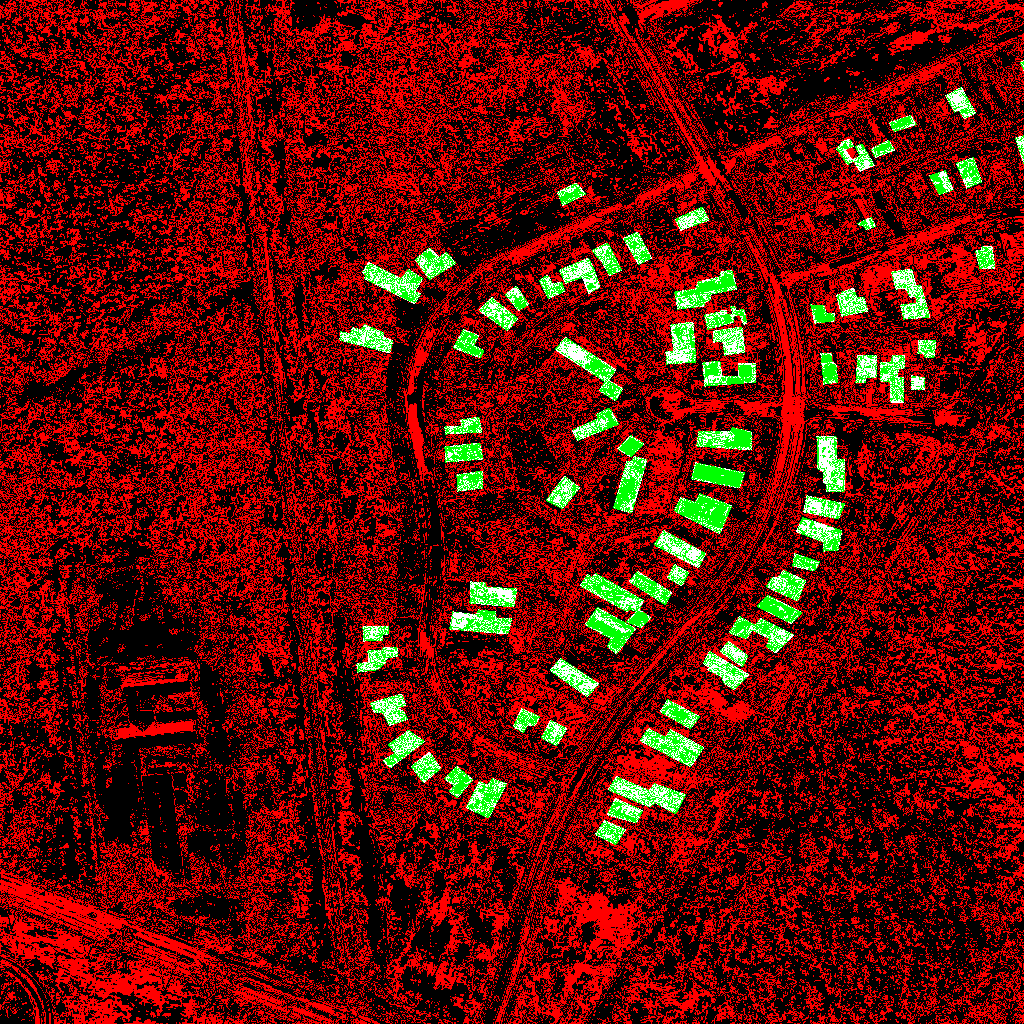} &
        \includegraphics[width=1.5cm]{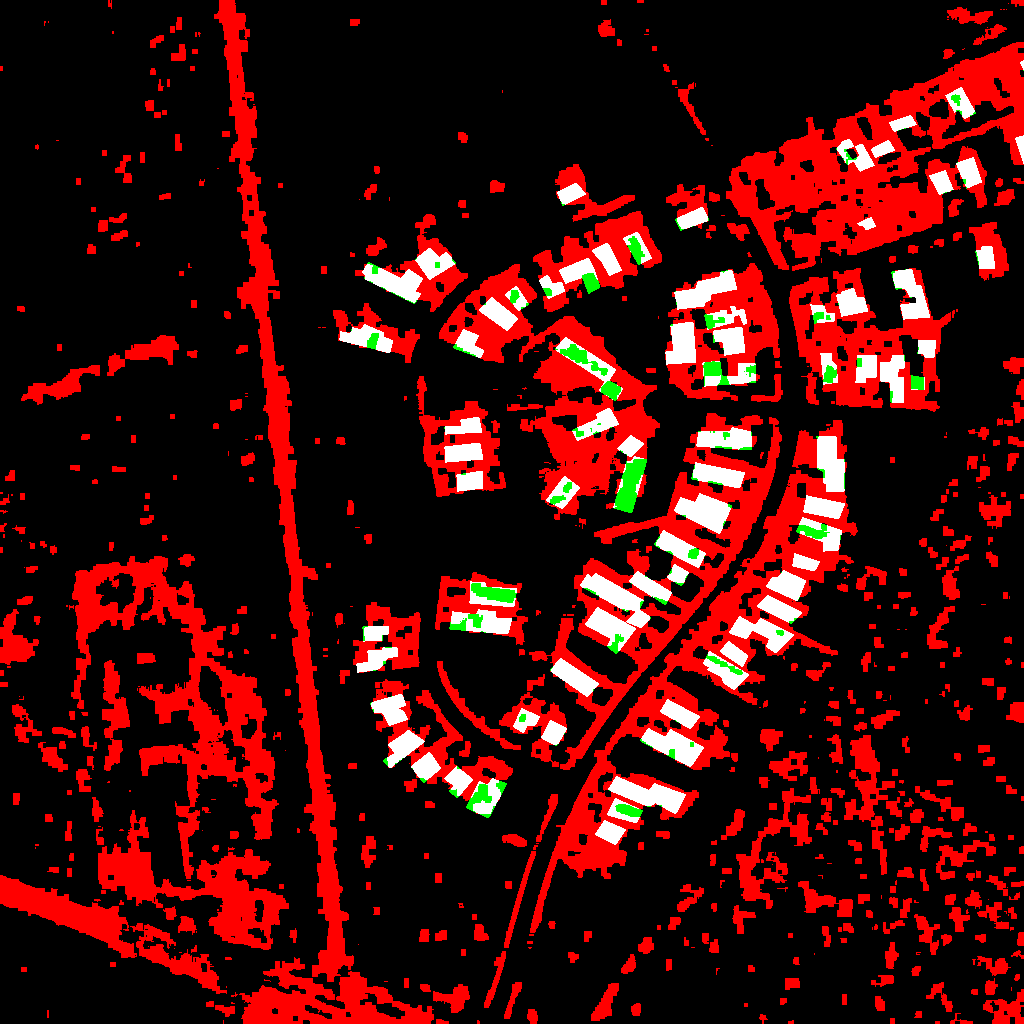}&
        \includegraphics[width=1.5cm]{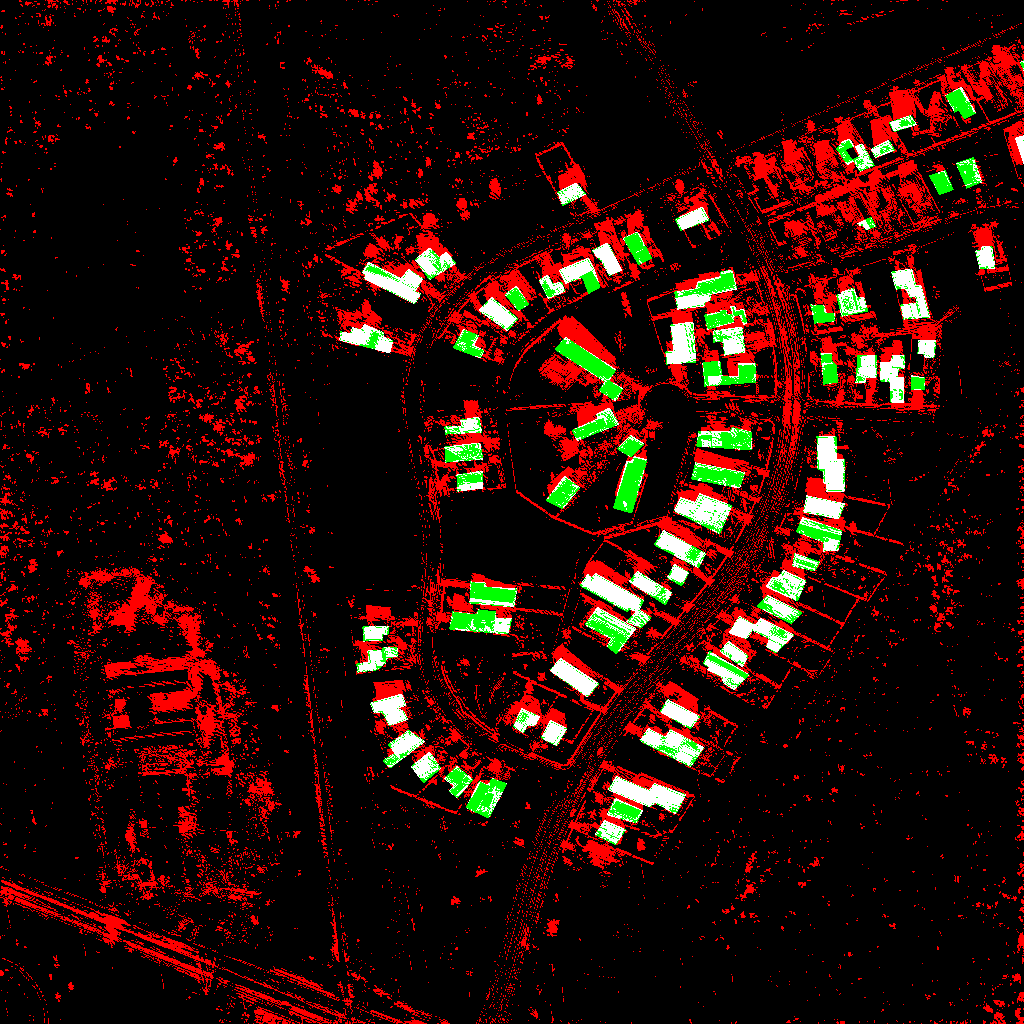}&
        \includegraphics[width=1.5cm]{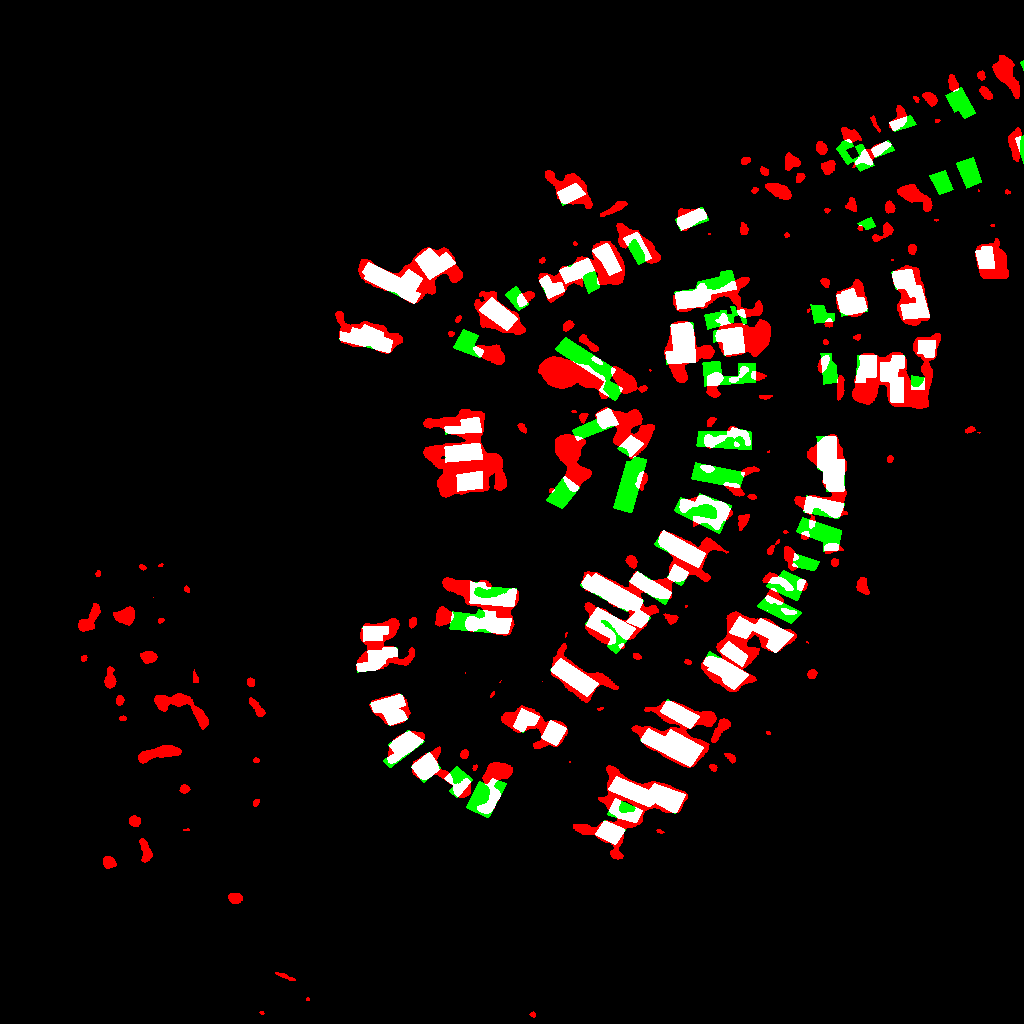}&
        \includegraphics[width=1.5cm]{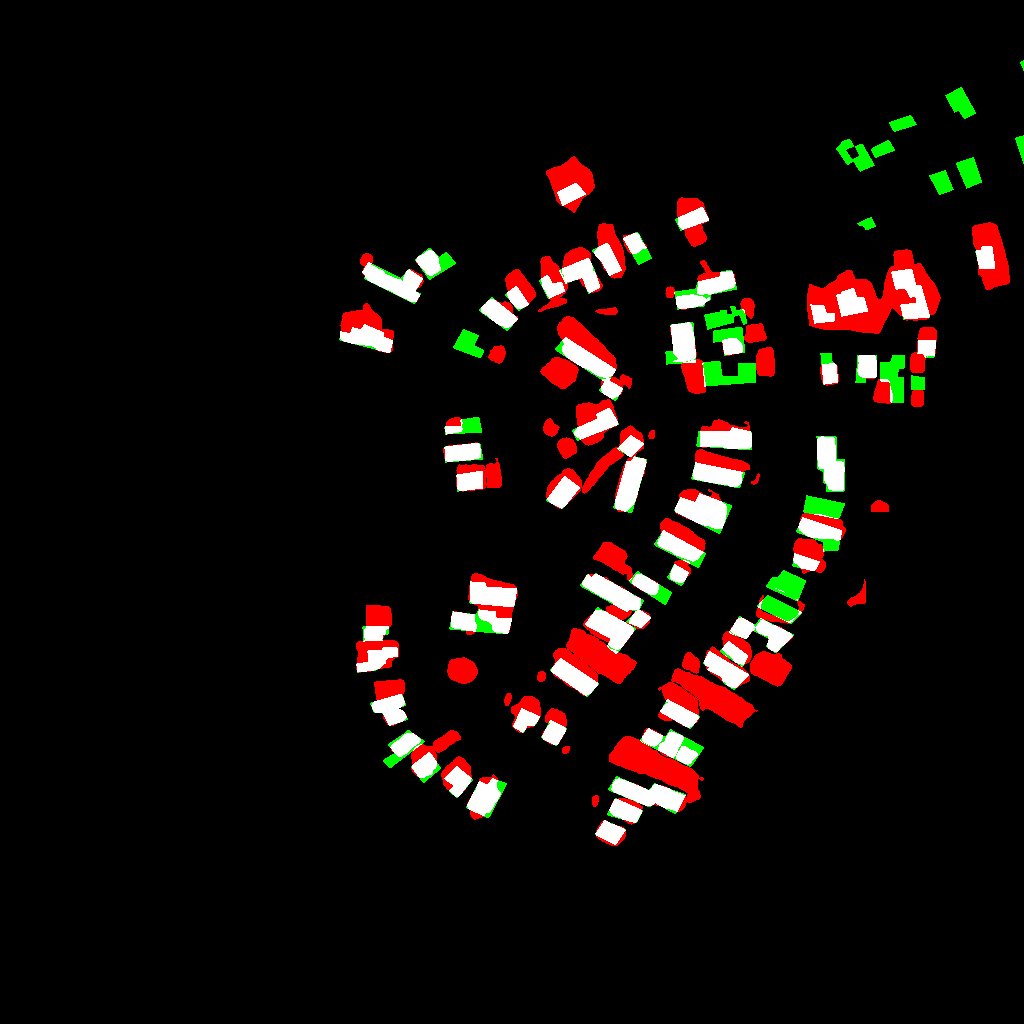}\\
        (i)&
        \includegraphics[width=1.5cm]{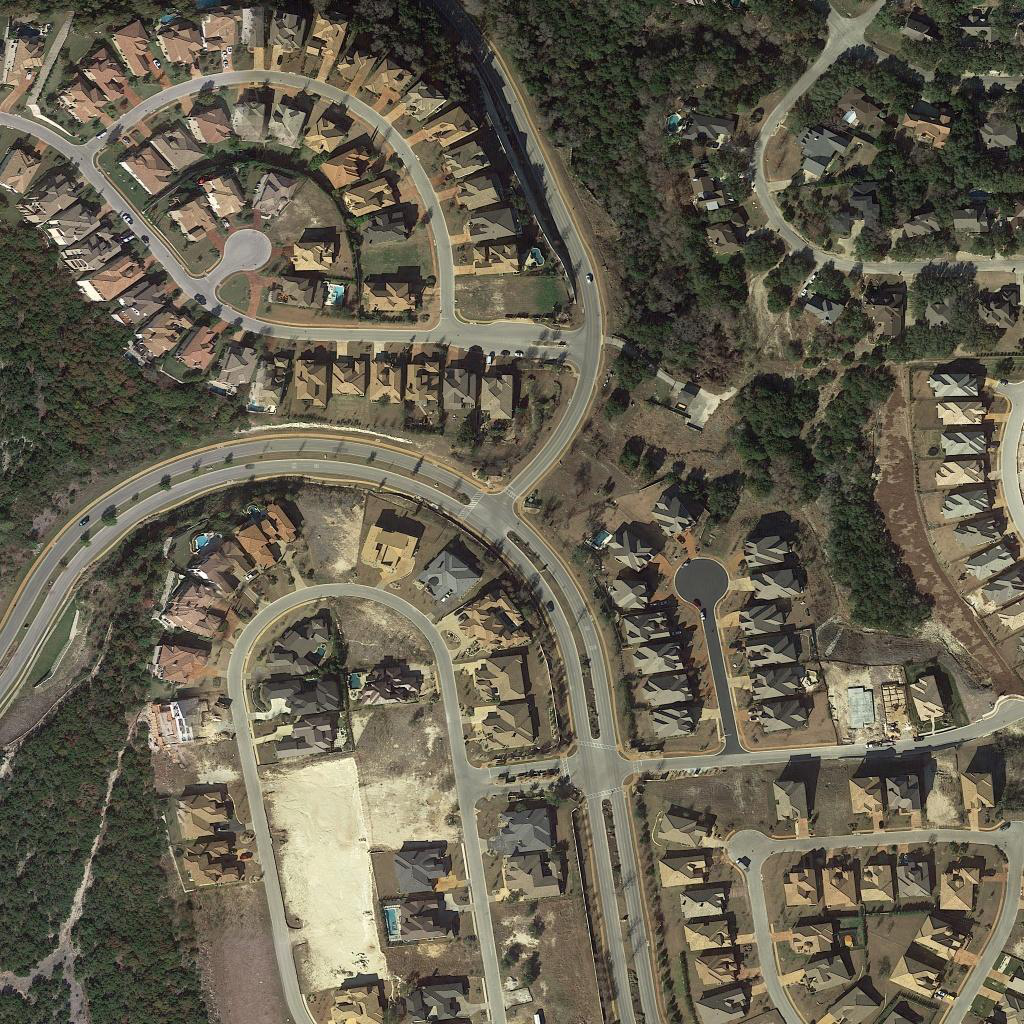} &
        \includegraphics[width=1.5cm]{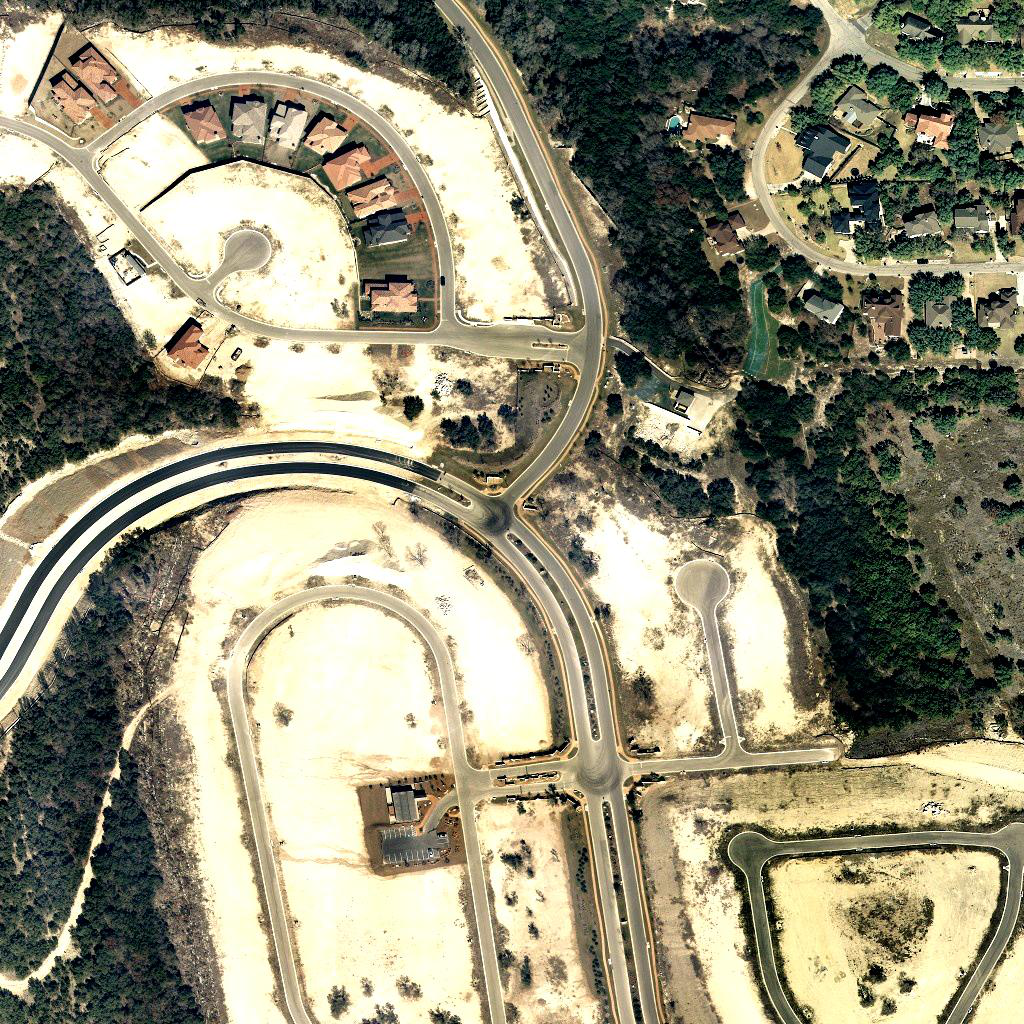} &
        \includegraphics[width=1.5cm]{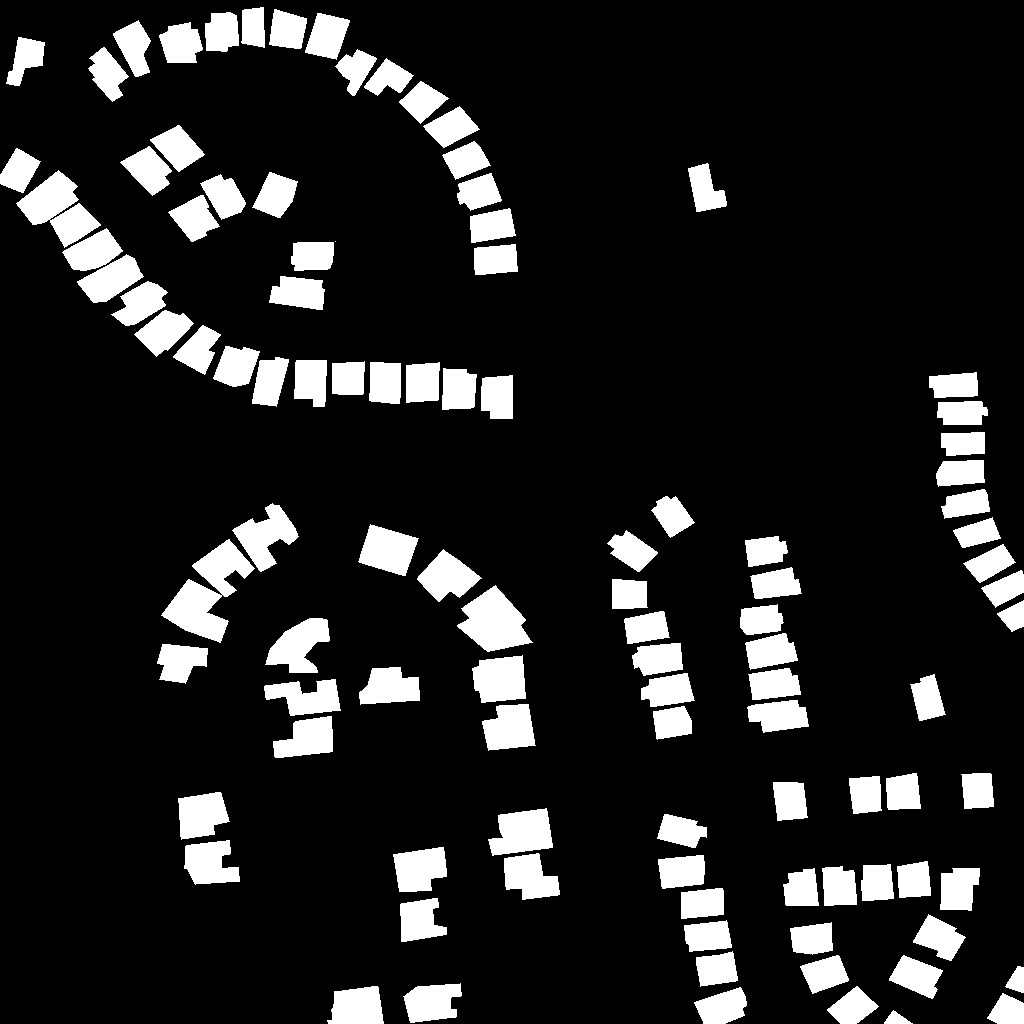} &
        \includegraphics[width=1.5cm]{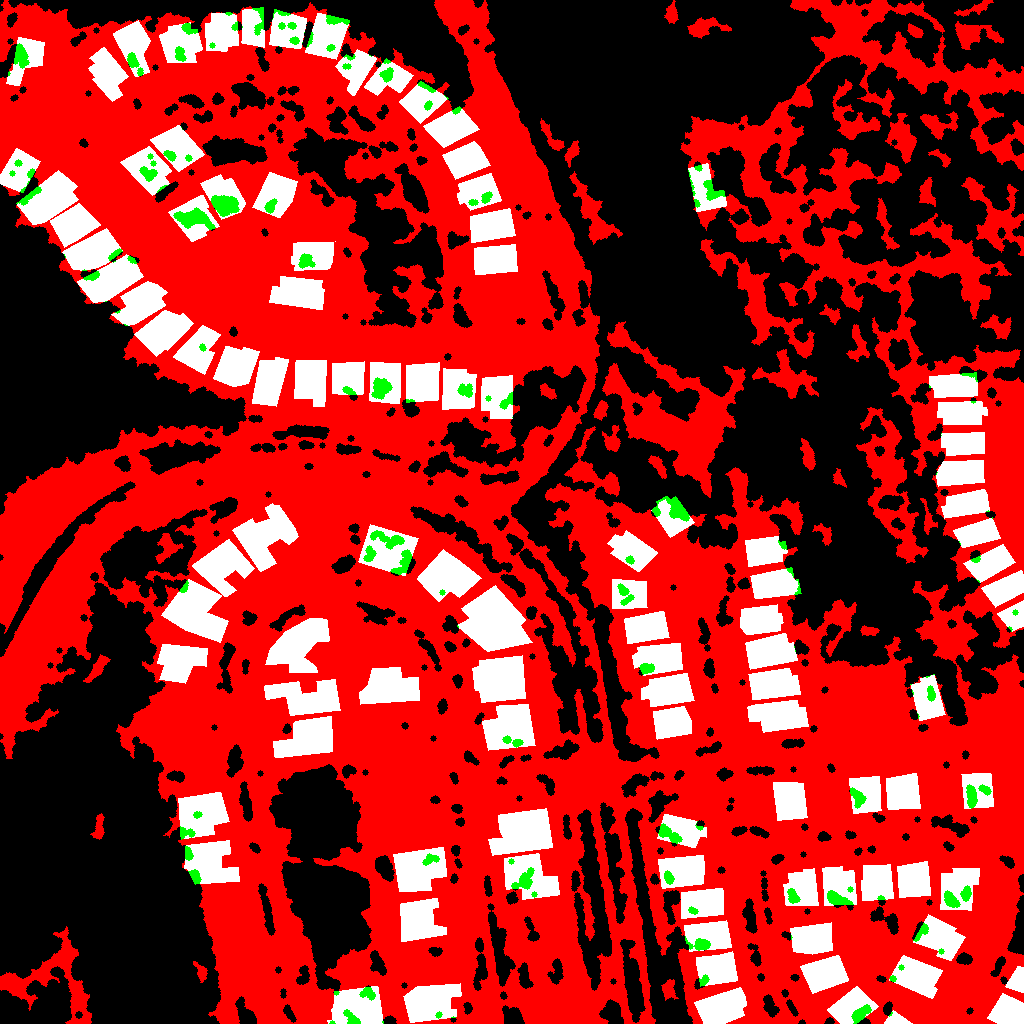} &
        \includegraphics[width=1.5cm]{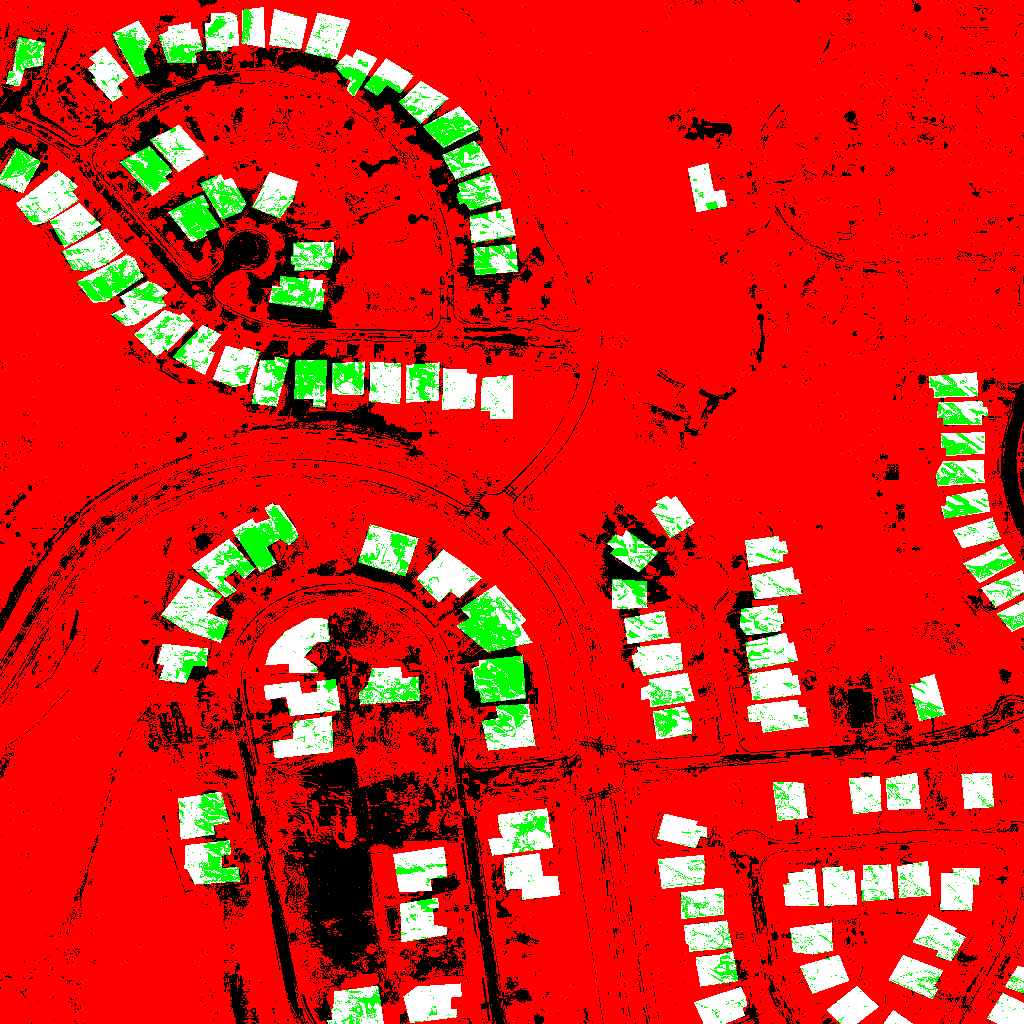} &
        \includegraphics[width=1.5cm]{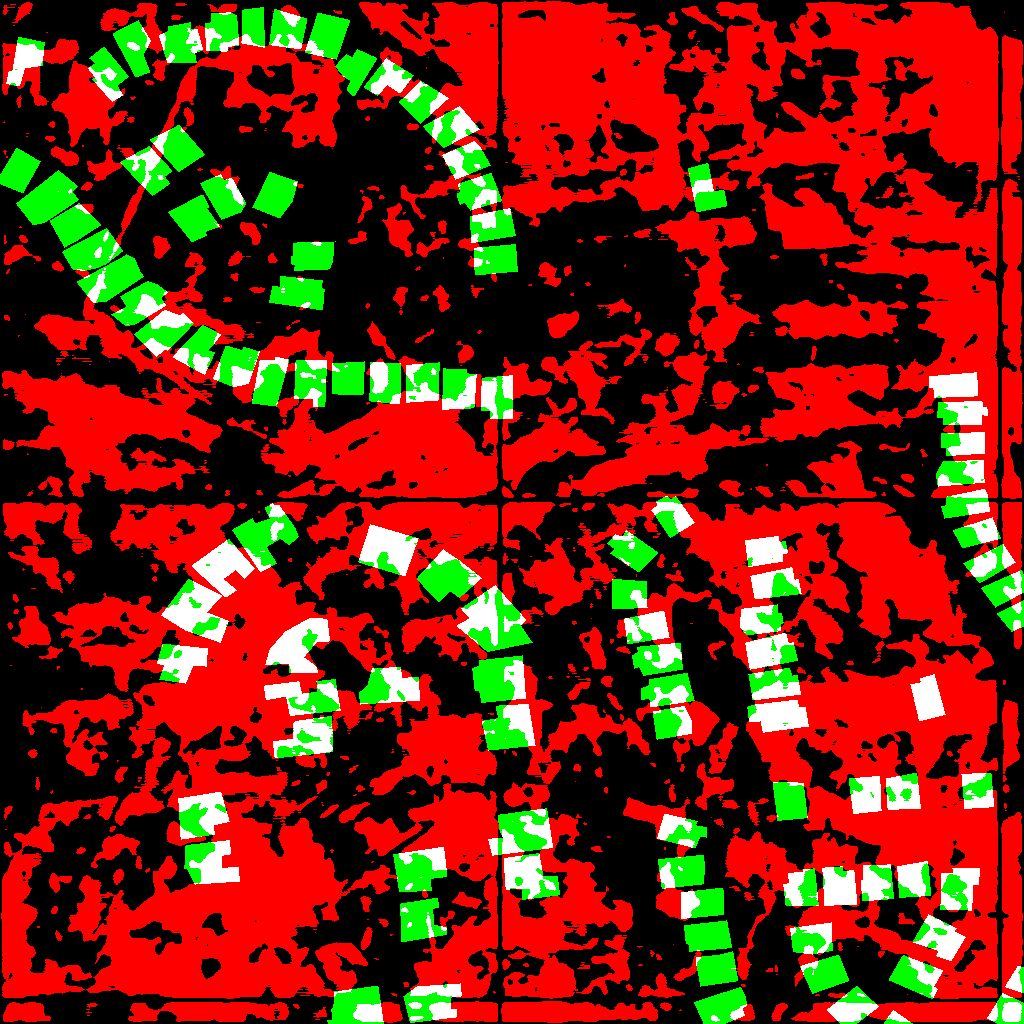} &
        \includegraphics[width=1.5cm]{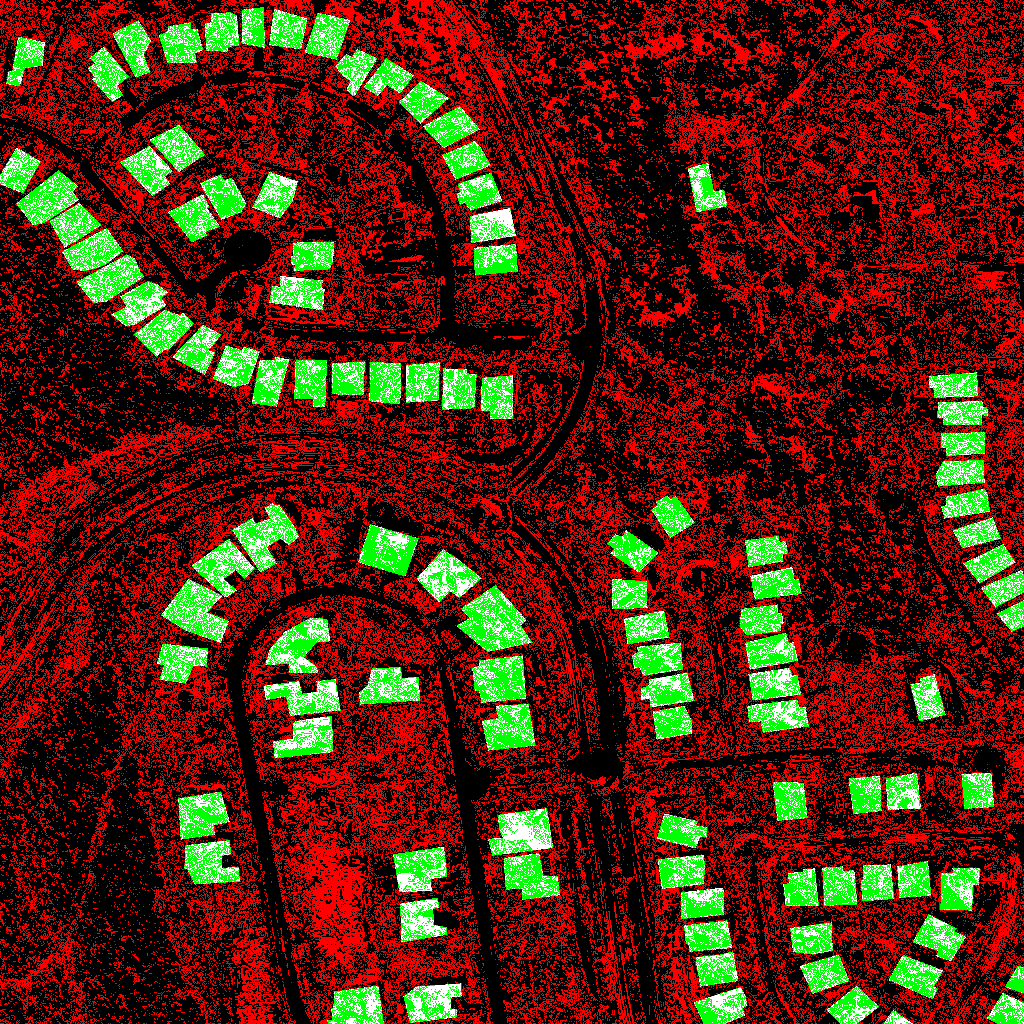} &
        \includegraphics[width=1.5cm]{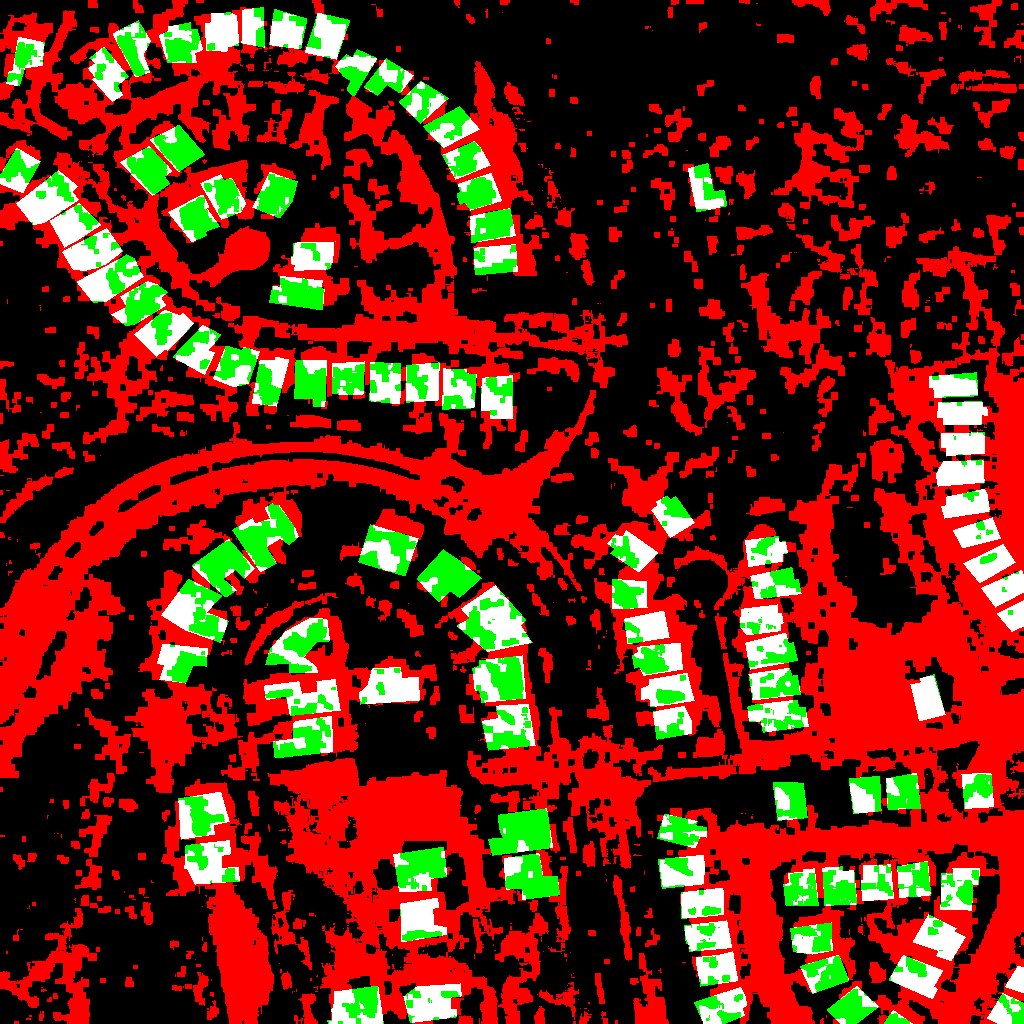}&
        \includegraphics[width=1.5cm]{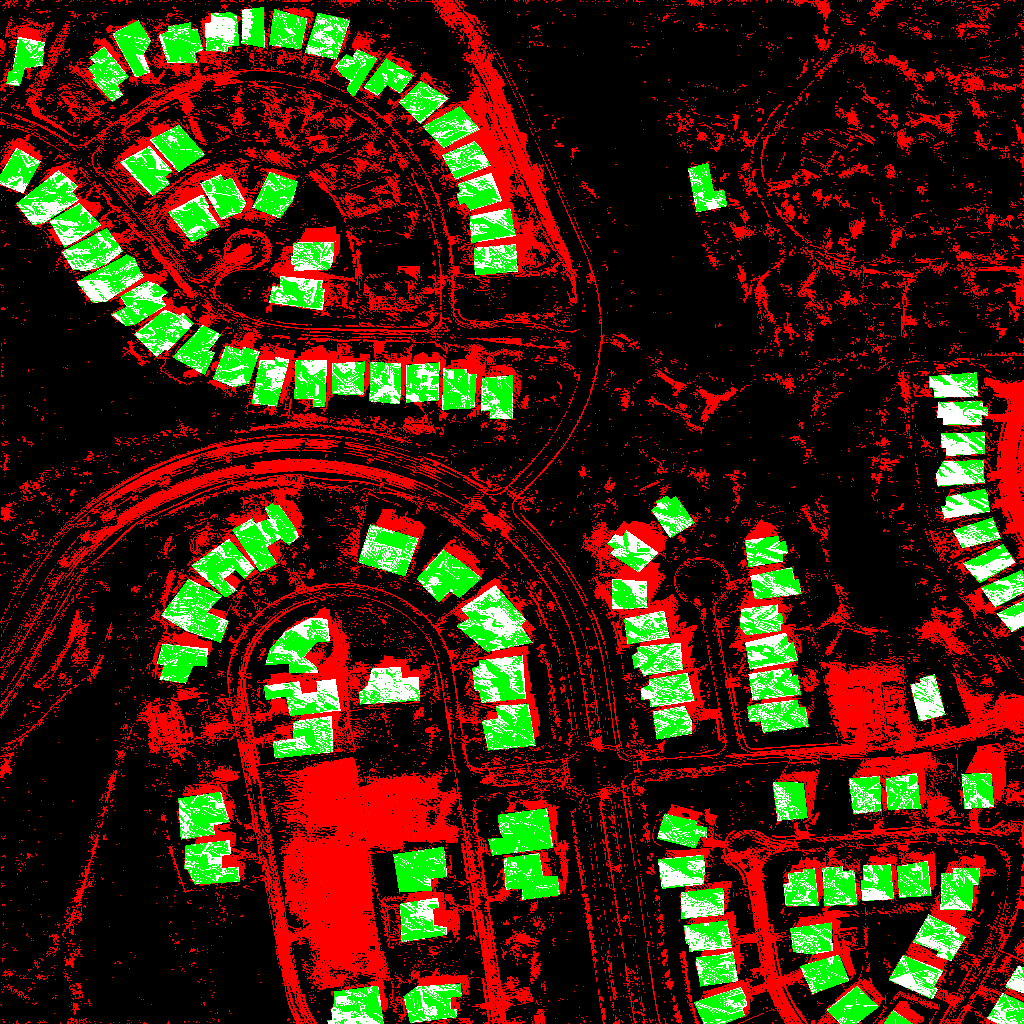}&
        \includegraphics[width=1.5cm]{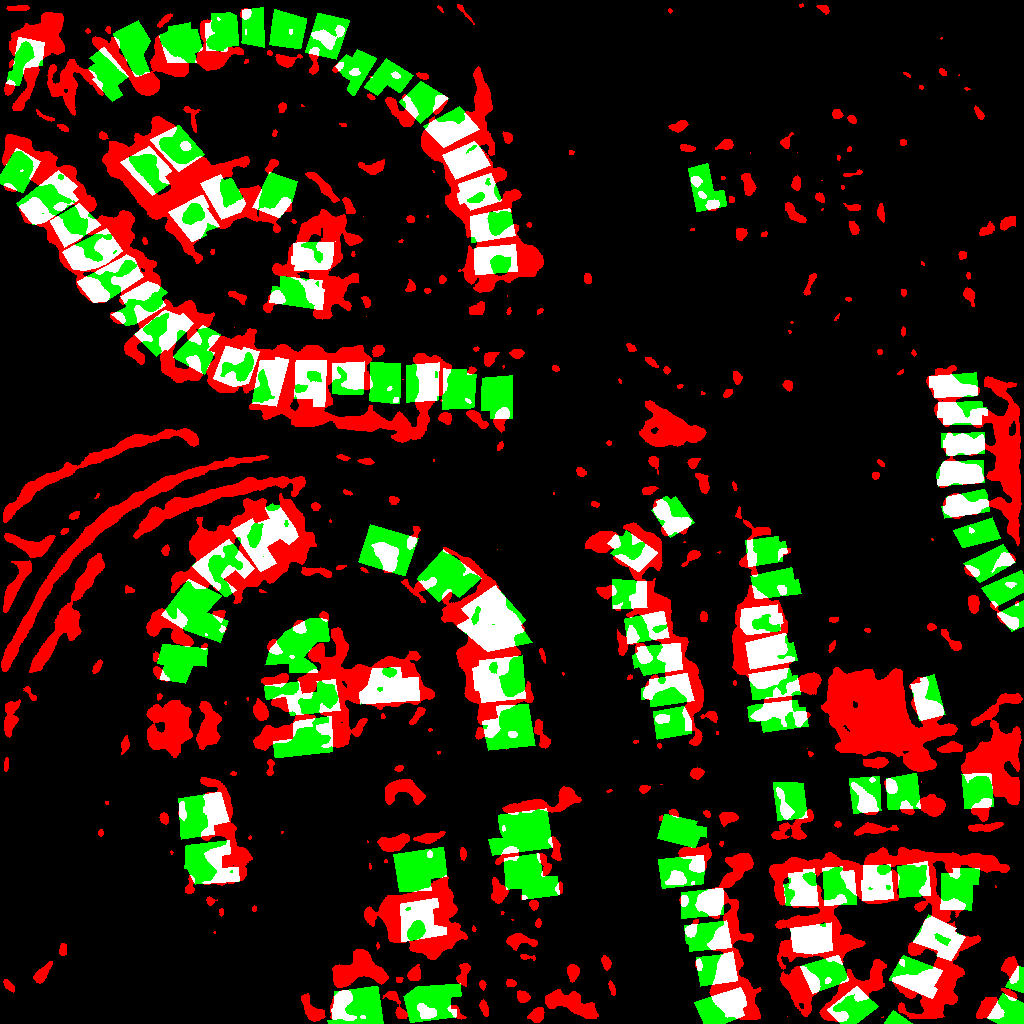}&
        \includegraphics[width=1.5cm]{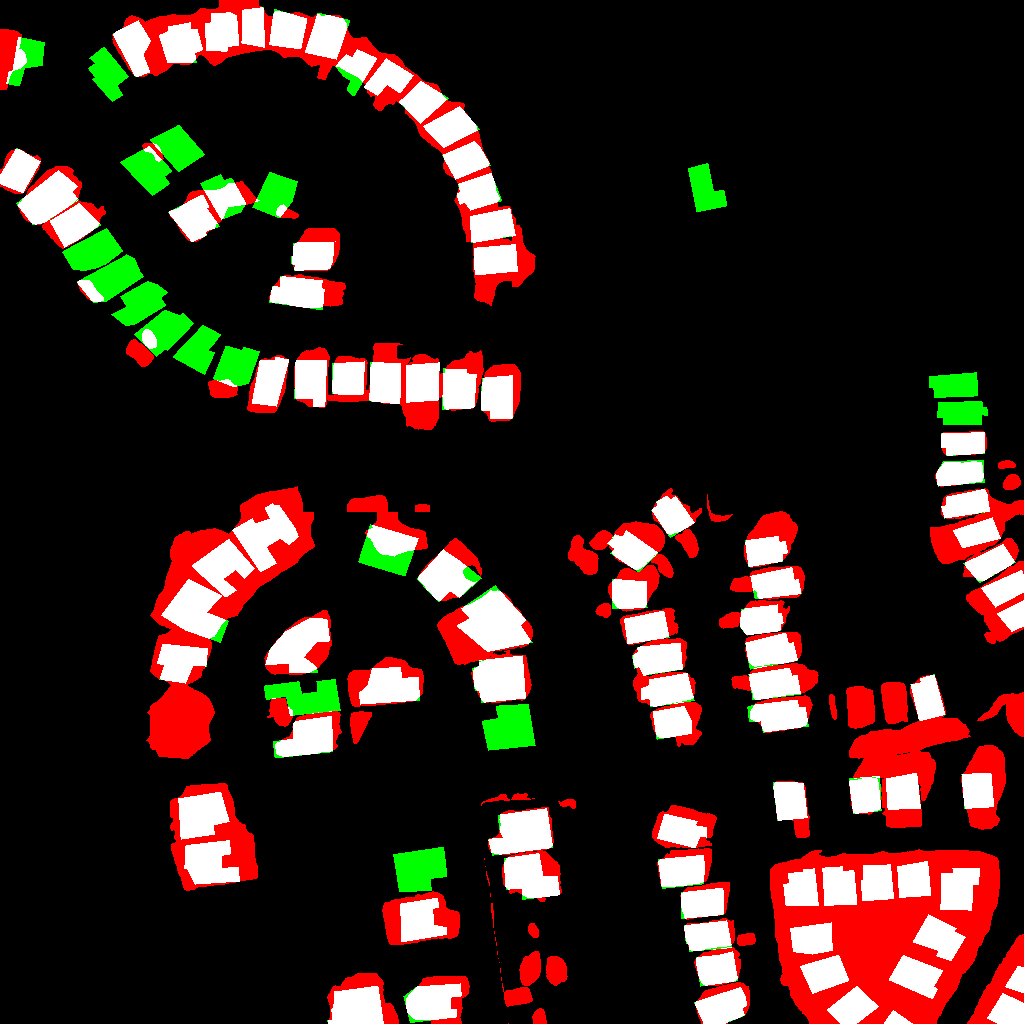}\\
        & $T_1$ image & $T_2$ image & GT & DCVA & DSFA & KPCA & CDRL & SiROC & FCD-GAN & I3PE & S2C {\small(Refined)}\\
    \end{tabular}
    \caption{Qualitative comparison between the proposed S2C framework and SOTA methods for UCD. The samples are selected from (a)-(c) CLCD, (d)-(f) SECOND, and (g)-(i) Levir datasets.}
    \label{Fig.vis_SOTA}
\end{figure*}

We conduct comparative experiments with various SOTA methods for UCD in RS. The compared methods include several non-parameter methods based on difference analysis, including the CVA-based methods \cite{Bruzzone2000diff}, ISFA \cite{Wu2014Slow}, DCVA \cite{saha2019unsupervised}, DSFA \cite{Du2019Unsupervised}, KPCA \cite{wu2021unsupervised} and SiROC \cite{Kondmann2022Spatial}. In addition, we also compare generative methods CDRL \cite{noh2022unsupervised} and FCD-GAN\cite{wu2023fully}, an augmentation-based method I3PE \cite{chen2023exchange} and a very recent VFM-based method AnyChange.\cite{zheng2024segment}. To facilitate a comprehensive assessment of accuracy, SAM-CD\cite{ding2024samcd} is also included in the comparison, representing the SOTA accuracy of supervised CD.

Table.\ref{Table.CompareSOTA} presents the quantitative results of this comparative study. Among the difference analysis-based methods, DCVA \cite{saha2019unsupervised} exhibits a notable advantage in $Rec$. It obtains the highest $Rec$ on CLCD and the second one on SECOND and Levir datasets. Among the methods presented in recent literature, I3PE achieves high $Pre$ across the three datasets. FCD-GAN \cite{wu2023fully} obtains superior $F_1$ on the CLCD dataset, and also achieves high $F_1$ on the SECOND dataset. AnyChange \cite{zheng2024segment}, a training-free approach leveraging VFM for CD, demonstrates satisfactory accuracy by achieving the highest $Rec$ on two datasets. However, it obtains a relatively low $Pre$, suggesting a considerable percentage of false alarms present in the results.

The proposed S2C yields substantial and consistent improvements in accuracy compared to the SOTA methods. The coarse prediction of S2C results in a sharp improvement of more than $30\%$ on the CLCD dataset, and approaching $20\%$ on the Levir dataset. Subsequent post-processing utilizing the IoU refinement further enhances its balance between $Pre$ and $Rec$. Notably, this refinement algorithm brings the greatest enhancement on the Levir dataset, where the coarse prediction is relatively blurred and has low $Pre$. The final results demonstrate an improvement of nearly $4\%$ in $F_1$ on the Levir dataset, which is recognized as a challenging benchmark for UCD algorithms given its building-focused annotation and the small number of changes. The improvements of the proposed S2C over the SOTA methods, after refinement, are quantified as $31\%$, $9\%$, and $23\%$ in $F_1$ for the respective datasets. The achieved level of accuracy significantly reduces the gap with fully-supervised methods, with observed reductions of $14\%$ and $16\%$ in $F_1$ on the CLCD and SECOND datasets, respectively. However, the Levir dataset still exhibits a large gap in $F_1$ between supervised and unsupervised methods, owing to its sparsity of change instances and building-focused annotations.

Fig.\ref{Fig.vis_SOTA} presents a qualitative analysis of the state-of-the-art (SOTA) methods. One can observe that most of the literature methods generate numerous false alarms due to the impact of temporal noise, such as spectral variations in Fig.\ref{Fig.vis_SOTA}(b)(c)(e)(g)(h), spatial misalignment in Fig.\ref{Fig.vis_SOTA}(d)(f), and radiometric insignificant changes in Fig.\ref{Fig.vis_SOTA}(a)(i). The results of FCD-GAN and I3PE have fewer false alarms but exhibit substantial limitations in comprehensively segmenting the major changes. In contrast, the proposed S2C demonstrates a marked reduction in errors. Its results effectively cover the majority of significant changes, although it still contains few false alarms.

\subsection{Unsupervised MMCD with S2C}

\begin{table}[t]
\centering
    \caption{Accuracy obtained by the proposed S2C with different encoders (Wuhan dataset).}
    \resizebox{1\linewidth}{!}{%
        \begin{tabular}{c|cc|c}
        \toprule
            Methods & $RGB$ Backbone & $SAR$ Backbone & $F_1$ (\%)\\
            \hline
            effi.SAM+SC & effi.SAM (vit-t) & effi.SAM (vit-t) & 9.76 \\
            effi.SAM+SC & Dino-v2 (vit-b) & Dino-v2 (vit-b) & 15.91 \\
            \hline
            S2C & effi.SAM (vit-t) & effi.SAM (vit-t) & 23.23 \\
            (dual VFMs) & Dino-v2 (vit-b) & Dino-v2 (vit-b) & 27.45 \\
            \hline
            \multirow{4}*{} & effi.SAM (vit-t) & ResNet18 & 32.04 \\
            S2C & effi.SAM (vit-s) & ResNet18 & 34.75 \\
            (single VFM)  & Dino-v2 (vit-b) & ResNet18 & 35.86 \\
             & fastSAM & ResNet34 & 38.26 \\
            \hline
            S2C & ResNet34 & ResNet34 & 41.28 \\
            (w/o. VFMs) & ResNet18 & ResNet18 & \textbf{41.96} \\
        \bottomrule
        \end{tabular} \label{Table.Het_Ablation} }
\end{table}

\begin{table}[t]
    \centering
    \caption{Quantitative evaluation of accuracy (\%) provided by SOTA unsupervised MMCD methods (Wuhan dataset).}
    \resizebox{1\linewidth}{!}{%
        \begin{tabular}{c|c|c|cccc}
        \toprule
            \multicolumn{2}{c|}{\multirow{2}*{Methods}} & \multirow{2}*{Reference} & \multicolumn{4}{c}{Accuracy}  \\
            \cline{4-7}
            \multicolumn{2}{c|}{} & &  $OA$ & $Pre$ & $Rec$ & $F_1$ \\  
            \hline
            \multirow{9}*{\rotatebox{90}{Homogeneous UCD}} & CVA & \textit{TGRS 2000} \cite{Bruzzone2000diff} & 59.81 & 13.91 & 36.12 & 20.09 \\
            & ISFA & \textit{TGRS 2014} \cite{Wu2014Slow} & 62.93 & 13.97 & 32.01 & 19.46\\
            & DCVA & \textit{TGRS 2019} \cite{saha2019unsupervised} & 47.10 & 14.29 & \underline{55.66} & 22.74\\
            & DSFA & \textit{TGRS 2019} \cite{Du2019Unsupervised} & 51.27 & 14.16 & 49.08 & 21.98\\
            & KPCA & \textit{TCYB 2021} \cite{wu2021unsupervised} & 53.47 & 13.84 & 44.52 & 21.11\\
            & CDRL & \textit{CVPR 2022} \cite{noh2022unsupervised} & 60.12 & 13.45 & 34.07 & 19.29\\
            & SiROC & \textit{TGRS 2022} \cite{Kondmann2022Spatial} & 70.09 & 15.05 & 24.52 & 18.65 \\
            & I3PE & \textit{ISPRS 2023} \cite{chen2023exchange} & 70.45 & 16.06 & 26.34 & 19.95\\
            & FCD-GAN & \textit{TPAMI 2023} \cite{wu2023fully} & \textbf{85.33} & 15.06 & 1.05 & 1.97\\
            \hline
            \multirow{4}*{\rotatebox{90}{MMCD}} & SR-GCAE & \textit{TGRS 2022} \cite{Chen2022Unsupervised} & 18.54 & 13.94 & \textbf{90.64} & 24.16\\
            & AGSCC & \textit{TNNLS 2022} \cite{Sun2022AGSCC}  & 75.48 & 10.31 & 9.79 & 10.04 \\
            & CAAE & \textit{TNNLS 2024} \cite{Luppino2024CAAE} & 75.43 & 13.71 & 14.30 & 14.00\\
            & LPEM & \textit{TNNLS 2024} \cite{sun2024LPEM} & 68.19 & \underline{19.70} & 41.43 & \underline{26.70} \\
            \hline    
            \multicolumn{2}{c|}{ S2C (w/o. VFMs) }& proposed & \underline{83.85} & \textbf{42.18} & 41.74 & \textbf{41.96} \\ 
        \bottomrule
        \end{tabular}
        }\label{Table.Compare_Het}
\end{table}

In this section, we present the experimental performance of the S2C for unsupervised MMCD. First, we conduct an ablation study to examine the efficacy of different feature encoders applied to the visible and SAR images. The results are presented in Table\ref{Table.Het_Ablation}. Due to substantial heterogeneity of SAR images relative to their training domains, employing two VFMs as feature extractors results in low accuracy. Consequently, we proceed to utilize a basic ResNet as $f_\zeta$, i.e., feature extractor for the SAR branch. Under this experimental setting, fastSAM outperforms Dino-v2 and effi.SAM as a more effective feature extractor for the optical branch. This may due to the low spatial resolution of the optical images in this dataset, where CNN-based backbones can better retain the spatial details. We proceed to examine the application of S2C without using any VFM. Surprisingly, training simple ResNet18 backbones as $f_\theta$ and $f_\zeta$ leads to the highest accuracy. This indicates that VFMs are also less effective in visible images, as this dataset has lower spatial resolution compared to the optical datasets in Table \ref{Table.Datasets}. Theoretically, utilizing RS FMs with effective generalization to both SAR and optical data could potentially enhance the accuracy of S2C. However, this hypothesis remains unexplored due to the absence of such effective FMs, particularly in the case of this specific dataset. Nonetheless, these experimental results affirm the efficacy of the S2C framework for UCD, despite the absence of the VFMs.

We further compare the accuracy of the S2C (with dual ResNet18 encoders) with the SOTA methods in Table \ref{Table.Het_Ablation}. The compared methods include not only those homogeneous UCD methods in Table \ref{Table.CompareSOTA}, but also several SOTA unsupervised MMCD methods with available implementations. The compared MMCD methods include two methods based on graph analysis: AGSCC \cite{Sun2022AGSCC} and LPEM \cite{sun2024LPEM}, a graph convolution-based method: SR-GCAE \cite{Chen2022Unsupervised} and a generative transcoding method: CAAE \cite{Luppino2024CAAE}. 

\begin{figure}[t]
\centering
    \setlength{\tabcolsep}{1pt}
    \begin{tabular}{>{\centering\arraybackslash}m{1.2cm}>{\centering\arraybackslash}m{1.2cm}>{\centering\arraybackslash}m{1.2cm}>{\centering\arraybackslash}m{1.2cm}>{\centering\arraybackslash}m{1.2cm}>{\centering\arraybackslash}m{1.2cm}>{\centering\arraybackslash}m{1.2cm}}
        \multicolumn{7}{c}{\includegraphics[width=8cm]{Vis_SOTA/ColorBar.png}}\\
        \includegraphics[width=1.2cm]{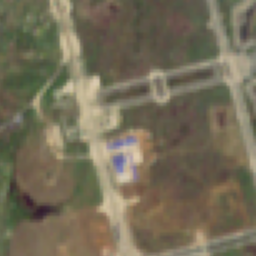} &
        \includegraphics[width=1.2cm]{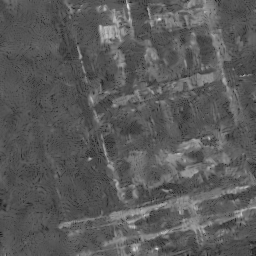} &
        \includegraphics[width=1.2cm]{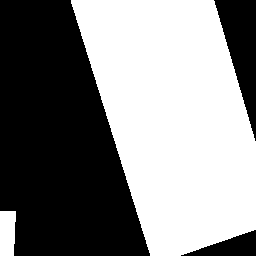} &
        \includegraphics[width=1.2cm]{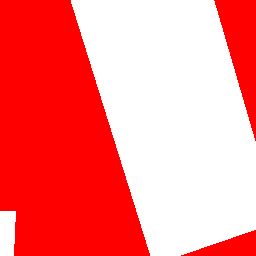} &
        \includegraphics[width=1.2cm]{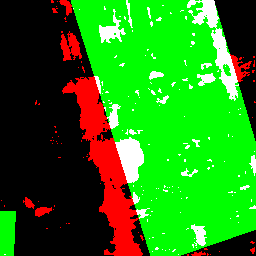} &
        \includegraphics[width=1.2cm]{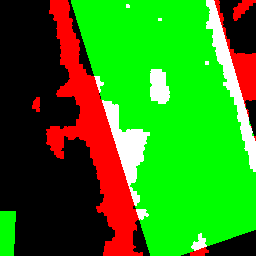} &
        \includegraphics[width=1.2cm]{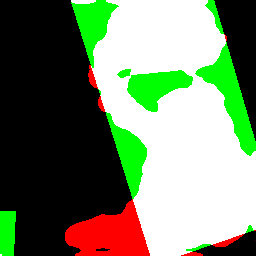}\\
        \includegraphics[width=1.2cm]{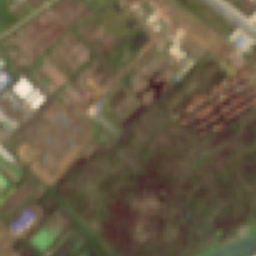} &
        \includegraphics[width=1.2cm]{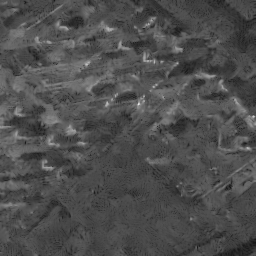} &
        \includegraphics[width=1.2cm]{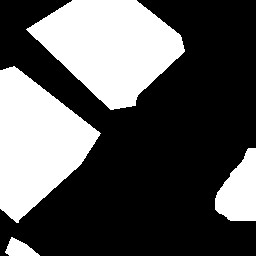} &
        \includegraphics[width=1.2cm]{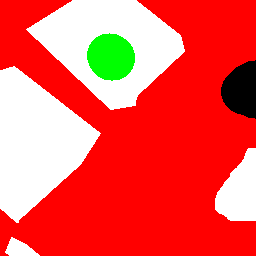} &
        \includegraphics[width=1.2cm]{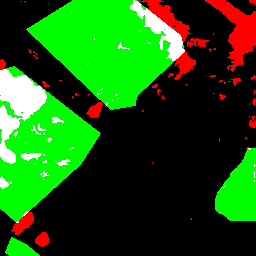} &
        \includegraphics[width=1.2cm]{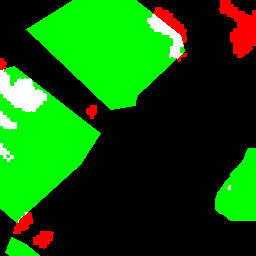} &
        \includegraphics[width=1.2cm]{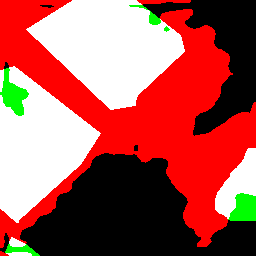}\\
        \includegraphics[width=1.2cm]{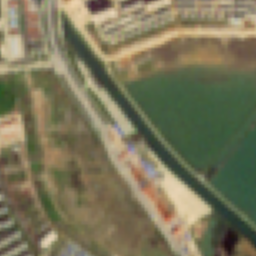} &
        \includegraphics[width=1.2cm]{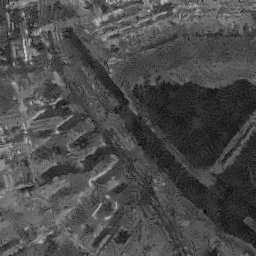} &
        \includegraphics[width=1.2cm]{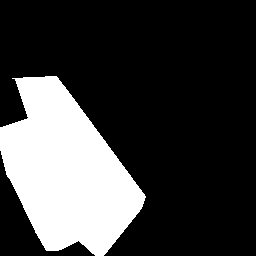} &
        \includegraphics[width=1.2cm]{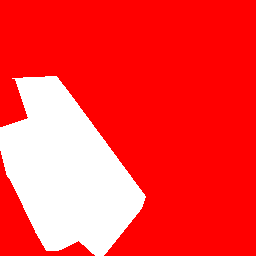} &
        \includegraphics[width=1.2cm]{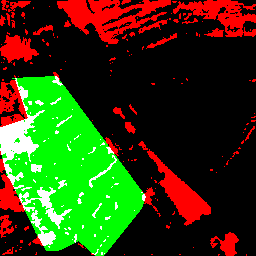} &
        \includegraphics[width=1.2cm]{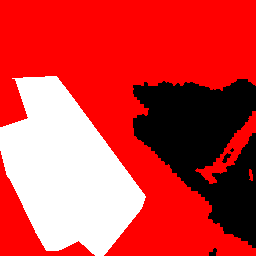} &
        \includegraphics[width=1.2cm]{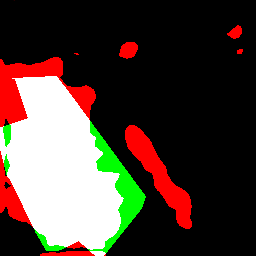}\\
        \includegraphics[width=1.2cm]{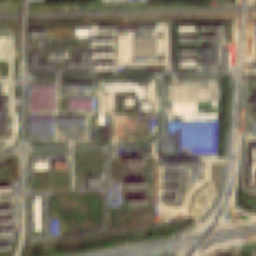} &
        \includegraphics[width=1.2cm]{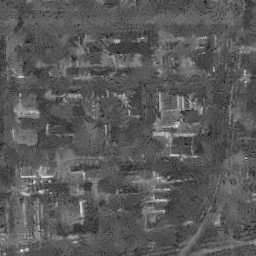} &
        \includegraphics[width=1.2cm]{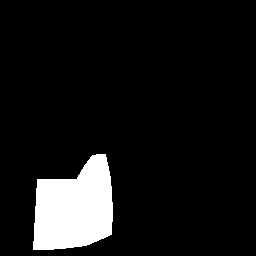} &
        \includegraphics[width=1.2cm]{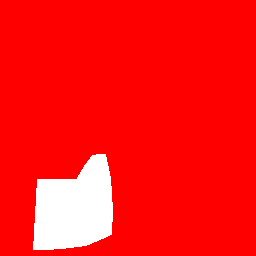} &
        \includegraphics[width=1.2cm]{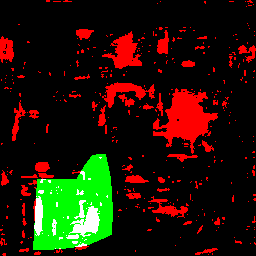} &
        \includegraphics[width=1.2cm]{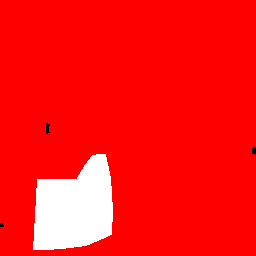} &
        \includegraphics[width=1.2cm]{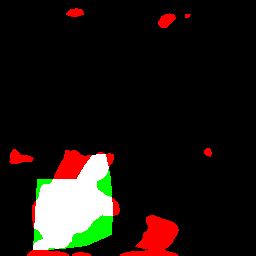}\\
        $T_1$ & $T_2$ & \multirow{2}*{GT} & SR- & \multirow{2}*{CAAE} & \multirow{2}*{LPEM} & S2C \\
        image & image & & GCAE & & & (proposed)
    \end{tabular}
    \caption{Qualitative comparison between the proposed S2C framework and SOTA methods for unsupervised MMCD.}
    \label{Fig.vis_SOTA_Het}
\end{figure}

The results indicate that homogeneous UCD methods, when applied to multimodal data, typically demonstrate reduced efficacy, with accuracy levels falling below 23\% in $F_1$. Most of the compared MMCD methods exhibit sensitivity to hyper-parameters, and fail to generalize effectively on the Wuhan heterogeneous dataset when with a substantial number of testing samples. Among these methods, the LPEM achieves the highest accuracy as indicated in $Pre$ and $F_1$. In contrast, the proposed S2C effectively identifies multimodal changes, surpassing literature methods by a margin exceeding 15\% in $F_1$. Fig.\ref{Fig.vis_SOTA_Het} present several samples of the MMCD results. One can observe that the SR-GCAE and LPEM produce considerable number of false alarms. These observations align with their evaluation metrics in Table \ref{Table.Het_Ablation} as indicated by high $Rec$ over $Pre$. In contrast, the CAAE identifies only limited areas of the changes. The proposed S2C method, although still exhibits ambiguity in delineating the object boundaries, effectively detects most of the multimodal changes.

\section{Conclusions}\label{sc5}

This study explores the formulation of a CL framework to explicitly model the unsupervised learning of semantic changes in multimodal RS images. To address the challenges posed by spectral variations, spatial misalignment, insignificant changes and multimodal heterogeneity, an S2C learning framework is developed for CD of HR RS images.
It consists of two novel CL paradigms, i.e., the CSC and CTC, which are both trained with consistency regularization to enhance robustness against different types of temporal noise. Notably, within the CTC paradigm, we present an innovative multi-temporal triplet learning strategy that addresses the existing gap in explicit difference learning. In addition, a set of novel techniques are developed to translate the VFM semantics into CD results, including grid sparsity regularization, negative cosine embedding of change probability, and an IoU matching-based refinement algorithm. Furthermore, with minimal adjustments, the S2C framework can be applied on unsupervised MMCD. Joint learning of semantic differences and consistencies enables S2C to align semantic representations across different image modalities, thus mapping domain-invariant changes. 

Experimental results reveal that the proposed S2C obtains significant accuracy improvements over the current SOTA, with advantages in $F_1$ exceeding 31\%, 9\%, 23\%, and 15\% across the four benchmark CD datasets, respectively. Additionally, S2C is efficient in training, requiring only a minimal quantity of 5 unlabeled sample pairs on the considered data set while inducing only a marginal reduction in accuracy. This highlights its potential for efficient deployment in practical applications. S2C can also be integrated into standard supervised CD methods to improve their accuracy, as it is essentially a collection of CD-specific training paradigms and post-processing algorithms that are independent of any particular DNN architectures.

A remaining limitation in S2C, as well as other UCD methods in the literature, is their lack of awareness of the particular applicational contexts. For example, S2C generates a substantial number of false alarms when applied to the Levir dataset, which focuses solely on building changes. To address this limitation, future studies are encouraged to explore language-driven UCD with the injection of user intents. Additionally, the S2C framework possesses the potential to be extended to more intricate UCD tasks, such as multi-class CD and time-series CD, which is left for further research investigations.

\bibliographystyle{IEEEtran}
\bibliography{refs}

\begin{thebibliography}{10}
\providecommand{\url}[1]{#1}
\csname url@samestyle\endcsname
\providecommand{\newblock}{\relax}
\providecommand{\bibinfo}[2]{#2}
\providecommand{\BIBentrySTDinterwordspacing}{\spaceskip=0pt\relax}
\providecommand{\BIBentryALTinterwordstretchfactor}{4}
\providecommand{\BIBentryALTinterwordspacing}{\spaceskip=\fontdimen2\font plus
\BIBentryALTinterwordstretchfactor\fontdimen3\font minus \fontdimen4\font\relax}
\providecommand{\BIBforeignlanguage}[2]{{%
\expandafter\ifx\csname l@#1\endcsname\relax
\typeout{** WARNING: IEEEtran.bst: No hyphenation pattern has been}%
\typeout{** loaded for the language `#1'. Using the pattern for}%
\typeout{** the default language instead.}%
\else
\language=\csname l@#1\endcsname
\fi
#2}}
\providecommand{\BIBdecl}{\relax}
\BIBdecl

\bibitem{bruzzone2009domain}
L.~Bruzzone and M.~Marconcini, ``Domain adaptation problems: A dasvm classification technique and a circular validation strategy,'' \emph{IEEE transactions on pattern analysis and machine intelligence}, vol.~32, no.~5, pp. 770--787, 2009.

\bibitem{wu2023fully}
C.~Wu, B.~Du, and L.~Zhang, ``Fully {{Convolutional Change Detection Framework With Generative Adversarial Network}} for {{Unsupervised}}, {{Weakly Supervised}} and {{Regional Supervised Change Detection}},'' \emph{IEEE Transactions on Pattern Analysis and Machine Intelligence}, vol.~45, no.~8, pp. 9774--9788, 2023.

\bibitem{lanza2011statistical}
A.~Lanza and L.~Di~Stefano, ``Statistical change detection by the pool adjacent violators algorithm,'' \emph{IEEE transactions on pattern analysis and machine intelligence}, vol.~33, no.~9, pp. 1894--1910, 2011.

\bibitem{taneja2015geometric}
A.~Taneja, L.~Ballan, and M.~Pollefeys, ``Geometric change detection in urban environments using images,'' \emph{IEEE transactions on pattern analysis and machine intelligence}, vol.~37, no.~11, pp. 2193--2206, 2015.

\bibitem{roth2022towards}
K.~Roth, L.~Pemula, J.~Zepeda, B.~Sch{\"o}lkopf, T.~Brox, and P.~Gehler, ``Towards total recall in industrial anomaly detection,'' in \emph{Proceedings of the IEEE/CVF conference on computer vision and pattern recognition}, 2022, pp. 14\,318--14\,328.

\bibitem{li2020siamese}
M.~D. Li, K.~Chang, B.~Bearce, C.~Y. Chang, A.~J. Huang, J.~P. Campbell, J.~M. Brown, P.~Singh, K.~V. Hoebel, D.~Erdo{\u{g}}mu{\c{s}} \emph{et~al.}, ``Siamese neural networks for continuous disease severity evaluation and change detection in medical imaging,'' \emph{NPJ digital medicine}, vol.~3, no.~1, p.~48, 2020.

\bibitem{bruzzone2012novel}
L.~Bruzzone and F.~Bovolo, ``A novel framework for the design of change-detection systems for very-high-resolution remote sensing images,'' \emph{Proceedings of the IEEE}, vol. 101, no.~3, pp. 609--630, 2012.

\bibitem{benedek2011building}
C.~Benedek, X.~Descombes, and J.~Zerubia, ``Building development monitoring in multitemporal remotely sensed image pairs with stochastic birth-death dynamics,'' \emph{IEEE Transactions on Pattern Analysis and Machine Intelligence}, vol.~34, no.~1, pp. 33--50, 2011.

\bibitem{robin2010contrario}
A.~Robin, L.~Moisan, and S.~Le~H{\'e}garat-Mascle, ``An a-contrario approach for subpixel change detection in satellite imagery,'' \emph{IEEE Transactions on pattern analysis and machine intelligence}, vol.~32, no.~11, pp. 1977--1993, 2010.

\bibitem{brunner2010earthquake}
D.~Brunner, G.~Lemoine, and L.~Bruzzone, ``Earthquake damage assessment of buildings using vhr optical and sar imagery,'' \emph{IEEE Transactions on Geoscience and Remote Sensing}, vol.~48, no.~5, pp. 2403--2420, 2010.

\bibitem{chen2021remote}
H.~Chen, Z.~Qi, and Z.~Shi, ``Remote sensing image change detection with transformers,'' \emph{IEEE Transactions on Geoscience and Remote Sensing}, vol.~60, pp. 1--14, 2021.

\bibitem{ding2024samcd}
L.~Ding, K.~Zhu, D.~Peng, H.~Tang, K.~Yang, and L.~Bruzzone, ``Adapting segment anything model for change detection in hr remote sensing images,'' \emph{IEEE Transactions on Geoscience and Remote Sensing}, vol.~62, pp. 1--11, 2024.

\bibitem{chen2021self}
Y.~Chen and L.~Bruzzone, ``Self-supervised change detection in multiview remote sensing images,'' \emph{IEEE Transactions on Geoscience and Remote Sensing}, vol.~60, pp. 1--12, 2021.

\bibitem{hong2018augmented}
D.~Hong, N.~Yokoya, J.~Chanussot, and X.~X. Zhu, ``An augmented linear mixing model to address spectral variability for hyperspectral unmixing,'' \emph{IEEE Transactions on Image Processing}, vol.~28, no.~4, pp. 1923--1938, 2018.

\bibitem{sohn2020fixmatch}
K.~Sohn, D.~Berthelot, N.~Carlini, Z.~Zhang, H.~Zhang, C.~A. Raffel, E.~D. Cubuk, A.~Kurakin, and C.-L. Li, ``Fixmatch: Simplifying semi-supervised learning with consistency and confidence,'' \emph{Advances in neural information processing systems}, vol.~33, pp. 596--608, 2020.

\bibitem{Kirillov2023Segment}
A.~Kirillov, E.~Mintun, N.~Ravi, H.~Mao, C.~Rolland, L.~Gustafson, T.~Xiao, S.~Whitehead, A.~C. Berg, W.-Y. Lo, P.~Dollar, and R.~Girshick, ``Segment anything,'' in \emph{Proceedings of the IEEE/CVF International Conference on Computer Vision (ICCV)}, October 2023, pp. 4015--4026.

\bibitem{yang2023revisiting}
L.~Yang, L.~Qi, L.~Feng, W.~Zhang, and Y.~Shi, ``Revisiting {{Weak-to-Strong Consistency}} in {{Semi-Supervised Semantic Segmentation}},'' in \emph{2023 {{IEEE}}/{{CVF Conference}} on {{Computer Vision}} and {{Pattern Recognition}} ({{CVPR}})}.\hskip 1em plus 0.5em minus 0.4em\relax IEEE, pp. 7236--7246.

\bibitem{chen2022self}
Y.~Chen and L.~Bruzzone, ``A self-supervised approach to pixel-level change detection in bi-temporal rs images,'' \emph{IEEE Transactions on Geoscience and Remote Sensing}, vol.~60, pp. 1--11, 2022.

\bibitem{zheng2024segment}
Z.~Zheng, Y.~Zhong, L.~Zhang, and S.~Ermon, ``Segment any change,'' \emph{Advances in Neural Information Processing Systems}, 2024.

\bibitem{ji2024segment}
W.~Ji, J.~Li, Q.~Bi, T.~Liu, W.~Li, and L.~Cheng, ``Segment anything is not always perfect: An investigation of sam on different real-world applications,'' pp. 617--630, 2024.

\bibitem{Bruzzone2000diff}
L.~{Bruzzone} and D.~F. {Prieto}, ``Automatic analysis of the difference image for unsupervised change detection,'' \emph{IEEE Transactions on Geoscience and Remote Sensing}, vol.~38, no.~3, pp. 1171--1182, 2000.

\bibitem{Gao2016Automatic}
F.~Gao, J.~Dong, B.~Li, and Q.~Xu, ``Automatic change detection in synthetic aperture radar images based on pcanet,'' \emph{IEEE Geoscience and Remote Sensing Letters}, vol.~13, no.~12, pp. 1792--1796, 2016.

\bibitem{saha2019unsupervised}
S.~Saha, F.~Bovolo, and L.~Bruzzone, ``Unsupervised deep change vector analysis for multiple-change detection in vhr images,'' \emph{IEEE Transactions on Geoscience and Remote Sensing}, vol.~57, no.~6, pp. 3677--3693, 2019.

\bibitem{wu2021unsupervised}
C.~Wu, H.~Chen, B.~Du, and L.~Zhang, ``Unsupervised change detection in multitemporal vhr images based on deep kernel pca convolutional mapping network,'' \emph{IEEE Transactions on Cybernetics}, vol.~52, no.~11, pp. 12\,084--12\,098, 2022.

\bibitem{Wu2014Slow}
C.~Wu, B.~Du, and L.~Zhang, ``Slow feature analysis for change detection in multispectral imagery,'' \emph{IEEE Transactions on Geoscience and Remote Sensing}, vol.~52, no.~5, pp. 2858--2874, 2014.

\bibitem{Du2019Unsupervised}
B.~Du, L.~Ru, C.~Wu, and L.~Zhang, ``Unsupervised deep slow feature analysis for change detection in multi-temporal remote sensing images,'' \emph{IEEE Transactions on Geoscience and Remote Sensing}, vol.~57, no.~12, pp. 9976--9992, 2019.

\bibitem{Kondmann2022Spatial}
L.~Kondmann, A.~Toker, S.~Saha, B.~Schölkopf, L.~Leal-Taixé, and X.~X. Zhu, ``Spatial context awareness for unsupervised change detection in optical satellite images,'' \emph{IEEE Transactions on Geoscience and Remote Sensing}, vol.~60, pp. 1--15, 2022.

\bibitem{Chen2022Unsupervised}
H.~Chen, N.~Yokoya, C.~Wu, and B.~Du, ``Unsupervised multimodal change detection based on structural relationship graph representation learning,'' \emph{IEEE Transactions on Geoscience and Remote Sensing}, vol.~60, pp. 1--18, 2022.

\bibitem{noh2022unsupervised}
H.~Noh, J.~Ju, M.~Seo, J.~Park, and D.-G. Choi, ``Unsupervised change detection based on image reconstruction loss,'' in \emph{Proceedings of the IEEE/CVF Conference on Computer Vision and Pattern Recognition}, 2022, pp. 1352--1361.

\bibitem{hong2023cross}
D.~Hong, B.~Zhang, H.~Li, Y.~Li, J.~Yao, C.~Li, M.~Werner, J.~Chanussot, A.~Zipf, and X.~X. Zhu, ``Cross-city matters: A multimodal remote sensing benchmark dataset for cross-city semantic segmentation using high-resolution domain adaptation networks,'' \emph{Remote Sensing of Environment}, vol. 299, p. 113856, 2023.

\bibitem{liu2019HyperCDReview}
S.~Liu, D.~Marinelli, L.~Bruzzone, and F.~Bovolo, ``A review of change detection in multitemporal hyperspectral images: Current techniques, applications, and challenges,'' \emph{IEEE Geoscience and Remote Sensing Magazine}, vol.~7, no.~2, pp. 140--158, 2019.

\bibitem{liu2018coupling}
J.~Liu, M.~Gong, K.~Qin, and P.~Zhang, ``A deep convolutional coupling network for change detection based on heterogeneous optical and radar images,'' \emph{IEEE Transactions on Neural Networks and Learning Systems}, vol.~29, no.~3, pp. 545--559, 2018.

\bibitem{wu2022CommonalityAE}
Y.~Wu, J.~Li, Y.~Yuan, A.~K. Qin, Q.-G. Miao, and M.-G. Gong, ``Commonality autoencoder: Learning common features for change detection from heterogeneous images,'' \emph{IEEE Transactions on Neural Networks and Learning Systems}, vol.~33, no.~9, pp. 4257--4270, 2022.

\bibitem{saha2021building}
S.~Saha, F.~Bovolo, and L.~Bruzzone, ``Building {{Change Detection}} in {{VHR SAR Images}} via {{Unsupervised Deep Transcoding}},'' \emph{IEEE Transactions on Geoscience and Remote Sensing}, vol.~59, no.~3, pp. 1917--1929.

\bibitem{Luppino2024CAAE}
L.~T. Luppino, M.~A. Hansen, M.~Kampffmeyer, F.~M. Bianchi, G.~Moser, R.~Jenssen, and S.~N. Anfinsen, ``Code-aligned autoencoders for unsupervised change detection in multimodal remote sensing images,'' \emph{IEEE Transactions on Neural Networks and Learning Systems}, vol.~35, no.~1, pp. 60--72, 2024.

\bibitem{sun2021itertivegraph}
Y.~Sun, L.~Lei, D.~Guan, and G.~Kuang, ``Iterative robust graph for unsupervised change detection of heterogeneous remote sensing images,'' \emph{IEEE Transactions on Image Processing}, vol.~30, pp. 6277--6291, 2021.

\bibitem{sun2024LPEM}
Y.~Sun, L.~Lei, D.~Guan, G.~Kuang, Z.~Li, and L.~Liu, ``Locality preservation for unsupervised multimodal change detection in remote sensing imagery,'' \emph{IEEE Transactions on Neural Networks and Learning Systems}, pp. 1--15, 2024.

\bibitem{prendes2015Multivariate}
J.~Prendes, M.~Chabert, F.~Pascal, A.~Giros, and J.-Y. Tourneret, ``A new multivariate statistical model for change detection in images acquired by homogeneous and heterogeneous sensors,'' \emph{IEEE Transactions on Image Processing}, vol.~24, no.~3, pp. 799--812, 2015.

\bibitem{touati2019multimodal}
R.~Touati, M.~Mignotte, and M.~Dahmane, ``Multimodal change detection in remote sensing images using an unsupervised pixel pairwise-based markov random field model,'' \emph{IEEE Transactions on Image Processing}, vol.~29, pp. 757--767, 2019.

\bibitem{lv2022cdhetreview}
Z.~Lv, H.~Huang, X.~Li, M.~Zhao, J.~A. Benediktsson, W.~Sun, and N.~Falco, ``Land cover change detection with heterogeneous remote sensing images: Review, progress, and perspective,'' \emph{Proceedings of the IEEE}, vol. 110, no.~12, pp. 1976--1991, 2022.

\bibitem{bandara2022revisiting}
\BIBentryALTinterwordspacing
W.~G.~C. Bandara and V.~M. Patel. Revisiting {{Consistency Regularization}} for {{Semi-supervised Change Detection}} in {{Remote Sensing Images}}. [Online]. Available: \url{http://arxiv.org/abs/2204.08454}
\BIBentrySTDinterwordspacing

\bibitem{mall2023changeaware}
U.~Mall, B.~Hariharan, and K.~Bala, ``Change-{{Aware Sampling}} and {{Contrastive Learning}} for {{Satellite Images}},'' in \emph{2023 {{IEEE}}/{{CVF Conference}} on {{Computer Vision}} and {{Pattern Recognition}} ({{CVPR}})}.\hskip 1em plus 0.5em minus 0.4em\relax IEEE, pp. 5261--5270.

\bibitem{zhao2024pixellevel}
M.~Zhao, X.~Hu, L.~Zhang, Q.~Meng, Y.~Chen, and L.~Bruzzone, ``Beyond {{Pixel-Level Annotation}}: {{Exploring Self-Supervised Learning}} for {{Change Detection With Image-Level Supervision}},'' \emph{IEEE Transactions on Geoscience and Remote Sensing}, vol.~62, pp. 1--16.

\bibitem{radford2021CLIP}
A.~Radford, J.~W. Kim, C.~Hallacy, A.~Ramesh, G.~Goh, S.~Agarwal, G.~Sastry, A.~Askell, P.~Mishkin, J.~Clark \emph{et~al.}, ``Learning transferable visual models from natural language supervision,'' in \emph{International conference on machine learning}.\hskip 1em plus 0.5em minus 0.4em\relax PMLR, 2021, pp. 8748--8763.

\bibitem{hong2024spectralgpt}
D.~Hong, B.~Zhang, X.~Li, Y.~Li, C.~Li, J.~Yao, N.~Yokoya, H.~Li, P.~Ghamisi, X.~Jia, A.~Plaza, P.~Gamba, J.~A. Benediktsson, and J.~Chanussot, ``Spectralgpt: Spectral remote sensing foundation model,'' \emph{IEEE Transactions on Pattern Analysis and Machine Intelligence}, 2024, dOI:10.1109/TPAMI.2024.3362475.

\bibitem{guo2024skysense}
X.~Guo, J.~Lao, B.~Dang, Y.~Zhang, L.~Yu, L.~Ru, L.~Zhong, Z.~Huang, K.~Wu, D.~Hu \emph{et~al.}, ``Skysense: A multi-modal remote sensing foundation model towards universal interpretation for earth observation imagery,'' in \emph{Proceedings of the IEEE/CVF Conference on Computer Vision and Pattern Recognition}, 2024, pp. 27\,672--27\,683.

\bibitem{wang2023cs}
L.~Wang, M.~Zhang, and W.~Shi, ``Cs-wscdnet: Class activation mapping and segment anything model-based framework for weakly supervised change detection,'' \emph{IEEE Transactions on Geoscience and Remote Sensing}, 2023.

\bibitem{chen2024change}
H.~Chen, J.~Song, and N.~Yokoya, ``Change detection between optical remote sensing imagery and map data via segment anything model (sam),'' \emph{arXiv preprint arXiv:2401.09019}, 2024.

\bibitem{dong2024changeclip}
S.~Dong, L.~Wang, B.~Du, and X.~Meng, ``Changeclip: Remote sensing change detection with multimodal vision-language representation learning,'' \emph{ISPRS Journal of Photogrammetry and Remote Sensing}, vol. 208, pp. 53--69, 2024.

\bibitem{hu2021lora}
\BIBentryALTinterwordspacing
E.~J. Hu, Y.~Shen, P.~Wallis, Z.~Allen-Zhu, Y.~Li, S.~Wang, L.~Wang, and W.~Chen. {{LoRA}}: {{Low-Rank Adaptation}} of {{Large Language Models}}. [Online]. Available: \url{http://arxiv.org/abs/2106.09685}
\BIBentrySTDinterwordspacing

\bibitem{chen2022selfsupervised}
Y.~Chen and L.~Bruzzone, ``A {{Self-Supervised Approach}} to {{Pixel-Level Change Detection}} in {{Bi-Temporal RS Images}},'' \emph{IEEE Transactions on Geoscience and Remote Sensing}, vol.~60, pp. 1--11.

\bibitem{bandara2023deep}
W.~G.~C. Bandara and V.~M. Patel, ``Deep metric learning for unsupervised remote sensing change detection,'' \emph{arXiv preprint arXiv:2303.09536}, 2023.

\bibitem{deuser2023sample4geo}
F.~Deuser, K.~Habel, and N.~Oswald, ``Sample4geo: Hard negative sampling for cross-view geo-localisation,'' in \emph{Proceedings of the IEEE/CVF International Conference on Computer Vision}, 2023, pp. 16\,847--16\,856.

\bibitem{liu2022cnntransformer}
M.~Liu, Z.~Chai, H.~Deng, and R.~Liu, ``A {{CNN-Transformer Network With Multiscale Context Aggregation}} for {{Fine-Grained Cropland Change Detection}},'' vol.~15, pp. 4297--4306.

\bibitem{yang2022asymmetric}
K.~Yang, G.-S. Xia, Z.~Liu, B.~Du, W.~Yang, M.~Pelillo, and L.~Zhang, ``Asymmetric {{Siamese Networks}} for {{Semantic Change Detection}} in {{Aerial Images}},'' \emph{IEEE Transactions on Geoscience and Remote Sensing}, vol.~60, pp. 1--18.

\bibitem{chen2020spatialtemporal}
H.~Chen and Z.~Shi, ``A {{Spatial-Temporal Attention-Based Method}} and a {{New Dataset}} for {{Remote Sensing Image Change Detection}},'' vol.~12, no.~10, p. 1662.

\bibitem{zhang2022domain}
C.~Zhang, Y.~Feng, L.~Hu, D.~Tapete, L.~Pan, Z.~Liang, F.~Cigna, and P.~Yue, ``A domain adaptation neural network for change detection with heterogeneous optical and sar remote sensing images,'' \emph{International Journal of Applied Earth Observation and Geoinformation}, vol. 109, p. 102769, 2022.

\bibitem{chen2023exchange}
H.~Chen, J.~Song, C.~Wu, B.~Du, and N.~Yokoya, ``Exchange means change: An unsupervised single-temporal change detection framework based on intra-and inter-image patch exchange,'' \emph{ISPRS Journal of Photogrammetry and Remote Sensing}, vol. 206, pp. 87--105, 2023.

\bibitem{ding2021adversarial}
L.~Ding, H.~Tang, Y.~Liu, Y.~Shi, X.~X. Zhu, and L.~Bruzzone, ``Adversarial shape learning for building extraction in vhr remote sensing images,'' \emph{IEEE Transactions on Image Processing}, vol.~31, pp. 678--690, 2021.

\bibitem{xiong2024efficientsam}
Y.~Xiong, B.~Varadarajan, L.~Wu, X.~Xiang, F.~Xiao, C.~Zhu, X.~Dai, D.~Wang, F.~Sun, F.~Iandola, R.~Krishnamoorthi, and V.~Chandra, ``{{EfficientSAM}}: {{Leveraged Masked Image Pretraining}} for {{Efficient Segment Anything}},'' in \emph{2024 {{IEEE}}/{{CVF Conference}} on {{Computer Vision}} and {{Pattern Recognition}} ({{CVPR}})}.\hskip 1em plus 0.5em minus 0.4em\relax IEEE, pp. 16\,111--16\,121.

\bibitem{he2016resnet}
K.~He, X.~Zhang, S.~Ren, and J.~Sun, ``Deep residual learning for image recognition,'' in \emph{Proceedings of the IEEE conference on computer vision and pattern recognition}, 2016, pp. 770--778.

\bibitem{zhao2023fast}
\BIBentryALTinterwordspacing
X.~Zhao, W.~Ding, Y.~An, Y.~Du, T.~Yu, M.~Li, M.~Tang, and J.~Wang. Fast {{Segment Anything}}. [Online]. Available: \url{http://arxiv.org/abs/2306.12156}
\BIBentrySTDinterwordspacing

\bibitem{oquab2024dinov2}
M.~Oquab, T.~Darcet, T.~Moutakanni, H.~V. Vo, M.~Szafraniec, V.~Khalidov, P.~Fernandez, D.~Haziza, F.~Massa, A.~El-Nouby, R.~Howes, P.-Y. Huang, H.~Xu, V.~Sharma, S.-W. Li, W.~Galuba, M.~Rabbat, M.~Assran, N.~Ballas, G.~Synnaeve, I.~Misra, H.~Jegou, J.~Mairal, P.~Labatut, A.~Joulin, and P.~Bojanowski, ``Dinov2: Learning robust visual features without supervision,'' 2023.

\bibitem{Sun2022AGSCC}
Y.~Sun, L.~Lei, D.~Guan, J.~Wu, G.~Kuang, and L.~Liu, ``Image regression with structure cycle consistency for heterogeneous change detection,'' \emph{IEEE Transactions on Neural Networks and Learning Systems}, vol.~35, no.~2, pp. 1613--1627, 2024.

\end{thebibliography}

\end{document}